\documentclass[10pt,twocolumn,letterpaper]{article}

\usepackage{3dv}

\usepackage{times}
\usepackage{epsfig}
\usepackage{graphicx}
\usepackage{amsmath}
\usepackage{amssymb}

\usepackage{placeins}
\usepackage{dblfloatfix} 
\usepackage{algorithm}
\usepackage{algorithmic}
\usepackage{array}
\usepackage{overpic}
\usepackage{bm}
\usepackage{wrapfig}

\usepackage{pgfplots}
\usepackage{pgf}
\usepackage{tikz}

\newcommand{\M}{\mathcal{M}}
\newcommand{\N}{\mathcal{N}}

\newcommand{\LM}{L^2(\M)}
\newcommand{\LN}{L^2(\N)}

\threedvfinalcopy 
\def\threedvPaperID{81} 
\def\httilde{\mbox{\tt\raisebox{-.5ex}{\symbol{126}}}}

\ifthreedvfinal\fi
\begin{document}

\title{High-Resolution Augmentation for Automatic Template-Based Matching of Human Models}

\author{Riccardo Marin\\
University of Verona\\
Strada le Grazie 15, Verona, Italy\\
{\tt\small riccardo.marin\_01@univr.it}
\and
Simone Melzi\\
University of Verona\\
Strada le Grazie 15, Verona, Italy\\
{\tt\small simone.melzi@univr.it}
\and
Emanuele Rodol\`a\\
Sapienza University of Rome\\
Via Salaria, 113, Roma, Italy\\
{\tt\small rodola@di.uniroma1.it}
\and
Umberto Castellani\\
University of Verona\\
Strada le Grazie 15, Verona, Italy\\
{\tt\small umberto.castellani@univr.it}
}

\maketitle

\begin{abstract}
   We propose a new approach for 3D shape matching of deformable human shapes. Our approach is based on the joint adoption of three different tools: an intrinsic spectral matching pipeline, a morphable model, and an extrinsic details refinement. By operating in conjunction, these tools allow us to greatly improve the quality of the matching while at the same time resolving the key issues exhibited by each tool individually. In this paper we present an innovative \emph{High-Resolution Augmentation} (HRA) strategy that enables highly accurate correspondence even in the presence of significant mesh resolution mismatch between the input shapes. This augmentation provides an effective workaround for the resolution limitations imposed by the adopted morphable model. The HRA in its global and localized versions represents a novel refinement strategy for surface subdivision methods. We demonstrate the accuracy of the proposed pipeline on multiple challenging benchmarks, and showcase its effectiveness in surface registration and texture transfer.
\end{abstract}

 \begin{figure*}[h]
 \begin{center}
  \begin{overpic}
  [trim=0cm 0cm 0cm 0cm,clip,width=0.80\linewidth]{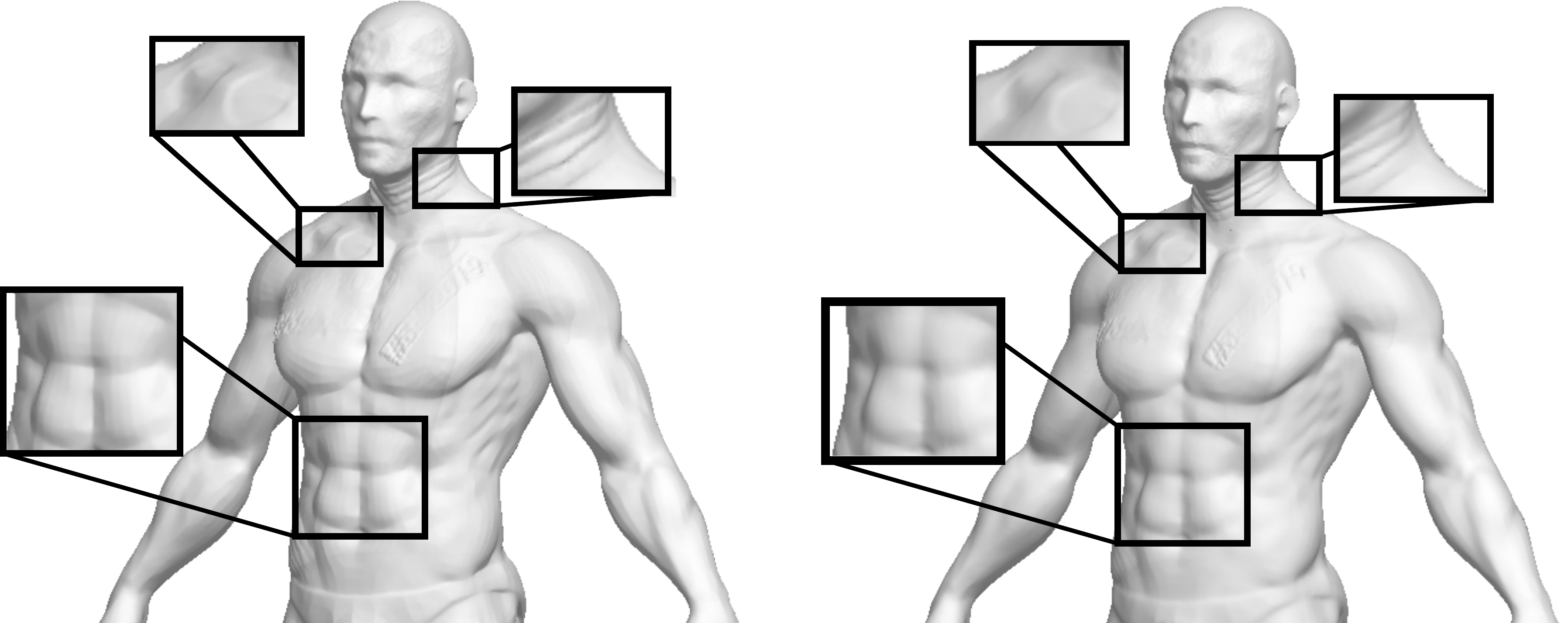}
  \put(23.5,41){\footnotesize Target}
  \put(76,41){\footnotesize HRA}


  \end{overpic}
      \caption{Example of a registration result obtained with our pipeline. We show the original (target from BadKing.com.au) surface on the left, and our registered result on the right. High-resolution geometric details (such as the wrinkles on the neck) are fully recovered by our pipeline, as put in evidence by the zoom-ins.}
    \label{fig:teaser}
\end{center}
\end{figure*}
\section{Introduction}
\label{sec:introduction}
Accurate shape matching is an essential tool in several applications in 3D vision and graphics, including shape registration~\cite{Hirshberg}, pose transfer~\cite{levy2006laplace}, shape remeshing~\cite{CMH} and shape modelling~\cite{Sumner} among others. 
A powerful direction to solve shape matching is provided by spectral geometry processing techniques.
In the spectral setting, the search for correspondences can be formulated as a matching problem in a higher-dimensional embedding space~\cite{ovsjanikov2012functional}.
The functional maps framework~\cite{ovsjanikov2012functional} constitutes an efficient and effective solution for spectral shape matching, and has been extended by several follow-up works \cite{rodola2017regularized,Nogneng2017,funAlgebra,zoomout}. To date, spectral approaches based on the functional maps represent the state of the art in deformable shape matching.
However, due to their band-limited representation, such approaches often suffer from poor point-wise re\-so\-lu\-tion especially in areas of high geometric detail. To face this limitation, Marin~\etal~\cite{FARM} proposed to combine the benefits of a 3D Morphable Model with those of functional maps, thereby unifying the intrinsic description of a surface with a strong extrinsic regularization. By doing so, the authors showed significant improvement over prior work in several challenging cases.

Despite these advantages, the approach of \cite{FARM} can not go beyond the resolution of the underlying parametric model, which can be a significant limiting factor whenever retaining the full resolution of the target shape is a strong requirement.
%
%
In this paper we propose a method to increase the resolution of the registered template, thus overcoming a key limitation of existing approaches as shown on a real example in Fig.~\ref{fig:teaser}. 
We show how our shape matching method can achieve very high-resolution correspondence, and we apply it to the tasks of shape registration and texture transfer. To summarize, the main contributions of our work are the following:
\begin{enumerate}
    \item We present a novel refinement strategy for shape re\-gi\-stra\-tion exploiting the advantages in using a surface subdivision scheme. This represents an innovative solution to face limitations imposed by the parametric model-based matching pipelines. 
    \item We provide insights about the capabilities of different 3D parametric models for the task of shape matching under different resolutions.
    \item We propose a novel, highly accurate matching pipeline for 3D human shapes; our pipeline is especially robust to dramatic changes in {\em mesh connectivity}, allowing to work with different levels of detail;
    \item Our pipeline considerably improves performance over the state-of-the-art methods in shape matching;
\end{enumerate}



\section{Related work}
\label{sec:related}
In 3D computer vision and graphics, shape matching is among the most widely studied topics due to its numerous applications. A complete review of all the existing shape matching techniques is out of the scope of this paper. For this reason, we limit our analysis to the methods that are most related to our work.

Descriptor-based shape matching methods are probably the most commonly used. The goal of descriptors is to define a compact, efficient, and as discriminating as possible representation for each point of the mesh. Correspondence can then be established by a nearest-neighbor assignment in descriptor space. 
A plethora of descriptors have been proposed in the literature. Some of them have been extended from computer vision to geometry processing such as spin images~\cite{Spin} or shape distributions~\cite{Osada}. The spectral representation provided by the Laplacian eigendecomposition is also at the basis of a large family of descriptors. We mention here the global point signatures \cite{GPS}, diffusion geometry-based signatures such as heat~\cite{sun2009concise} and wave kernel signatures~\cite{aubry2011wave}, descriptors defined upon the windowed Fourier transform \cite{WFT}, as well as its anisotropic version \cite{AWFT}. Other descriptors encode extrinsic properties of the surface, as done by the SHOT descriptor both in its original \cite{SHOT} and improved version \cite{CVPR}.

Descriptors often find use within more sophisticated pipelines, such as those based on seeking for an alignment between the spectral embeddings of the input shapes (also known as spectral matching). A tremendously effective solution for this revolves around the functional maps representation~\cite{ovsjanikov2012functional}. The key idea is that solving for a correspondence in the {\em functional} (as opposed to point-to-point) domain can be much more efficient and stable. Due to its efficiency and compactness, functional maps have been exploited in several different settings, including partial shape matching \cite{rodola2016partial_old}, non-isometric matching \cite{cosmo2019isospectralization} and deep learning-based pipelines for dense correspondence \cite{FMNET}; we refer to \cite{ovsjanikov16} for a more complete survey on functional maps. Several works also attempt to improve the quality of the mapping by extending the spectral basis to induce better function approximation or promote point-wise maps~\cite{Nogneng2017,funAlgebra}. More recently, an efficient algorithm for functional map optimization (called ZoomOut) \cite{zoomout} has been proposed; differently from previous approaches, this method is able to infer a precise high-frequency map from a very low-pass initialization.

Recently, several works highlight the necessity of looking for matches between the vertices of a mesh and the entire surface seen as a collection of triangles (sub-vertex or ``precise'' matching). The main rationale of such approaches is that vertex-to-vertex correspondences directly rely on the discrete representation of the underlying continuous surface in addition to mesh connectivity. In \cite{ezuz2017deblurring}, the authors propose a conversion from vertex-to-vertex to vertex-to-surface map given functional maps as input, together with a denoising and deblurring strategy. The recent SHREC'19 Connectivity benchmark~\cite{SHREC19} explored such a setting for human shape correspondence, where each mesh has a dramatically different connectivity. It turns out that most state-of-the-art approaches heavily depend on discretization and connectivity, suggesting that much still needs to be done in this direction.

Most closely related to this paper are the works \cite{deformation} and \cite{FARM}. Both works build upon the general idea of a deformation-driven approach for shape matching. 
%
In particular, the FARM method~\cite{FARM} produces a high-quality correspondence through an iterative deformation of a given parametric model (the authors adopt the SMPL~\cite{loper15} morphable model) into a given target shape. The approach relies on a spectral matching step based on the functional maps representation. The pipeline automatically detects 5 landmarks~\cite{bestpaper}, and proceeds with a two-stage process. The first stage seeks for a registration between the parametric model and the target shape; supposing to be now more isometric to the target shape, the second operates between the deformed version of the parametric model and the target shape itself. When the input consists of two shapes to be matched one to another, a map between them can be composed by the maps computed toward the parametric model.

\section{Background}\label{sec:background}
%
We first introduce some mathematical preliminaries that will be instrumental for the discussion of our method.

\vspace{1ex}\noindent\textbf{Continuous human shapes.}
We represent a human shape through its external surface, modeled as a compact and connected smooth 2-dimensional Riemannian manifold $\M$ (possibly with a boundary $\partial \M$) embedded into $\mathbb{R}^3$. This object is naturally equipped with a standard metric~\cite{docarmo}, which induces an inner product $\langle f, g \rangle_{\M} = \int_\M f(x)g(x) dx$, for every pair of functions $f$ and $g$ defined on $\M$. We denote by $L^2(\M) = \{ f:\M\to \mathbb{R}~ \vert~ \int_\M |f(x)|^2 dx < \infty \}$ the space of square-integrable real functions on $\M$.

The Laplace-Beltrami operator (LBO) $\Delta_\M : L^2(\M)\to L^2(\M)$ is a positive semi-definite operator that corresponds to the standard Laplace operator on Euclidean domains. The LBO admits an eigendecomposition which gives rise to an extended Fourier analysis on manifolds.
In particular, we have ${\Delta}_\M \phi_k = \lambda_k \phi_k$, where $0=\lambda_1 \leq \lambda_2 \leq \hdots$ are real eigenvalues, and $\{\phi_k\}_{k\ge1}$ are the corresponding eigenfunctions forming an orthonormal basis of $L^2(\M)$.
In analogy to classical Fourier analysis, any function $f \in L^2(\M)$ can be expanded in the Laplacian eigenbasis as:
\begin{align}
f(x) = \sum_{k\geq 1} \langle f, \phi_k \rangle_{\M} \phi_k(x)\,.
\label{eq:fourier}
\end{align}

\vspace{1ex}\noindent\textbf{Discretization.}
In the discrete setting, $\M$ is represented as a triangular mesh $(\mathcal{V}, \mathcal{E})$, where $\mathcal{V}$ is the set of  vertices and $\mathcal{E}$ is the list of  edges. 
 If $n_{\M}$ is the number of vertices of $\M$, then a function defined on $\M$ is represented by a vector of length $n_{\M}$ which contains the value of the function at the corresponding vertex. In the discrete case, the LBO is represented as a $n_{\M} \times n_{\M} $ matrix $\bm{\Delta}_{\M} = \mathbf{A}_{\M}^{-1} \mathbf{W}_{\M}$. The {\em mass} matrix $\mathbf{A}_{\M}$ is a diagonal matrix with the area elements associated with each vertex.
The {\em stiffness} matrix $\mathbf{W}_{\M}$ locally encodes the geometry of the surface around each vertex. We refer to \cite{pinkall1993computing} for additional details.

\vspace{1ex}\noindent\textbf{Functional maps}~\cite{ovsjanikov2012functional} is a framework that allows to robustly and compactly transfer information among surfaces. Here we only provide a short overview of the original framework, and refer to \cite{ovsjanikov2012functional,ovsjanikov16} for a more thorough exposition.
Let $\M$ and $\N$ a pair of shapes with $n_{\M}$ and  $n_{\N}$ vertices, and let $\{\phi_h\}_{h\ge1},\{\psi_l\}_{l\ge1}$ be the respective LBO eigenfunctions.
Given a point-wise map $\pi: \N\to\M$, the functional maps framework converts the matching problem (i.e. the search for a precise estimation of $\pi$) from the point-to-point domain to a mapping between functional spaces. This conversion is obtained through the linear operator $T: \LM \to \LN$, defined as $T(f) = f \circ \pi$. 
In the Laplacian eigenbases, the operator $T$ is encoded by a matrix $\mathbf{C}=(c_{hl})$, the entries of which are explicitly defined by:
\begin{equation}\label{eq:cij}
c_{hl} = \langle T{\phi}_h , {\psi}_l \rangle_\N \,.
\end{equation}
Therefore, the matrix $\mathbf{C}$ represents the change of basis transformation between the two functional spaces. More formally:
\begin{equation}\label{eq:c}
T(f) = T \sum_{h} \langle f, {\phi}_h \rangle_\M {\phi}_h = \sum_{hl} c_{hl} \langle f, {\phi}_h \rangle_\M \psi_{l} .
\end{equation}

As suggested in \cite{ovsjanikov2012functional}, we truncate the summations in Equation~\eqref{eq:c} to fixed values $k_{\M}$ and $k_{\N}$ respectively. This  corresponds to considering only the first $k_{\M}$ and $k_{\N}$ eigenfunctions as truncated bases for the functional spaces, giving rise to a band-limited representation of the functions. This leads to an especially compact representation of the correspondence, which is now encoded as a small matrix $\mathbf{C} \in \mathbb{R}^{k_{\N}\times k_{\M}}$ (the values of $k_{\M}$ and $k_{\N}$ are usually small and do not exceed a few hundreds); compared to the classical representation as a point-to-point correspondence matrix of dimensions $n_{\N}\times n_{\M}$, having such a $\mathbf{C}$ with $k_{\M} \ll n_{\M}$ and $k_{\N} \ll n_{\N}$ allows solving correspondence problems much more efficiently.

\section{Our method}
\label{sec:method}
In \cite{FARM}, Marin~\etal proposed an automatic pipeline for human body shape registration. Their method is based on three main steps.
Given a parametric model for humans (the authors used SMPL~\cite{loper15}, see an example in Figure~\ref{fig:mapcomparison}), a target body shape and five landmarks (head, hands and feet), the method allows to find an accurate functional correspondence between the two shapes. This is used to deform the template closer to the target surface, and repeat the process to improve the matching quality. The final registration is obtained after this two-stage pipeline.

The pipeline above has two major limitations: (1) The poor runtime efficiency due to the two iterative registration steps; and (2) the reliance of the overall registration quality upon the connectivity of the adopted parametric model (just about $\sim$7K vertices in the case of SMPL).
In the sequel we show how to address these limitations, at the same time yielding significant improvement in standard applications.

\subsection{One-step dense correspondence} 
The problem of estimating a high-quality functional correspondence was recently addressed by the elegant refinement method proposed in \cite{zoomout}, and named ZoomOut. This method starts from a given initial (functional) matching between the two shapes $\N$ and $\M$, and iteratively performs the following two steps:
\begin{enumerate}
\item convert the $k_{\N} \times k_{\M}$ functional map to a pointwise map;
\item convert the pointwise map to a $k_{\N}+1\times k_{\M}+1$ functional map;
\end{enumerate}
The process proceeds while increasing the values of $k_{\N}$ and $k_{\M}$ at each iteration. The map obtained through this iterative approach is completely determined by the initial map; remarkably, it is a descriptor-free algorithm. It was further observed that, as more and more high frequencies are included in the functional map representation, the level of geometric detail accurately mapped by the estimated correspondence also increases. In our experiments, we show that it is possible to directly optimize the registration of the parametric model by applying this kind of refinement to the matching obtained in the first step of \cite{FARM}.
FARM also adopts a refinement strategy in order to filter out coarse mismatches. We show that, thanks to the matching step described above, this additional refinement can be avoided.

We refer to Figure~\ref{fig:mapcomparison} for a qualitative evaluation of ZoomOut on our data. In the figure, we compare the matching provided by the initial functional map of size $50\times 30$, the two rounds of the refinement proposed in FARM, and the ZoomOut correspondence. The matching quality of ZoomOut can be further appreciated in comparison with the previous matching pipeline in Figure~\ref{fig:FAUST}. These results confirm that to estimate good correspondences it is not necessary an intermediate registration to get more isometric shapes. For this reason we can avoid the two rounds estimation strategy adopted in FARM. We have empirically seen that this let us to save around $30$\% of time.

 \begin{center}
 \begin{figure*}[t!]
  \vspace{0.5cm}
  \begin{overpic}
  [trim=0cm 0cm 0cm 0cm,clip, width=0.95\linewidth]{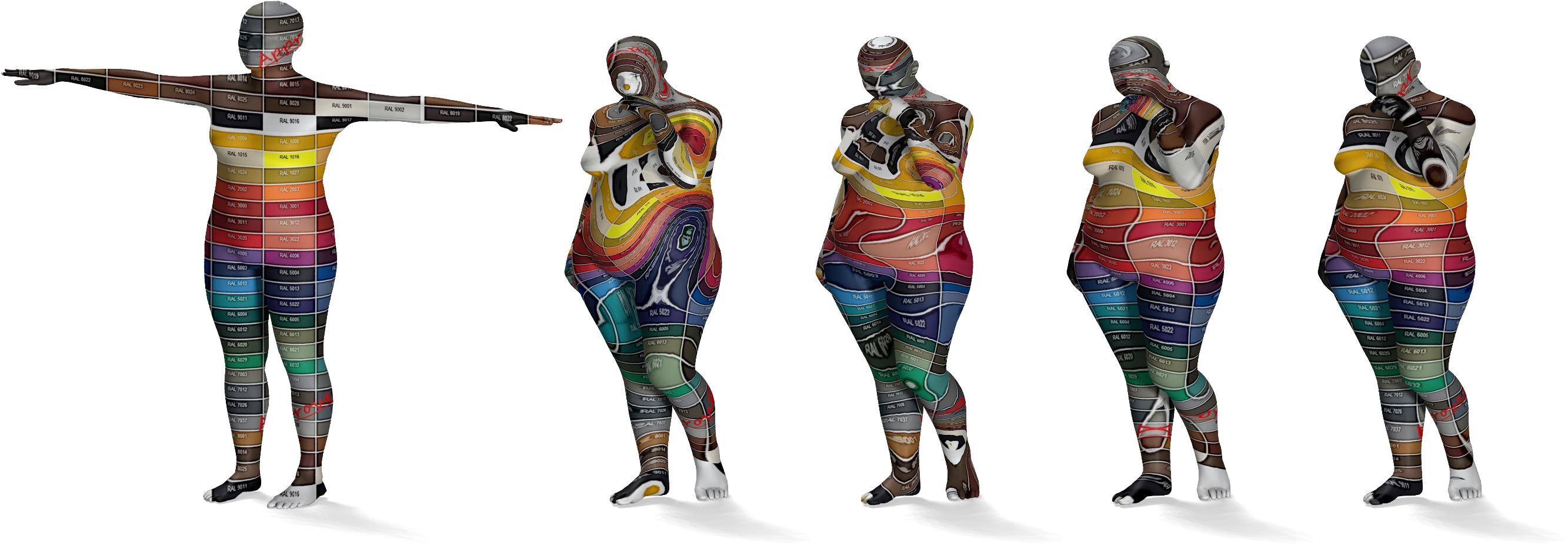}
  \put(12.5,1){\footnotesize Source}  \put(39.3,1){\footnotesize Init}
  \put(55.8,1){\footnotesize R1}
  \put(71.8,1){\footnotesize R2}
  \put(86.8,1){\footnotesize \textbf{Our}}
  \end{overpic}
      \caption{A ruined FMAP correspondence in a non isometric case visualized through texture transfer. From left to right: the source texture visualized on the SMPL model; the initialization of the correspondence (Init); the map refined in the first round of FARM (R1); the map refined in the second round of FARM (R2); Our refinement (Our). R1 and Our start on the same Init correspondence.}
    \label{fig:mapcomparison}
\end{figure*}
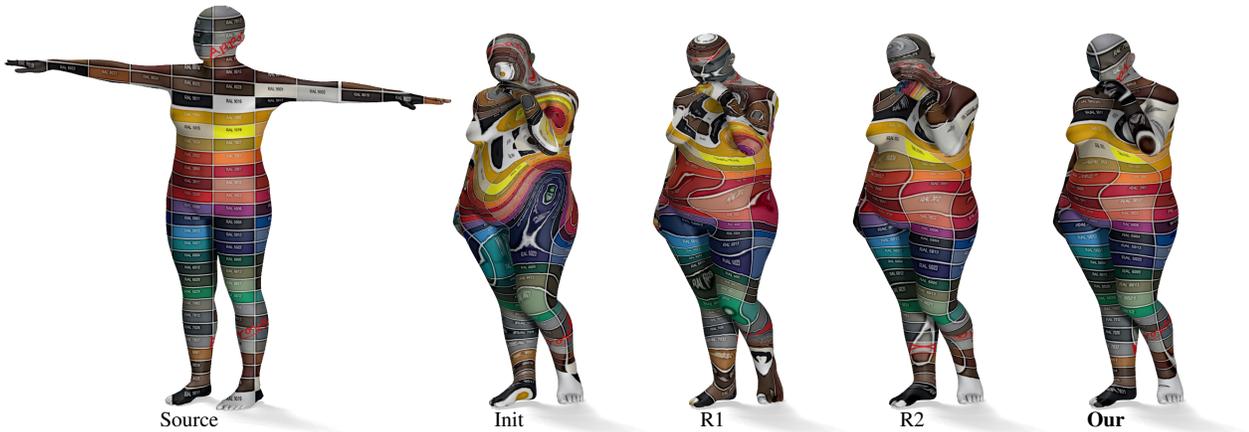
\end{center}
%


\subsection{High-Resolution Augmentation}
The main novelty of the proposed method is the High-Resolution Augmentation (HRA).
To highlight the contribution of this step, we first consider standard results with a parametric model.
A parametric model is nothing more than a fixed template, for which a set of parameters govern a group of deformations. 
Every deformation of the model is obtained over the same template, providing the same connectivity for all the generated shapes. As a direct consequence, the model has a fixed resolution in all its poses.

%
%
\begin{center}
 \begin{figure*}[t!]
 \vspace{0.5cm}
 \centering
\begin{overpic}[trim=0cm 0cm 0cm 0cm,clip,width=0.85\linewidth]{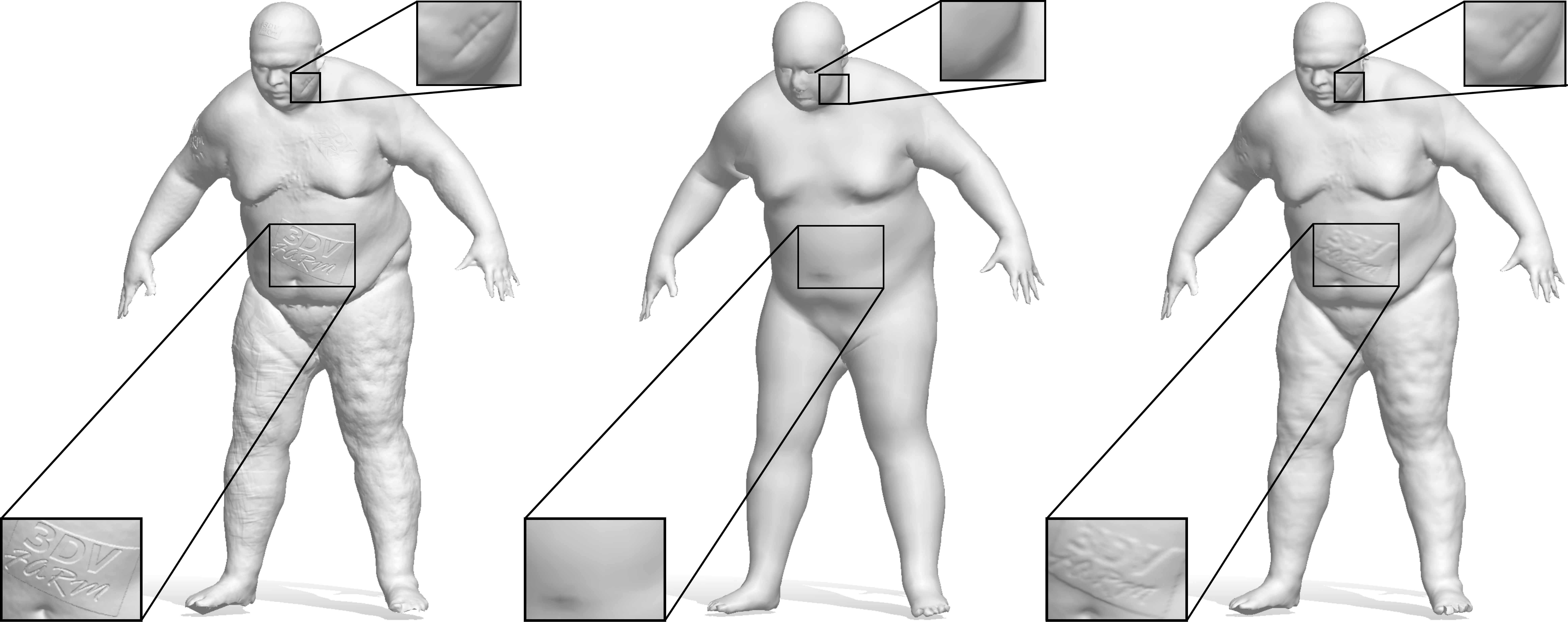}
\put(14.3,-1.2){\footnotesize Original ($2.1$M)}
\put(48.4,-1.2){\footnotesize FARM ($6.8$K)}
\put(82.4,-1.2){\footnotesize HR ($421$K)}
\end{overpic}
\vspace*{2mm}
      \caption{A comparison on registering a highly detailed real scan from \cite{dfaust} adorned with artist details and patterns. On the left the target, (2,1 million vertices), in the middle FARM registration (6,8 thousands vertices) and on the right our result (421 thousands vertices). Using one-fifth of the target resolution, we are already able to acquire the finest details.}
    \label{fig:details}
\end{figure*}
\end{center}

The quality of the details that can be recovered by the parametric model is thus limited to the ones that can be represented by its connectivity.
As can be seen in Figure~\ref{fig:details}, the registration obtained by FARM is drastically inferior to HRA; SMPL has few vertices (6890) to catch all the details.
In contrast, HRA allows us to achieve a very high-quality registration that can finely reproduce all the details encoded by the scan.

HRA is applied to the template once its registration to the target shape is concluded.
Then a subdivision method is applied to the template (3 times recursively)  obtaining a mesh with a larger number of vertices. We alternate these iterations with a minimization of the As-Rigid-As-Possible (ARAP) energy \cite{Igarashi05} in order to fit the extrinsic details of the geometry of the target shape.

In our experiments, we compare three different subdivision methods: the Barycentric Subidvision (BCS), Upsample~\cite{SubSur00}, and the popular  Loop~\cite{Loop87}. 
\setlength{\columnsep}{0pt}
\setlength{\intextsep}{0pt}
\begin{wrapfigure}[9]{r}{0.25\textwidth}
\begin{center}
\begin{overpic}[trim=0cm 0cm 0cm 0cm,clip,width=0.93\linewidth]{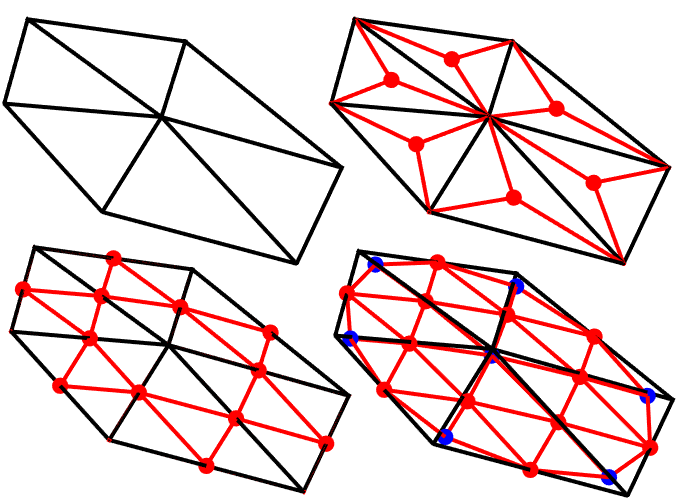}
\put(12,73){\footnotesize{Input}}
\put(6,-3){\footnotesize{Upsample}}
\put(64,-3){\footnotesize{Loop}}
\put(66,73){\footnotesize{BCS}}
\end{overpic}
\end{center}
\end{wrapfigure}
For completeness here  we briefly describe the three subdivision methods, highlighting the difference between them. A visualization of the three techniques is provided in the inset Figure, where the initial edges and vertices of the mesh are depicted in black while the newly added edges and vertices are depicted in red and blue respectively.

BCS splits each triangle by adding for each face a vertex in the position of the barycenter. Then, the original triangle is substituted by the three smaller triangles where each original vertex is connected to the barycentric one. We provide our own implementation for BCS, which we consider as a baseline since it is the simplest approach among the three. The main drawback of this method is that at each iteration, the triangle aspect ratio increases (\ie, the ratio between the shortest and longest edge). For this reason, the ARAP energy becomes quite unstable during our optimization, because it relies on a {\em rigid} preservation of the edge lengths. 
With BCS, in the current formulation we can apply the subdivision at most two times. We also noticed that the registration results have artifacts arising from the wild connectivity.

The Upsample method~\cite{SubSur00} requires to add a vertex for each edge of the mesh. In this scheme, each triangle is subdivided in four sub-triangles. If we consider the older vertices as \emph{Even} and the new ones as \emph{Odd}, Upsample adds the \emph{Odds} without modifying the \emph{Evens} positions. In this case we have a more regular mesh compared to BCS. As a drawback, we note that Upsample flatten large areas of the surface. Without a smoothing, the triangles becomes smaller but they get stuck in their rigidity relation. We observed this method is more stable than BCS, and it allows us to perform three iterations of subdivision without energy collapse.

Finally we tried Loop ~\cite{Loop87}, the more sophisticated and popular method for surface subdivision. This approach adopts the same strategy as Upsample by adding a vertex on each edge and splitting the faces in four triangles. Then, a smoothening step is performed where Even and Odd positions are recomputed as the weighted means of their neighborhoods. This method yields stable results, does not give rise to evident artifacts, and furthermore it injects non-rigid changes, permitting to the ARAP energy to start over in case it gets locked by locally strong deformations.

We select the Loop method mainly due to the latter observation, and to the quality of the results it is able to provide (see Figure~\ref{fig:subdivions} for comparisons). The final mesh produced by three iterations of Loop subdivision has a number of faces that is equal to $4^{3}$ times the initial number of faces of the template (\eg, SMPL grows from 13,776 to 881,664 triangles). 
 \begin{center}
  \begin{figure}[t!]
  \vspace{0.5cm}
  \begin{overpic}
  [trim=0cm 0cm 0cm 0cm,clip,width=1\linewidth]{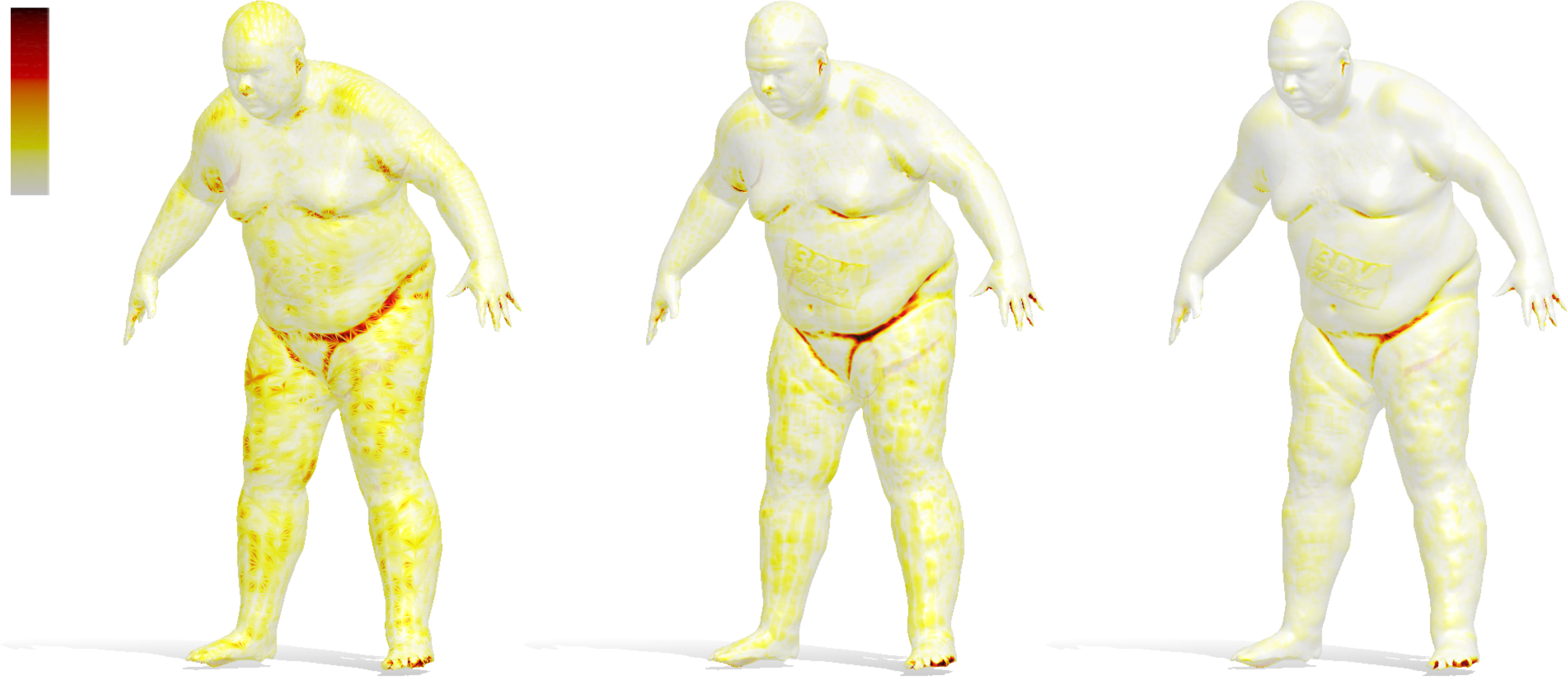}
  \put(3,29.5){\tiny{0mm}}
  \put(3,43){\tiny{5mm}}
  \put(15,44.5){\footnotesize{BCS}}
  \put(46,44.5){\footnotesize{Upsample}}
  \put(83,44.5){\footnotesize{Loop}}
  \end{overpic}

      \caption{Comparison of different subdivision methods. From left: BCS, Upsample, Loop. Baycentric is strongly penalized by its rigid structure that causes ARAP regularization to become unstable after 2 subdivision. Colors represent distance of template to target surface, with saturation at 5 millimetres.}
    \label{fig:subdivions}
\end{figure}
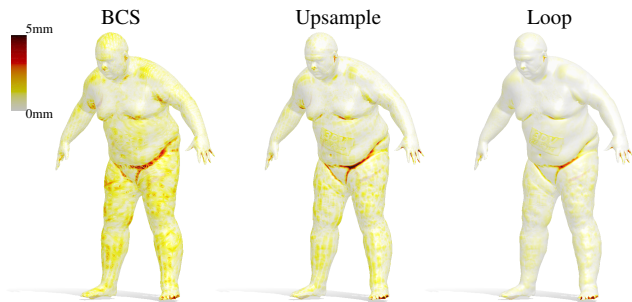
\end{center}
%

Our approach is therefore an iterative combination of subdivision and optimization. Details are captured progressively, and the smoothness induced by Loop subdivision at each iteration allows to meet the ARAP constraints as the surface gets closer to the target geometry.


\subsection{Localized High-Resolution Augmentation}
As can be seen in Figures~\ref{fig:teaser} and~\ref{fig:details}, the most of the details presented by a shape are localized in small regions such as the face traits, the sharp abs on the belly or local pattern like cellulite.
This suggests to us that a detail refinement over the whole shape is an overkill, and the problem can be better addressed by refining only proper local regions. Local refinement rises two main problems: firstly, we need to estimate automatically the regions where to apply our refinement. Secondly, Loop~\cite{Loop87} subdivision method is not naturally applicable locally, since it generate vertices over the edges that require linkage with other triangles.
We propose a strategy that is aimed to solve both these problems. 
By our experiments we found that the \emph{mean curvature} ($H$) is a good indicator for high detailed region. We select over the target the regions where $|H|>0.03$.
Then, we project these regions over the template using nearest neighbor, and we propagate the selection over a surrounding geodesic circumference.
At this point, we would subdivide only these local patches, and then re-attach them coherently to the unaffected surface. To do this, we identify the Odd vertices on the boarder of the subdivided patches. These are the new vertices that belong both to one subdivided face and to one that is not subdivided. We link these vertices with the \emph{opposite vertex on the face unaffected by subdivision}, and consequently we split the face in two new faces.
The only special case we need to handle is when an outer triangle is adjacent to more then one subdivided triangles. When it happens, we include all these outer triangles into the subdivided region. We proceed including outer triangles until this anomaly has been fixed.
We would like to remark that the new surface is still in correspondence with the older one, and so with all other meshes locally subdivided in this way.
%
This local version of the HRA  provides a more efficient fitting to the details of a given target shape. Also, it constitutes a new method for the local surface subdivision.

 \begin{center}
  \begin{figure}[t!]
  \vspace{0.5cm}
  \begin{overpic}
  [trim=0cm 0cm 0cm 0cm,clip,width=1\linewidth]{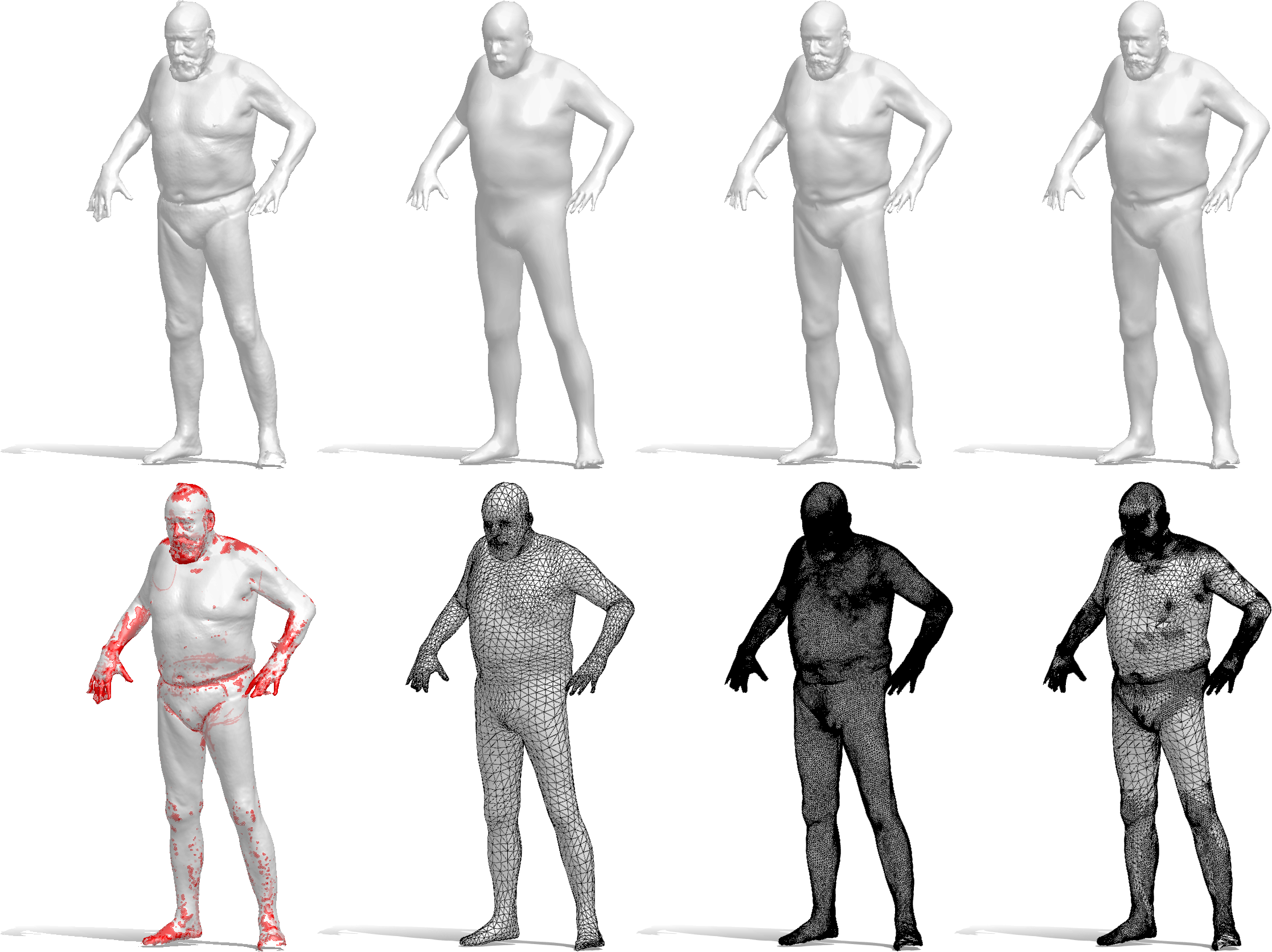}
  \put(15,-3){\footnotesize{190K}}
  \put(39.5,-3){\footnotesize{6.8K}}
  \put(65,-3){\footnotesize{110K}}
  \put(91,-3){\footnotesize{68K}}
  
  \put(13,76){\footnotesize{Target}}
  \put(38,76){\footnotesize{ZOSR}}
  \put(62.5,76){\footnotesize{HR}}
  \put(86,76){\footnotesize{Local}}
  \end{overpic}
  \vspace{0.05cm}
      \caption{Localized adaptive refinement. In the first column, the target and the regions with $|H|>0.03$. Then, we show the FARM output, HR and HRA. Notice how the majority of the details (e.g. face beard) have been caught with half of the vertices.}
    \label{fig:local}
\end{figure}
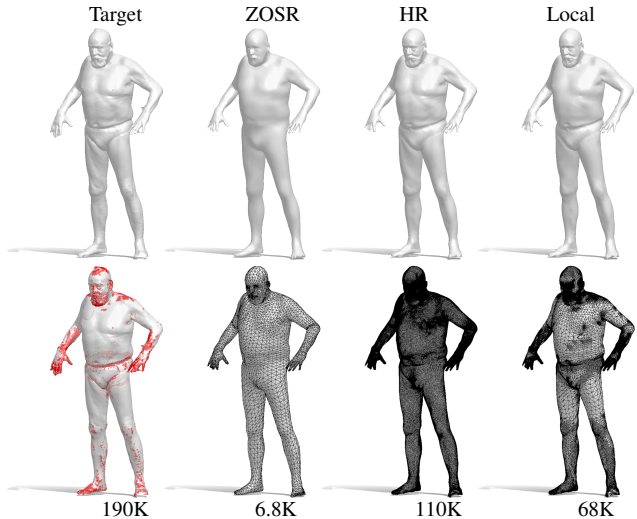
\end{center}
\section{Results}
\label{sec:results}
In this Section, we collect the experiments and applications of the proposed method. 
%
All the experiments are performed on MATLAB 2018, on a machine with 32GB of RAM and an Intel 3,6 GHz Core i7. The code of the proposed method  will be made publicly available upon acceptance.

\subsection{Point-to-point matching}
We evaluate our method in point-to-point matching task on FAUST~\cite{FAUST} and TOSCA~\cite{TOSCA} datasets.
We evaluate both with and without the use of High-Resolution Augmentation strategy (denoted as \emph{HR} and \emph{ZOSR} respectively).
We compare our results with 5 different state-of-the-art approaches: \emph{RMH}~\cite{RHM}, \emph{PMF}~\cite{PMF}, \emph{BCICP}~\cite{BCICP}, \emph{ZoomOut}~\cite{zoomout} and \emph{FARM}\cite{FARM}. 
All these methods refine the same initial matching that is added to the evaluations and denoted by \emph{Ini}. The ZoomOut matching is the one exploited by our methods for the parametric model registration.
Learning based-approaches are excluded for a fair comparison.
We evaluate the matching quality through the cumulative
error protocol proposed in~\cite{KimBIM}.
\begin{figure}[t!]
  \centering
  \input{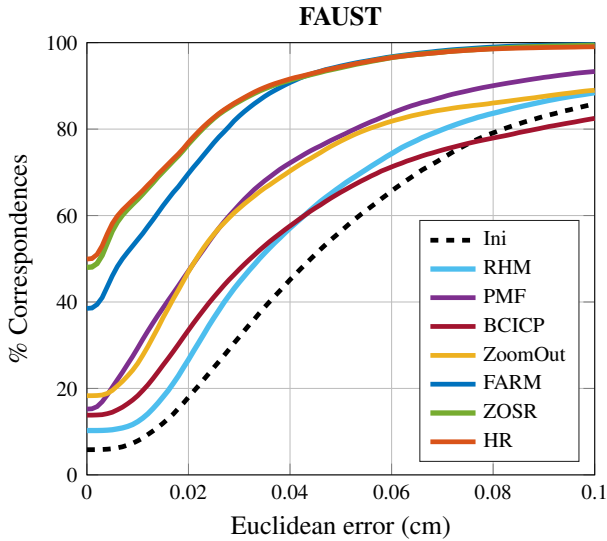}
\caption{Correspondence comparison curves over FAUST.}
    \label{fig:FAUST}
\end{figure}
In Figure~\ref{fig:FAUST}, we report the average on 10 pairs each of which is composed by one shape of FAUST and the SMPL template. The considered FAUST shapes are all different subjects in different poses in order to explore the non-isometric cases.
SMPL and FAUST shapes share the same connectivity thus it is possible to evaluate the matching quality.
As can be seen both ZOSR and HR outperform all the competitors and in particular FARM.
HR slightly improves ZOSR results, although SMPL owns the same connectivity of FAUST and thus the HRA is not necessary. This confirms that Loop subdivision allows us to achieve  better minimization of the ARAP energy. 

\begin{figure}[t!]
  \centering
  \input{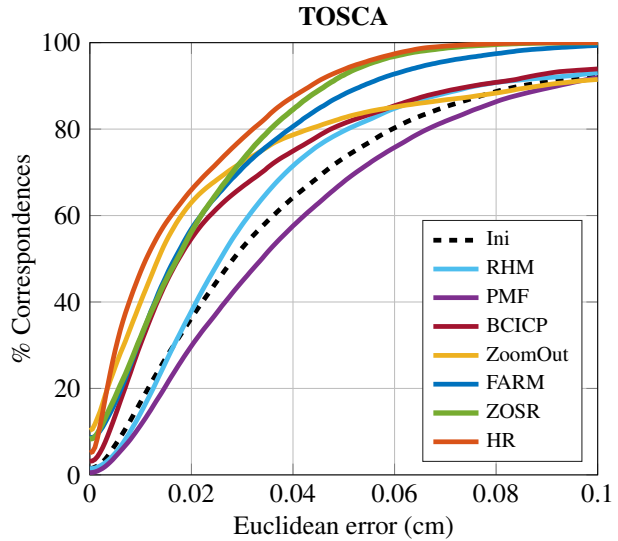}
\caption{Correspondence comparison curves over TOSCA.}
    \label{fig:TOSCA}
\end{figure}

In Figure~\ref{fig:TOSCA} we visualize the average comparison on 7 pairs of the David class of the TOSCA dataset. These shapes has the same connectivity and subject, but largely differ for the poses. Also in this case ZOSR and HR outperform all the competitors. The meshes of David from TOSCA contains around $52$K vertices that are many more than the 6890 of the SMPL mesh. For this reason, the improvement achieved by HR is more clear in this case. These result confirm that the HRA clearly improves the performance of the proposed method.
%
%
%
\subsection{Texture transfer}
In Figure~\ref{fig:SHREC}, we visualize three qualitative results of the proposed method in the texture transfer application underlying quantitative results showed in the previous paragraphs. We Consider three pairs of shapes from the 430 that compose the SHREC'19 Connectivity benchmark. These shapes present a large set of non-isometric deformations and connectivities. The texture transfer quality of our method can be appreciated on the fine details that we are able to transfer as the text \textit{``Approved''} highlighted in the zoom-in of the shapes in the middle. The high resolution obtained allows us to transfer also a picture of an artist as done for the pair on the right of the Figure \ref{fig:SHREC}.

\subsection{Human body registration}
We complete the registration of a large amount of shapes from various datasets, and also over some ad-hoc modified ones to test our capability in catching details. In Figure~\ref{fig:details} we have a variety of different local patterns: cellulite, synthetic letters, scars and also dynamic body tissues. 
All these details are well represented by our method. Also in Figure~\ref{fig:subdivions} we present a quantitative result on the same shape: the colors encode the distance between registered template and target surface. Few points saturate the error at 5 millimetres in all three subdivision strategies; it is also interesting notice how BCS and Upsample connectivity affect the geometry fitting (e.g. on the legs). Finally, in Figure ~\ref{fig:local} we show our adaptive strategy to local optimization. Our inference permits to use just half of the vertices to obtain the same result quality. 
We included further examples of human bodies registration in the \textbf{Supplementary Material}.

 \begin{center}
  \begin{figure*}[h]
  \setlength{\tabcolsep}{0pt}
  \begin{tabular}{lcr}
     
    \begin{minipage}{0.35\linewidth}
     
        \begin{overpic}
        [trim=0cm 0cm 0cm 0cm,clip,width=1\linewidth]{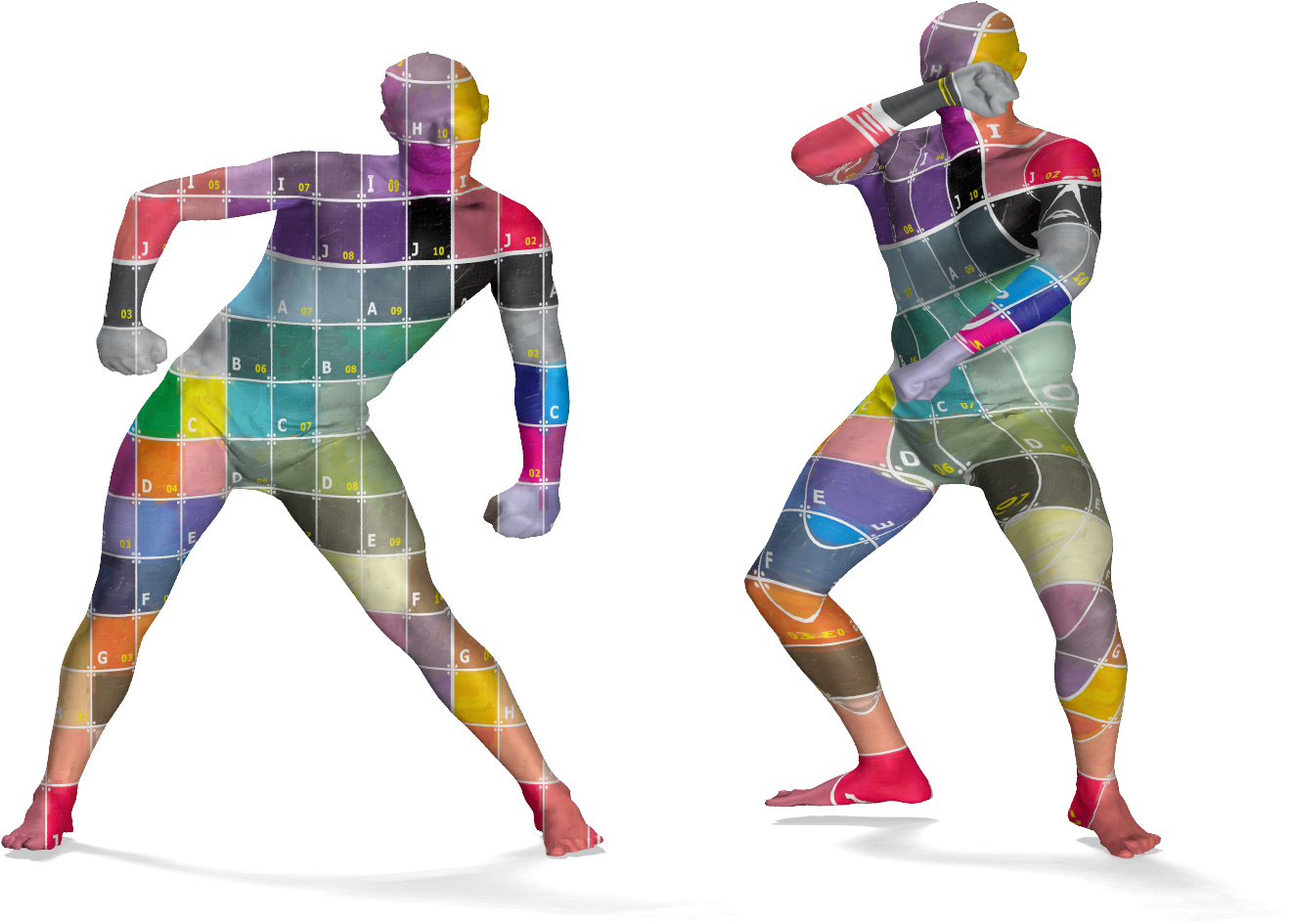}
\put(12,-2.5){\footnotesize Original}
       \put(64,-2.5){\footnotesize Transfer}        \end{overpic}
  
    \end{minipage}
    & 
    \hspace{0.05cm}  
    
    \begin{minipage}{0.32\linewidth}
        \begin{overpic}
        [trim=0cm 0cm 0cm 0cm,clip,width=1\linewidth]{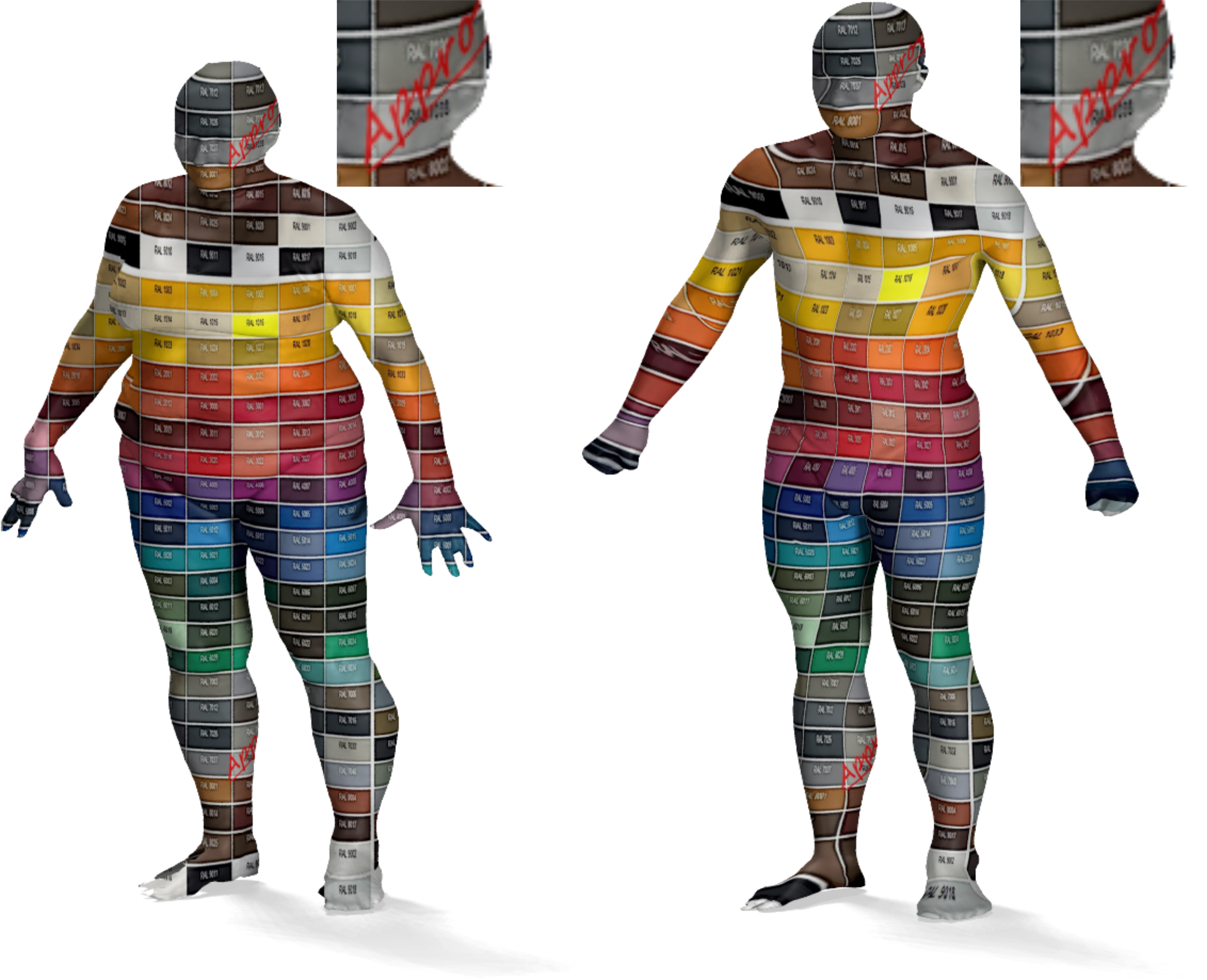}
        \put(12,-2.5){\footnotesize Original}
       \put(61,-2.5){\footnotesize Transfer}
        \end{overpic}

    \end{minipage}
    & 
    \hspace{0.05cm}  
    
  \begin{minipage}{0.33\linewidth}
   \vspace{0.05cm}
   
        \begin{overpic}
        [trim=0cm 0cm 0cm 0cm,clip,width=1\linewidth]{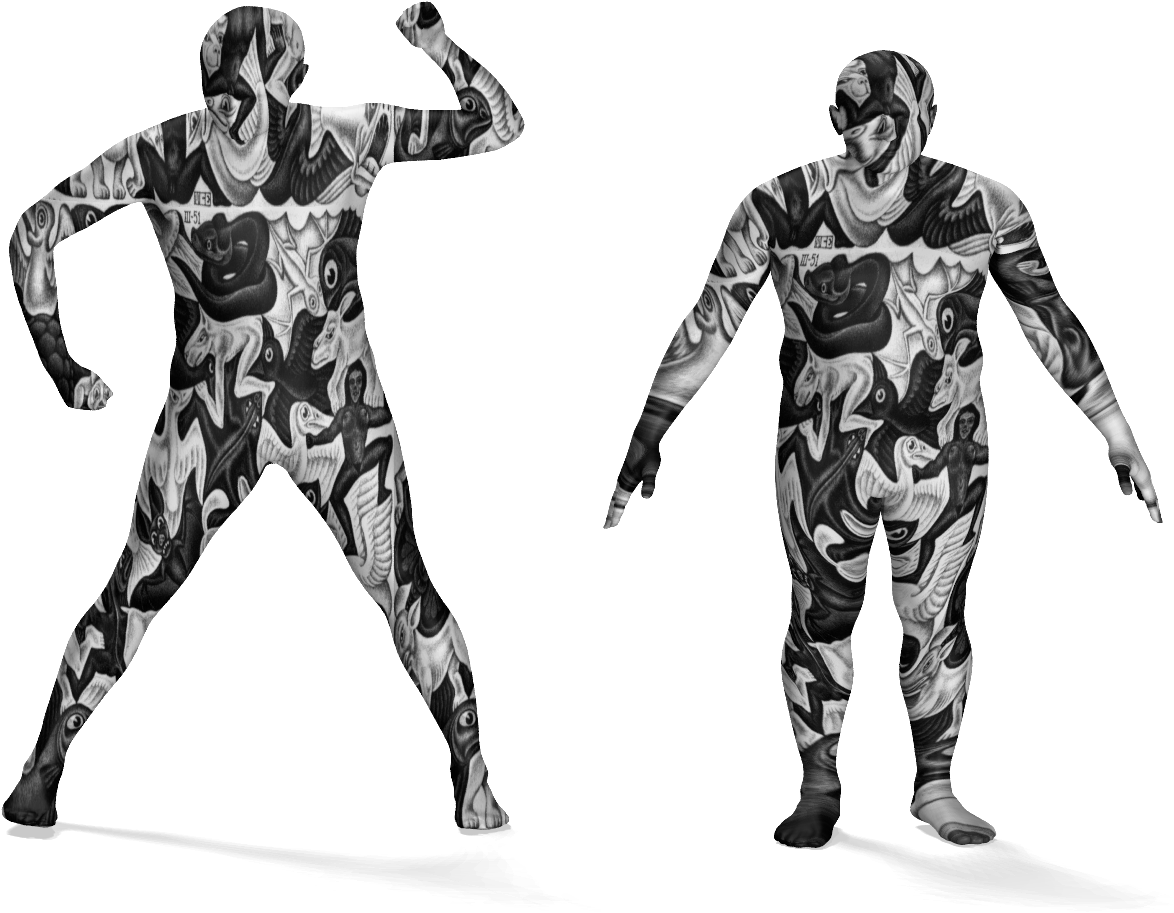}
\put(11,-2.5){\footnotesize Original}
       \put(65,-2.5){\footnotesize Transfer}        \end{overpic}

    \end{minipage}
    

    
    \vspace{0.3cm}
	
  \end{tabular}
    \caption{Our qualitative results in the texture transfer application on 3 different pairs from the SHREC`19 Connectivity benchmark~\cite{SHREC19}.}
    \label{fig:SHREC}
\end{figure*}
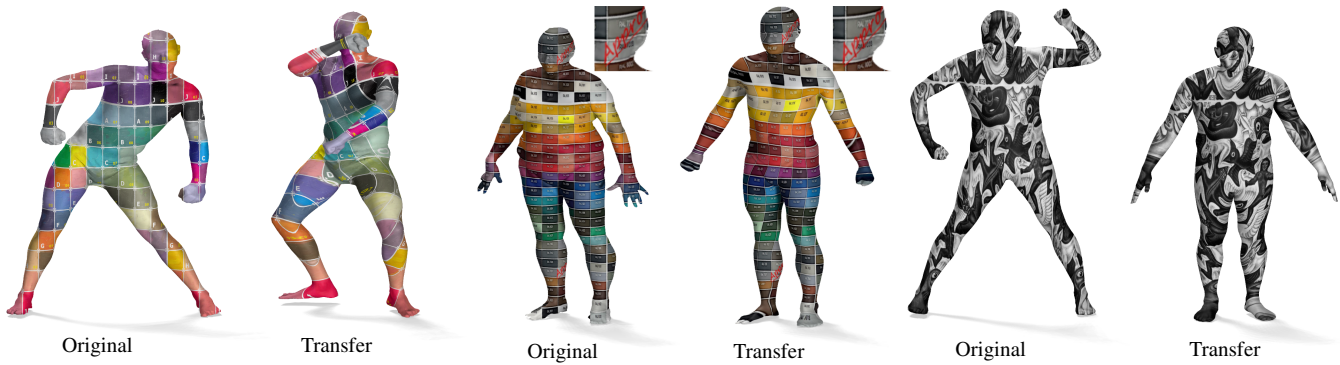
\end{center}
 \begin{center}
  \begin{figure*}[!h]
  \vspace{0.5cm}
  \begin{overpic}
  [trim=0cm 0cm 0cm 0cm,clip,width=1\linewidth]{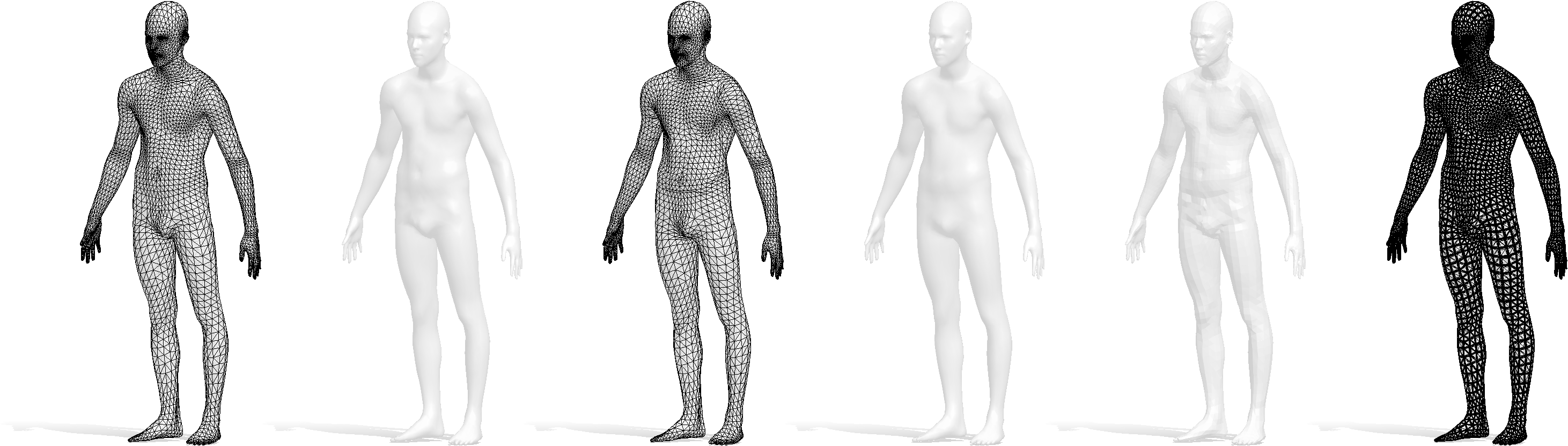}
  \put(16.5,26){\footnotesize Target}
  \put(50,26){\footnotesize FARM}
  \put(84.3,26){\footnotesize HR}

  \end{overpic}
      \caption{Registration of a low-resolution mesh with our HRA method. From left: target and its wiremesh; FARM result; HR result. }
    \label{fig:LOW}
\end{figure*}
\end{center}
 \begin{center}
  \begin{figure}[h!]
  \vspace{0.5cm}
  \begin{overpic}
  [trim=0cm 0cm 0cm 0cm,clip,width=1\linewidth]{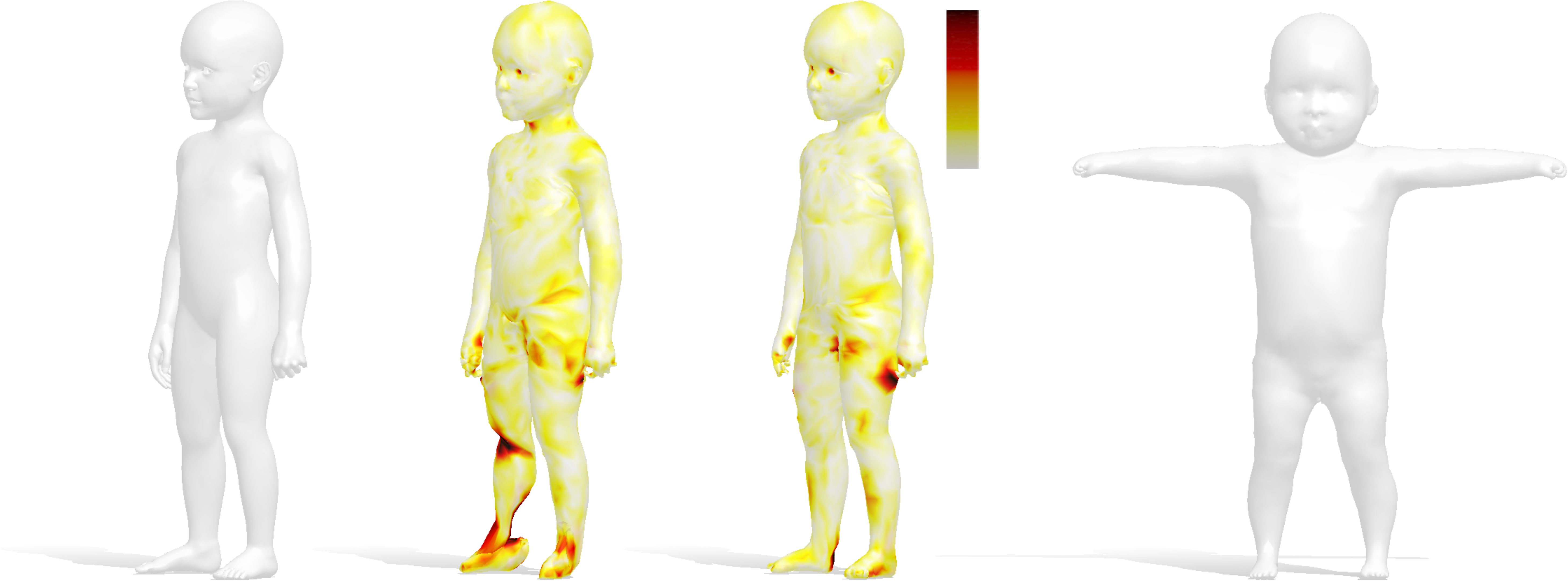}
  \put(10,39){\footnotesize{Target}}
  \put(30,39){\footnotesize{FARM}}
  \put(50,39){\footnotesize{ZOSR}}
   \put(81,39){\footnotesize{SMIL}}
   \put(62.5,37){\tiny{5mm}}
   \put(62.5,25){\tiny{0mm}}
   
  \end{overpic}
      \caption{Comparison using the SMIL morphable model. From left to right: the target shape, the FARM registration, our method. Finally, the template of SMIL. Colors represent distance of template to target surface, with saturation at 5 millimetres.}
    \label{fig:SMIL}
\end{figure}
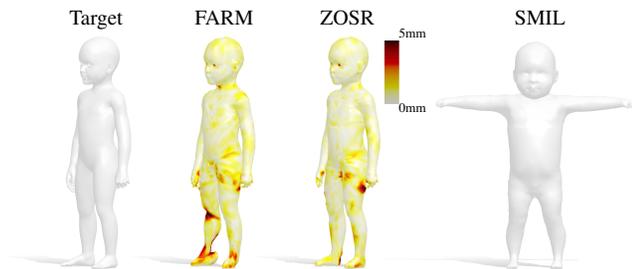
\end{center}
%
%
%

\vspace{-2.5cm}\noindent\textbf{Challenging cases.}
We also present some experiments on a few challenging cases. 
Firstly, we emphasize the positive effect of the HRA also in the case of low resolution shapes. In Figure~\ref{fig:LOW}, we consider a FAUST shape with only 6890 vertices and we apply our method together with the proposed HRA. Notice that using the proposed technique the result is geometry sharpening, and some traits become more evident. The High-resolution mesh coherently fit the original connectivity despite the large quantity of added vertices. Moreover the HRA does not cause error or collapsing.
Finally, Figure~\ref{fig:SMIL} is an original experiment: we test our method not only by changing the morphable model or domain (as shown in FARM with SCAPE and KIDS~\cite{KIDS}), but both together. We substitute the SMPL model with the SMIL morphable model~\cite{SMIL}, a parametric model similar to SMPL where the template respects more the proportions of infant shapes instead of adult ones. The proportions of the kid shape (on the left of Figure~\ref{fig:SMIL}) and SMIL template are different, thus also in this setting strong non-isometries are addressed. We correct the landmark detection heuristic in both FARM and ZOSR pipelines to perform correctly on different body proportions. All other settings are left unchanged. As can be seen, FARM fails dramatically on the right leg; it cannot be used without parameters redefinition. ZOSR performs robustly thanks to ZoomOut refinement. 
Furthermore, we prove that the proposed method can adopt different morphable models without additional effort.  
\section{Conclusions}
\label{sec:conclusions}
In this paper we presented a new approach for 3D shape matching of deformable human shapes. Our approach jointly exploits a spectral matching method, a parametric  model, and an extrinsic high-resolution refinement strategy. The proposed \emph{High-Resolution Augmentation}, in its global and in its innovative localized version, is able to fill the gap between the parametric model resolution and a general target geometry, also in the case of large mesh resolution differences. The quantitative evaluation shows that our approach outperforms the competitors on standard benchmarks. The HRA constitutes a promising solution to overcome the parametric models resolution limitations, giving rise to future directions in the high definition modeling. 

{\footnotesize
\section*{Acknowledgments} 
\vspace{-1.5ex}
\noindent
ER is supported by the ERC Starting Grant no. 802554 (SPECGEO). This work has been partially supported by project MIUR Excellence Departments 2018-2022.
}

{\small
\bibliographystyle{ieee}
\bibliography{egbib}
}
\clearpage
\newpage


\threedvfinalcopy 
\def\threedvPaperID{81} 
\def\httilde{\mbox{\tt\raisebox{-.5ex}{\symbol{126}}}}

\ifthreedvfinal\pagestyle{empty}\fi

\appendix
\renewcommand*{\thesection}{\arabic{section}}
\null
\begin{center}
      \begin{tabular}[!t]{c}
          \centering  
    {\Large \bf Supplementary Materials \par}
      \vspace*{24pt}
      \large
      \lineskip .5em
      \end{tabular}

      \par
      
      \vskip .5em
      \vspace*{12pt}
\end{center}

\textit{In this document we collect additional visualizations and results that were not included in the main paper due to lack of space.\\
We also provide \textbf{videos} for some of the presented registrations, that we find meaningful to understand our method performances. In the videos the error is shown as distance of target to the registration (i.e. white if there is no error, black if the error is $0.5$ centimeter or above). \vspace{0.5cm}\\
This document is divided in Sections, and in particular: \\
in Section \ref{sec:localized} we show some results of our localized adaptive refinement, with mesh detail.\\ 
In Section \ref{sec:HRA} we provide results of High-Resolution Augmentation, with front-back view.\\ 
In Section \ref{sec:ZOSR} there are massive registrations tests, with the error distance plotted over the registered template, with saturation at $1$ centimeter. \\
In Section \ref{sec:SMIL} we show cases of KIDS registration using ZOSR. \\
Finally in Section \ref{sec:TextureTransfer} we show our capability in texture transfer task.\vspace{0.5cm}\\
To avoid repetition over the entire document, the 3D morphable models used as tempaltes are SMPL \cite{SMPL} for adults and SMIL \cite{SMIL} for kids. The datasets used in our experiments are: SCAPE \cite{Anguelov05},
     TOSCA \cite{TOSCA},
     SPRING \cite{SPRING},
    MoSh \cite{MOSHMOCAP},
    CAESAR \cite{caesar} in the MPI free available version \cite{CAESAR-PIS},
   Princeton \cite{PRINCETON},
    SHREC14 \cite{SHREC2014},
    K3D-hub \cite{K3DHUB},
   Artist made models from BadKing.com.au website,
    KIDS \cite{KIDS}.}
\clearpage \newpage

 \section{Localized adaptive refinement}
\label{sec:localized}
Some examples of our localized adaptive refinement strategy.
In general we can achieve almost the same quality with half of the HR vertices. Since our strategy aims to refine local regions it is reasonable that some global surface behaviors has been bartered for high-frequencies in critical places (e.g. face). A combination of global and local approach is left for future works.
 \begin{center}
  \begin{figure}
  \vspace{0.5cm}
  \begin{overpic}
  [trim=0cm 0cm 0cm 0cm,clip,width=1\linewidth]{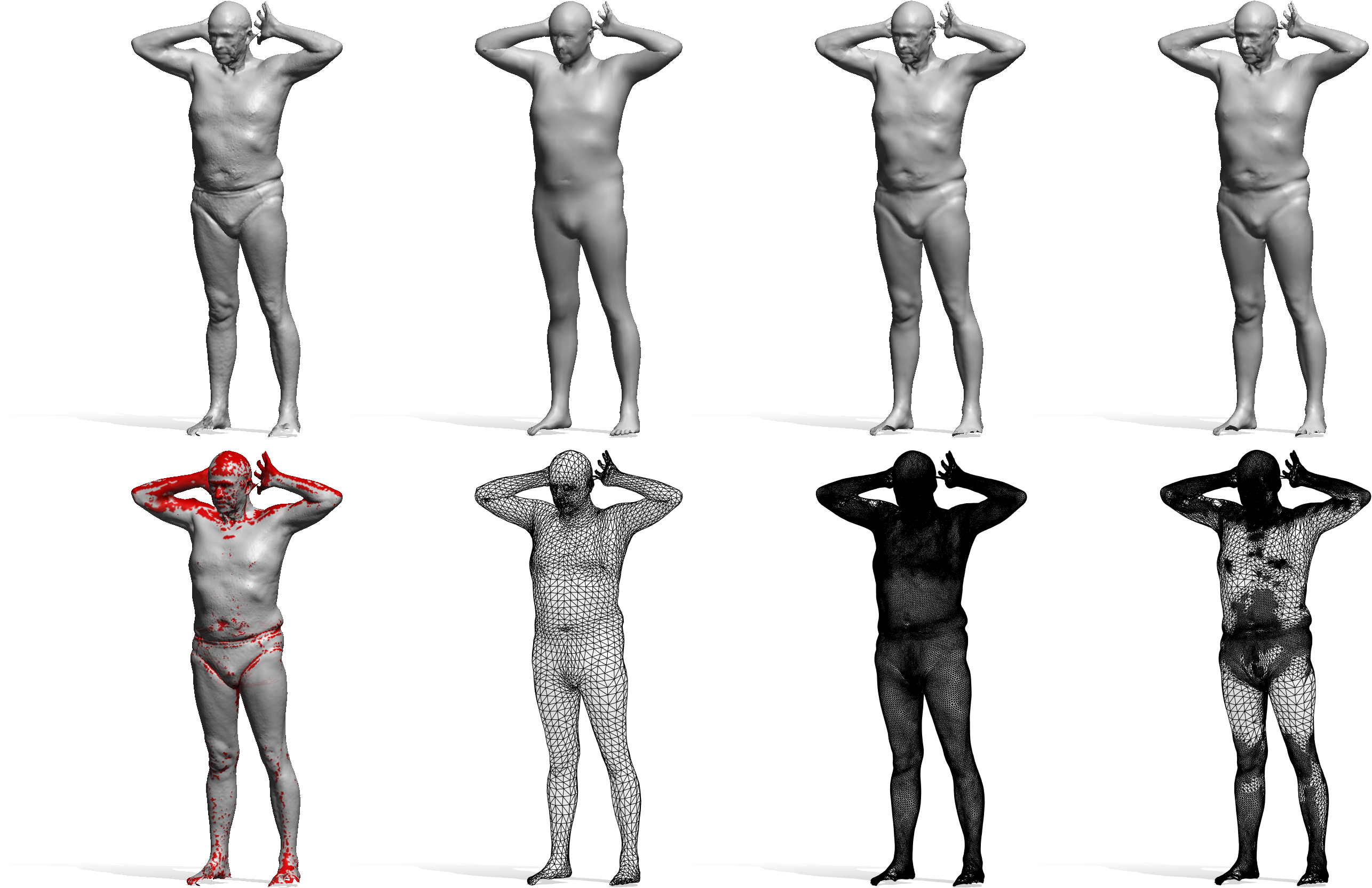}
  \put(15,-3){\footnotesize{205K}}
  \put(39.5,-3){\footnotesize{6.8K}}
  \put(65,-3){\footnotesize{110K}}
  \put(91,-3){\footnotesize{69K}}
  
  \put(13,67){\footnotesize{Target}}
  \put(38,67){\footnotesize{ZOSR}}
  \put(64,67){\footnotesize{HR}}
  \put(87,67){\footnotesize{Local}}
  \end{overpic}
  \vspace{0.2cm}
  \caption{}
    \label{fig:local}
\end{figure}
\end{center}

 \begin{center}
  \begin{figure}
  \vspace{0.5cm}
  \begin{overpic}
  [trim=0cm 0cm 0cm 0cm,clip,width=1\linewidth]{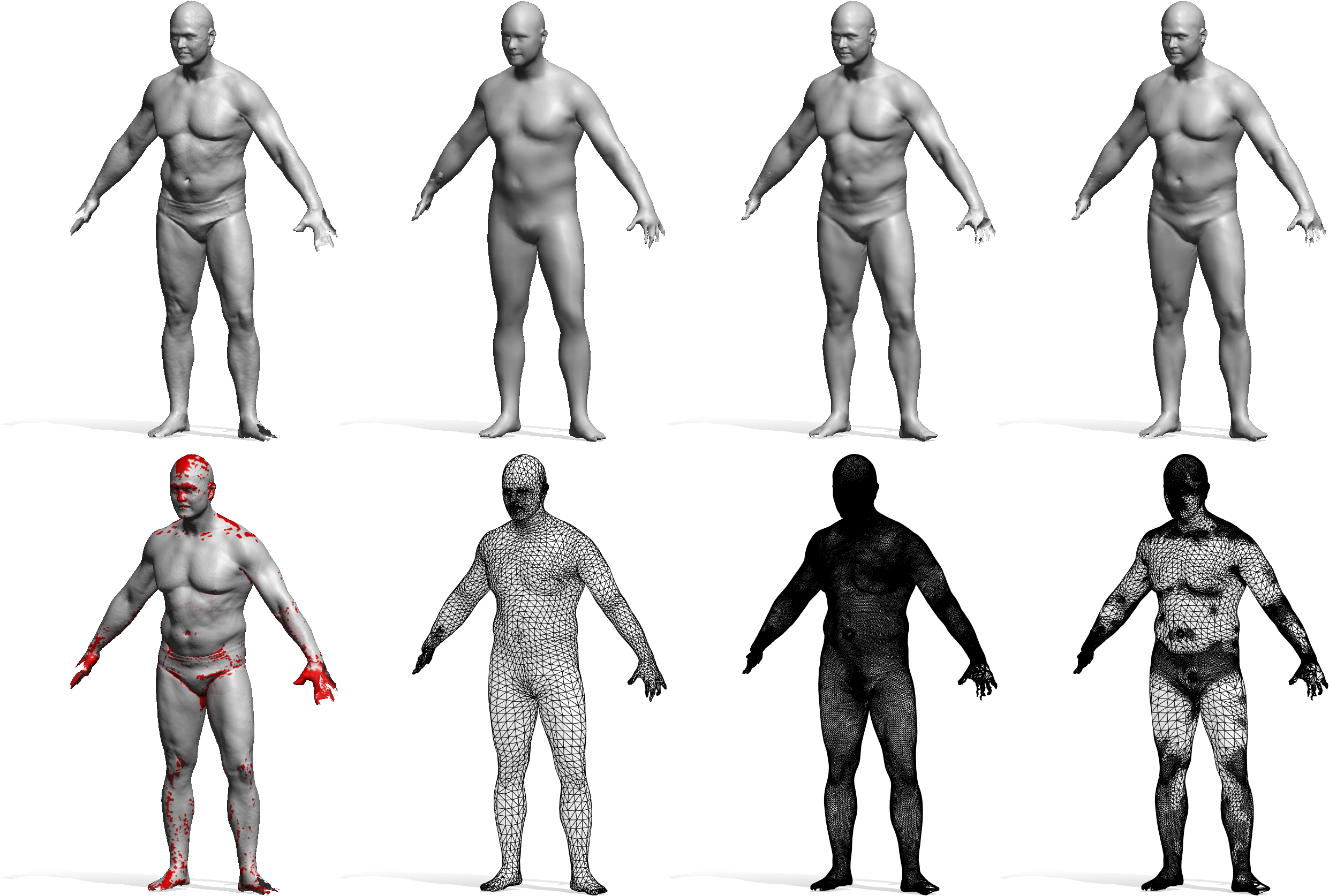}
  \put(13,-3){\footnotesize{213K}}
  \put(37.5,-3){\footnotesize{6.8K}}
  \put(63,-3){\footnotesize{110K}}
  \put(89,-3){\footnotesize{59K}}
  
  \put(11,70){\footnotesize{Target}}
  \put(36,70){\footnotesize{ZOSR}}
  \put(62.5,70){\footnotesize{HR}}
  \put(86,70){\footnotesize{Local}}
  \end{overpic}
  \vspace{0.2cm}
  \caption{}
    \label{fig:local}
\end{figure}
\end{center}

  \vspace{3cm}
 \begin{center}
  \begin{figure}
  \vspace{0.5cm}
  \begin{overpic}
  [trim=0cm 0cm 0cm 0cm,clip,width=0.8\linewidth]{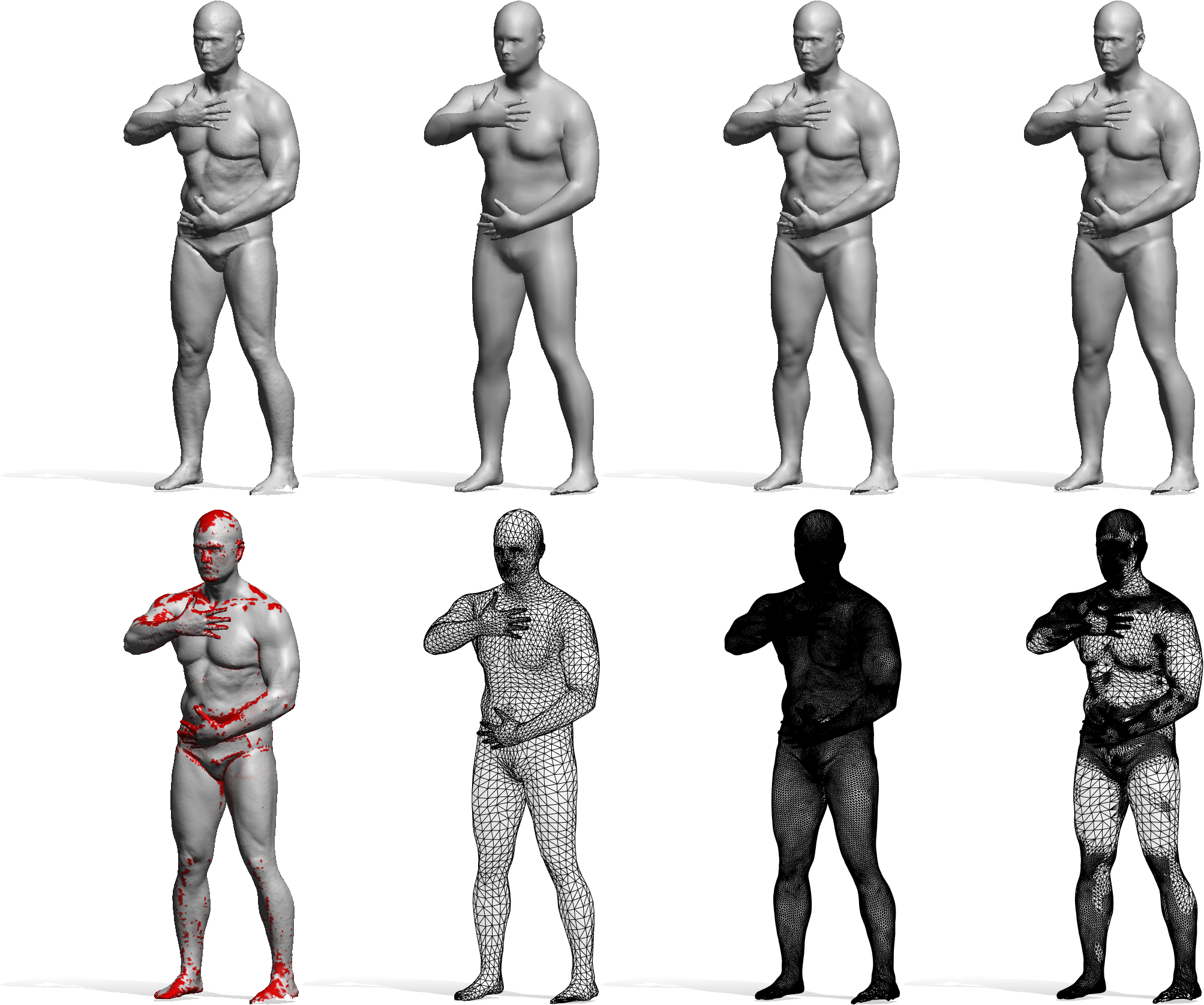}
  \put(15,-3){\footnotesize{204K}}
  \put(39.5,-3){\footnotesize{6.8K}}
  \put(65,-3){\footnotesize{110K}}
  \put(91,-3){\footnotesize{66K}}
  
  \put(14,85){\footnotesize{Target}}
  \put(38,85){\footnotesize{ZOSR}}
  \put(65,85){\footnotesize{HR}}
  \put(88.5,85){\footnotesize{Local}}
  \end{overpic}
  \vspace{0.2cm}
  \caption{}
    \label{fig:local}
\end{figure}
\end{center}

\clearpage \newpage

\section{High-Resolution Augmentation}
\label{sec:HRA}
In this section we collect experiments on some highly detailed meshes. All these meshes have a resolution heavily higher then the SMPL template. We tested our capability over a wide set of different type of details, obtaining always compelling quality.
They are all registered with three subdivision steps (i.e. the final template has 440K vertices).
We show back and front view for each experiment, compared also to the ZOSR results. 

 \begin{center}
  \begin{figure}
  \vspace{0.5cm}
  \begin{overpic}
  [trim=0cm 0cm 0cm 0cm,clip,width=1.8\linewidth]{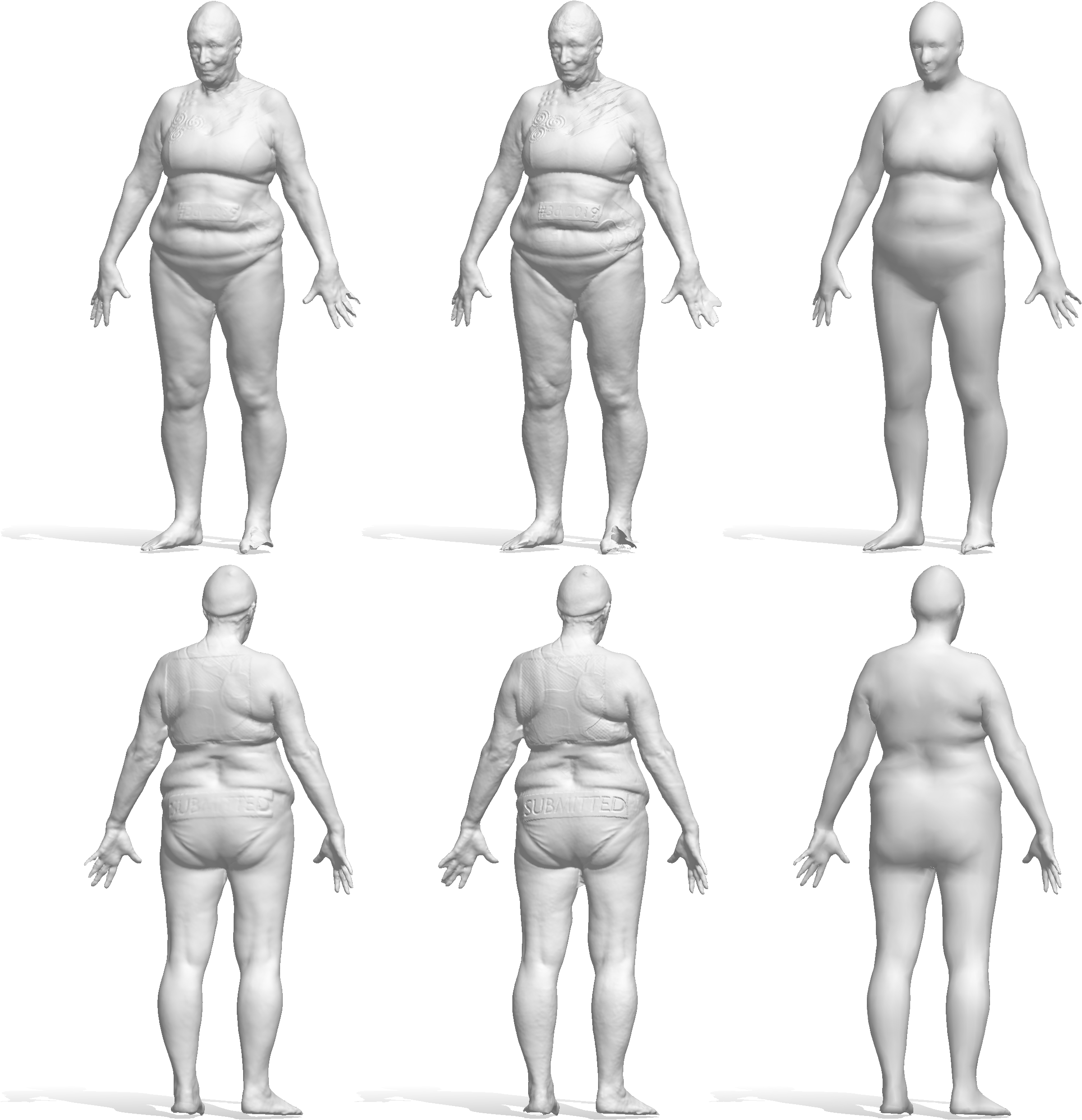}
  
  \put(18,100){\footnotesize{HR}}
  \put(49,100){\footnotesize{Target}}
  \put(81,100){\footnotesize{ZOSR}}
  \end{overpic}
  \vspace{0.05cm}
      \caption{HRA artist adorned - 427K vertices}
    \label{fig:HRA1}
\end{figure}
\end{center}

 \begin{center}
  \begin{figure*}
  \vspace{0.5cm}
  \begin{overpic}
  [trim=0cm 0cm 0cm 0cm,clip,width=\linewidth]{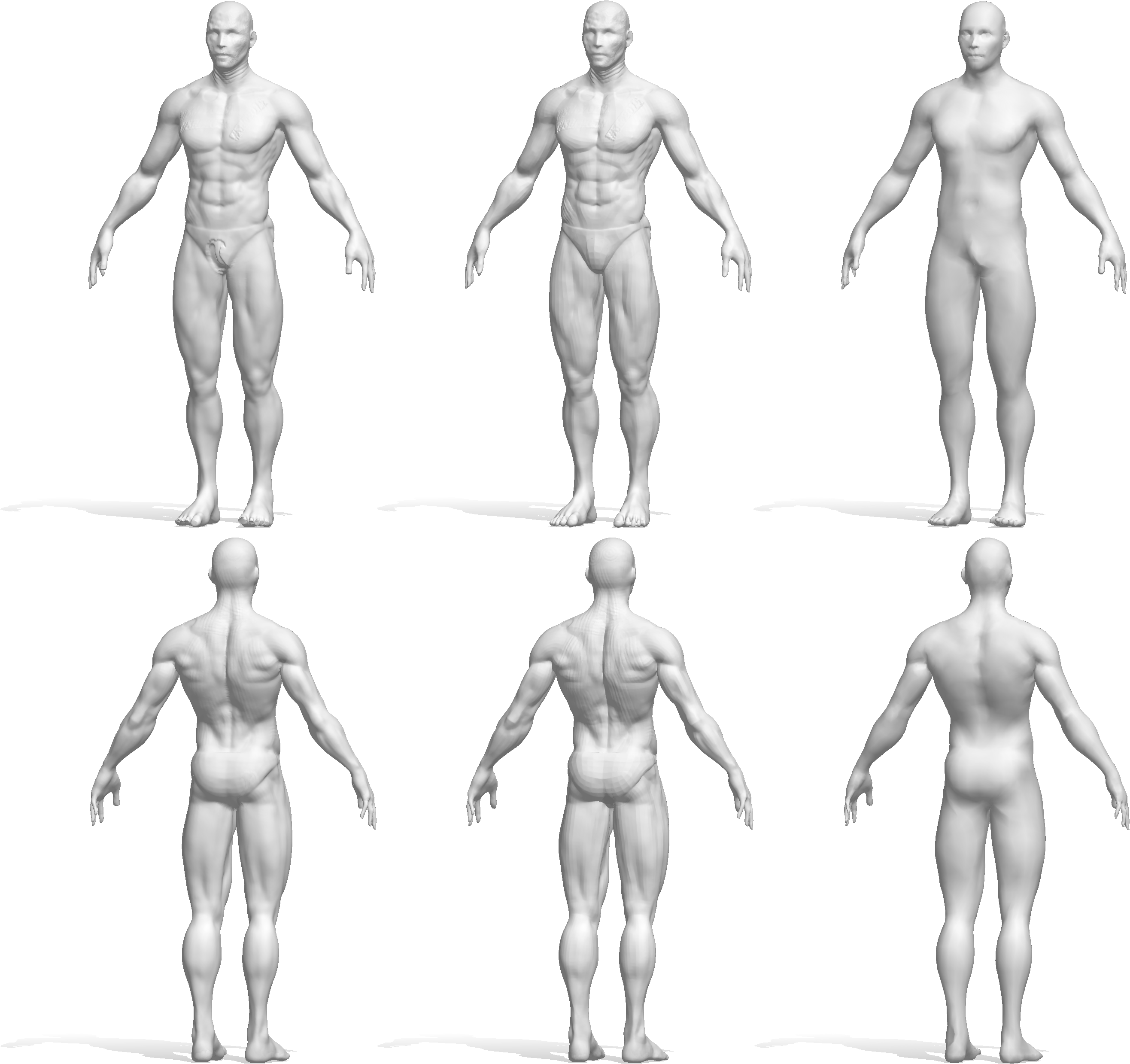}
  
  \put(19,97){\footnotesize{HR}}
  \put(51.5,97){\footnotesize{Target}}
  \put(85,97){\footnotesize{ZOSR}}
  \end{overpic}
  \vspace{0.05cm}
      \caption{HRA - 339K vertices}
    \label{fig:HRA3}
\end{figure*}
\end{center}

 \begin{center}
  \begin{figure*}[t!]
  \vspace{0.5cm}
  \begin{overpic}
  [trim=0cm 0cm 0cm 0cm,clip,width=\linewidth]{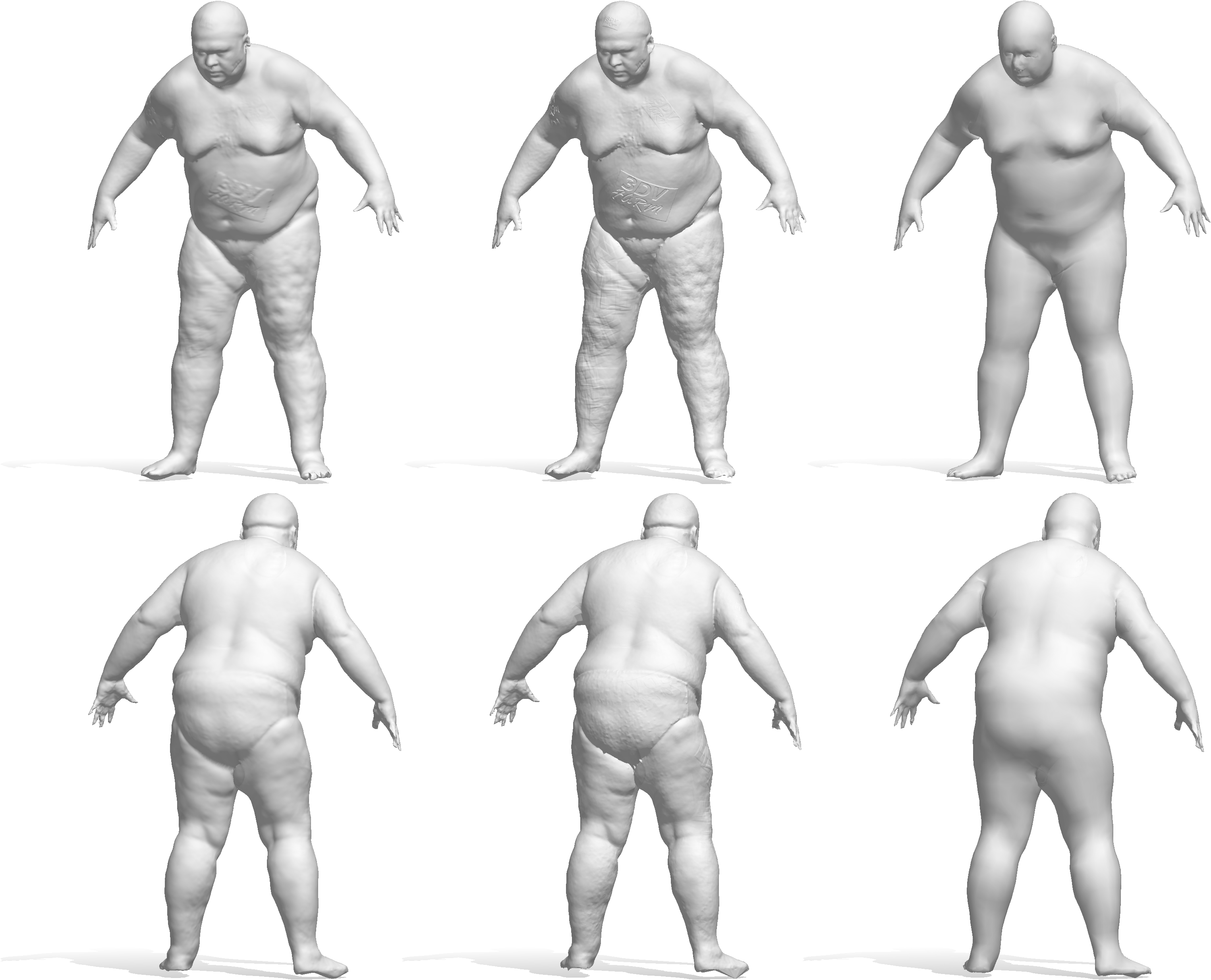}
  
  \put(18,83){\footnotesize{HR}}
  \put(50,83){\footnotesize{Target}}
  \put(84,83){\footnotesize{ZOSR}}
  \end{overpic}
  \vspace{0.05cm}
      \caption{HRA artist adorned - 2,1M vertices}
    \label{fig:HRA2}
\end{figure*}
\end{center}

 \begin{center}
  \begin{figure*}[t!]
  \vspace{0.5cm}
  \begin{overpic}
  [trim=0cm 0cm 0cm 0cm,clip,width=0.9\linewidth]{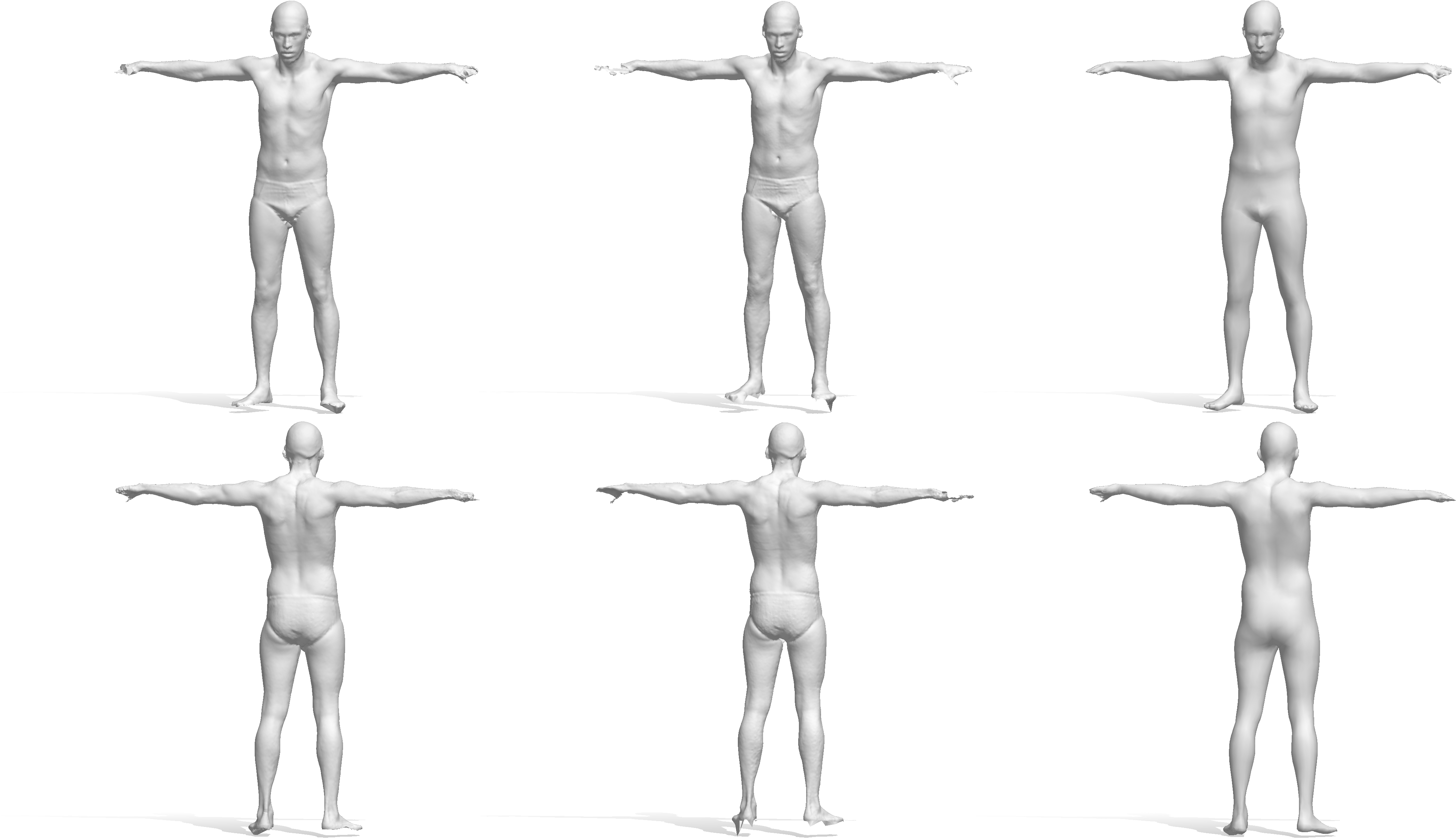}
  
  \put(18.5,59){\footnotesize{HR}}
  \put(51.5,59){\footnotesize{Target}}
  \put(84.5,59){\footnotesize{ZOSR}}
  \end{overpic}
  \vspace{0.05cm}
      \caption{HRA - 118K vertices}
    \label{fig:HRA4}
\end{figure*}
\end{center}

 \begin{center}
  \begin{figure*}[t!]
  \vspace{0.5cm}
  \begin{overpic}
  [trim=0cm 0cm 0cm 0cm,clip,width=0.9\linewidth]{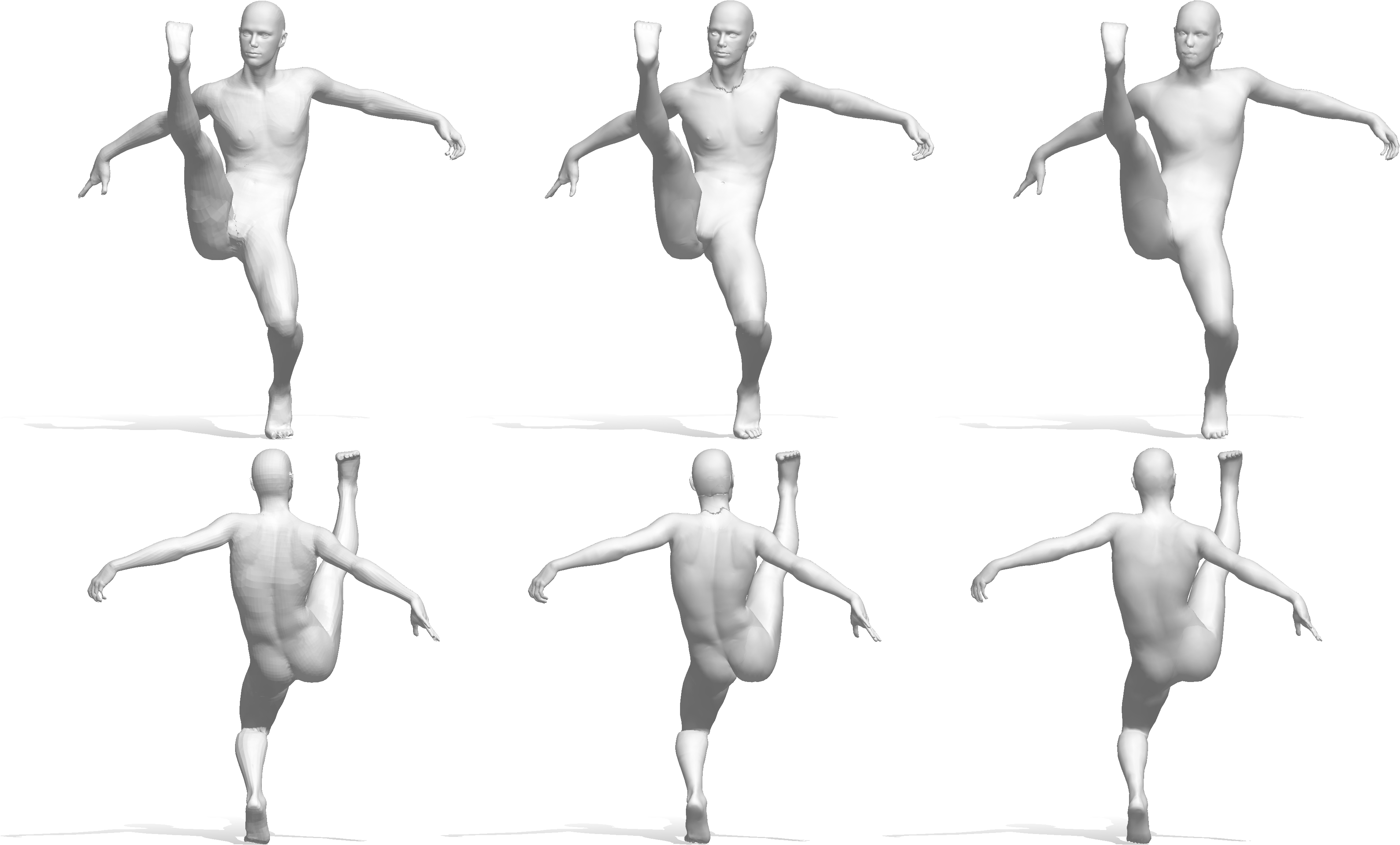}
  
  \put(18,62){\footnotesize{HR}}
  \put(50,62){\footnotesize{Target}}
  \put(84,62){\footnotesize{ZOSR}}
  \end{overpic}
  \vspace{0.05cm}
      \caption{HRA - 53K vertices}
    \label{fig:HRA5}
\end{figure*}
\end{center}

\clearpage \newpage

 \section{ZOSR}
\label{sec:ZOSR}
\begin{center}
  \begin{figure}[t!]
  \vspace{0.5cm}
  \begin{overpic}
  [trim=0cm 0cm 0cm 0cm,clip,width=\linewidth]{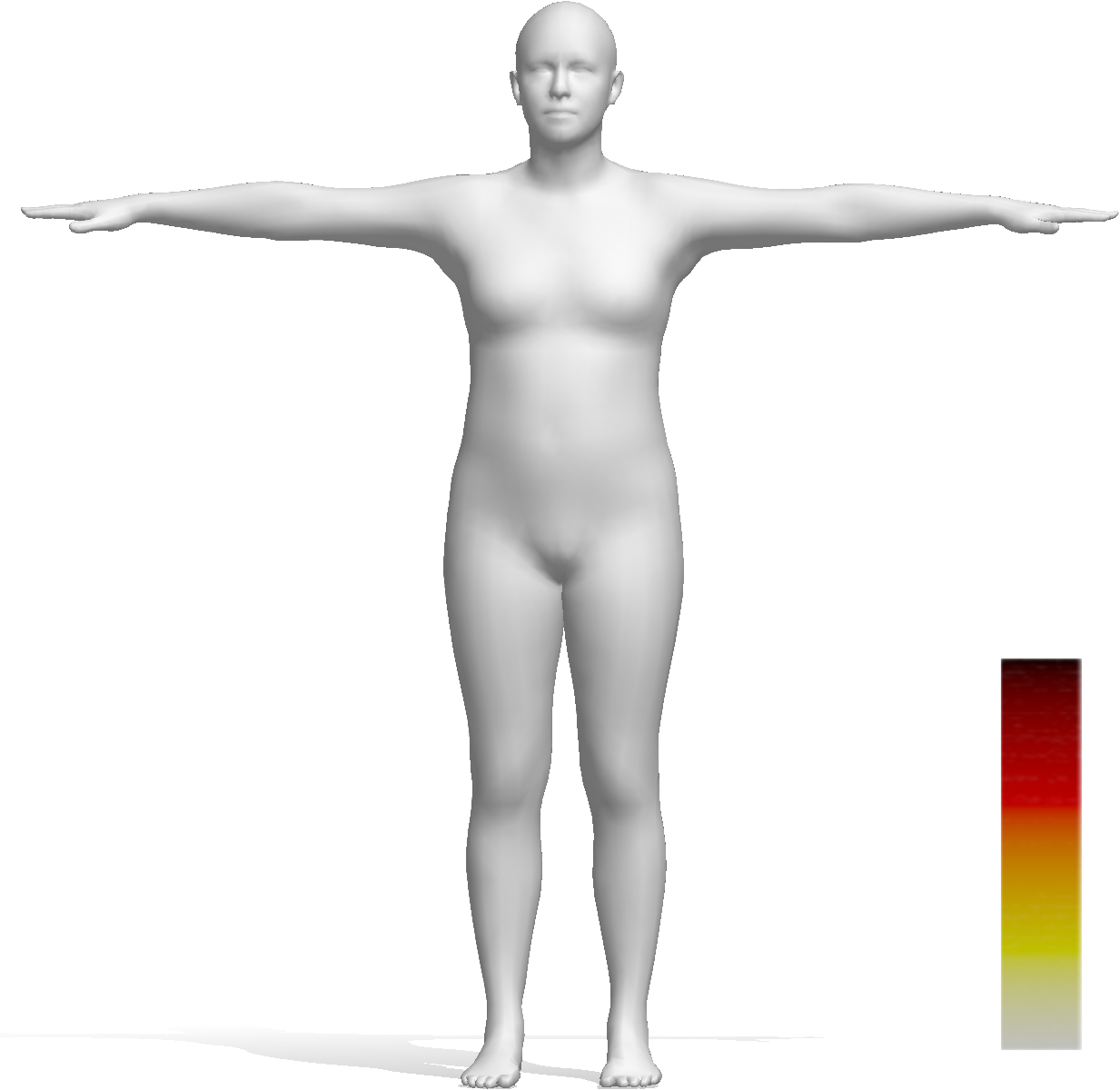}

\put(98,1){\footnotesize{0cm}}
\put(98,40){\footnotesize{1cm}}
  \end{overpic}
  \vspace{0.5cm}
  \caption{SMPL template}
      
    \label{fig:smpl}
\end{figure}
\end{center}

Here we show some some example of ZOSR registration without HR. This provide an idea about the variety of possible target shapes. Error in this case is plotted over the registered template with saturation at \textbf{1 centimeter}. The SMPL template has been shown in Figure \ref{fig:smpl}. It is the same used for the results in Section \ref{sec:HRA}.

 \begin{center}
  \begin{figure}[t!]
  \vspace{0.5cm}
  \begin{overpic}
  [trim=0cm 0cm 0cm 0cm,clip,width=\linewidth]{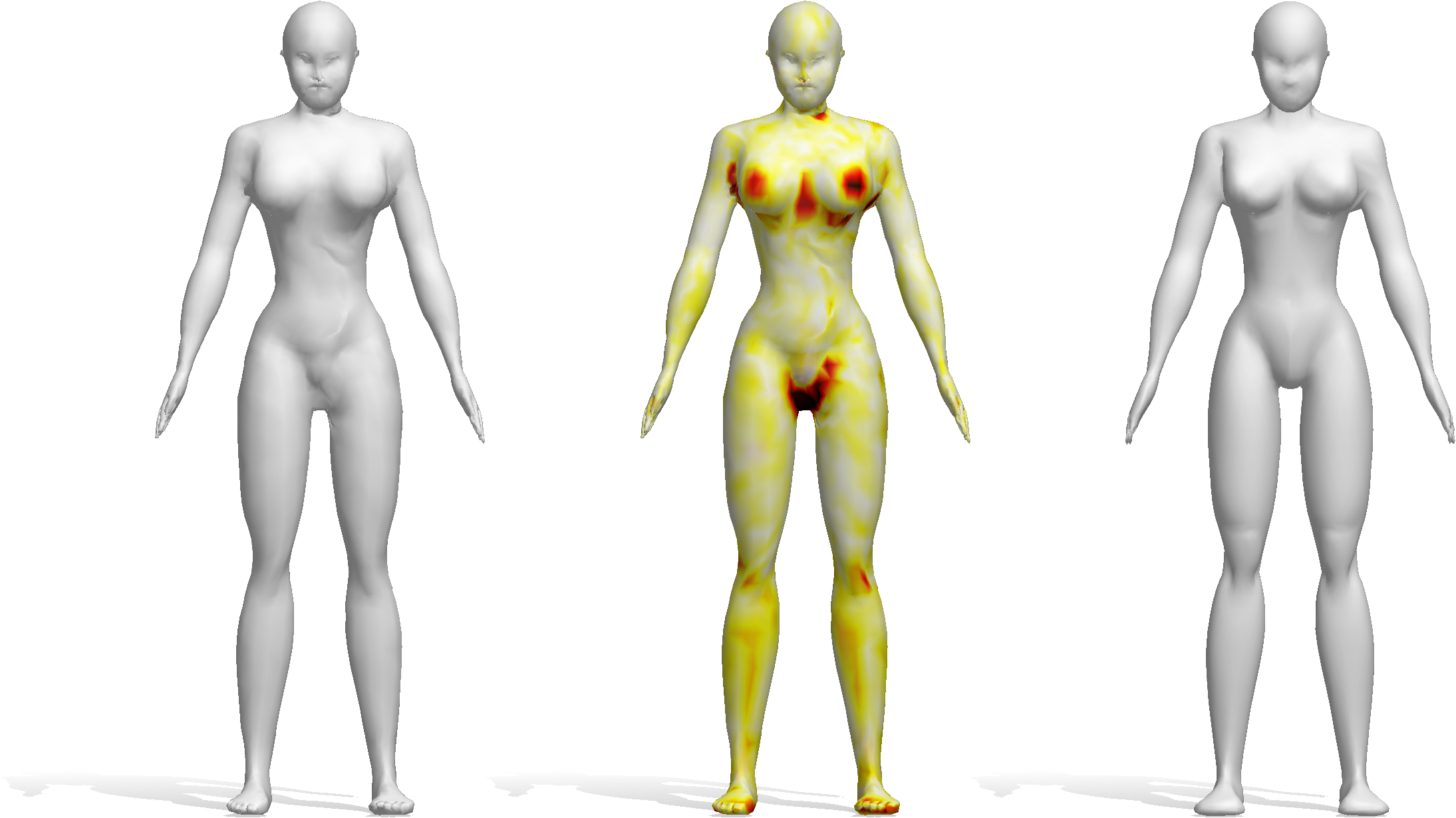}
  
  \put(18,58){\footnotesize{ZOSR}}
  \put(50,50){\footnotesize{}}
  \put(84,58){\footnotesize{Target}}
  \end{overpic}

    \label{fig:local}
\end{figure}

\end{center}
 \begin{center}
  \begin{figure}[t!]
  \vspace{0.5cm}
  \begin{overpic}
  [trim=0cm 0cm 0cm 0cm,clip,width=\linewidth]{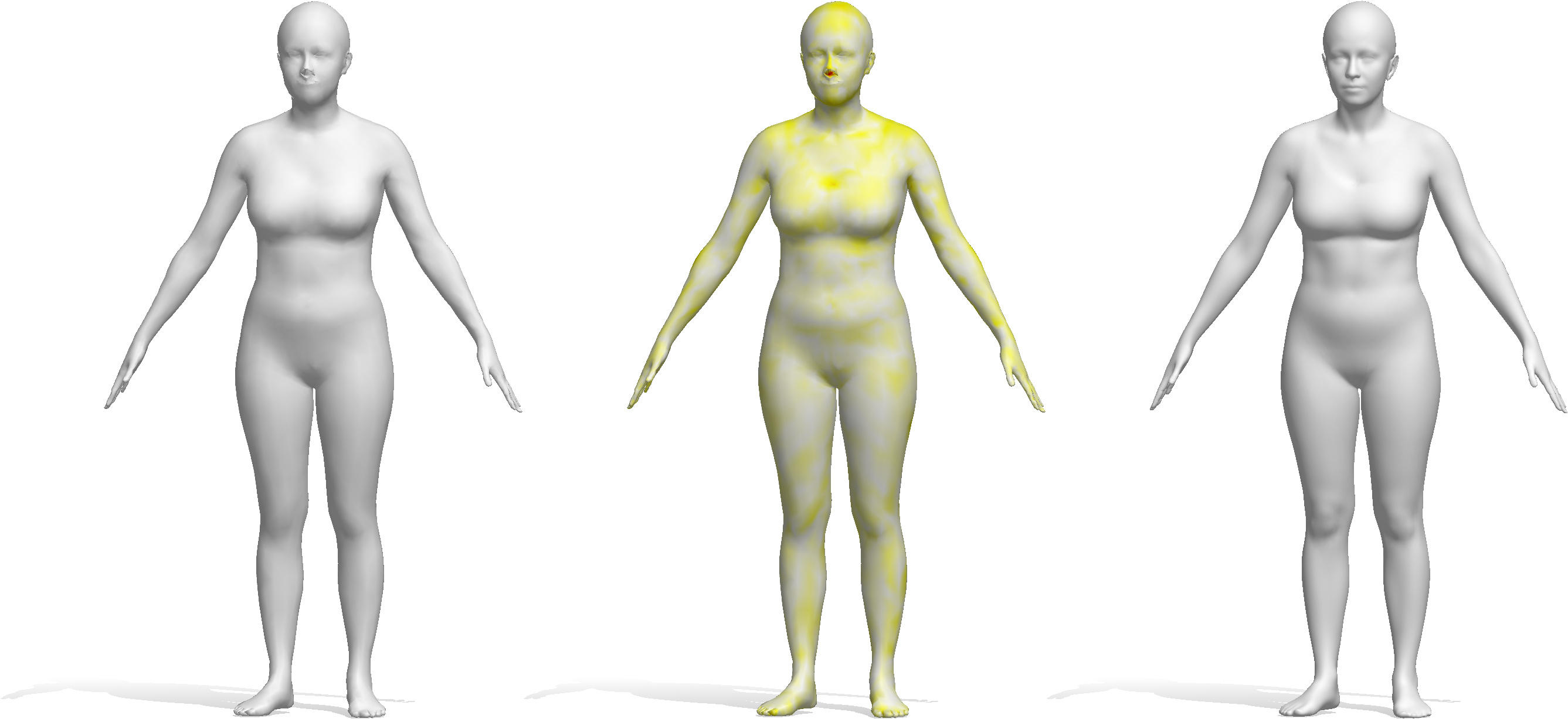}
  
  \put(17,48){\footnotesize{ZOSR}}
  \put(50,50){\footnotesize{}}
  \put(82,48){\footnotesize{Target}}
  \end{overpic}

    \label{fig:local}
\end{figure}
\end{center}

 \begin{center}
  \begin{figure}[t!]
  \vspace{0.5cm}
  \begin{overpic}
  [trim=0cm 0cm 0cm 0cm,clip,width=\linewidth]{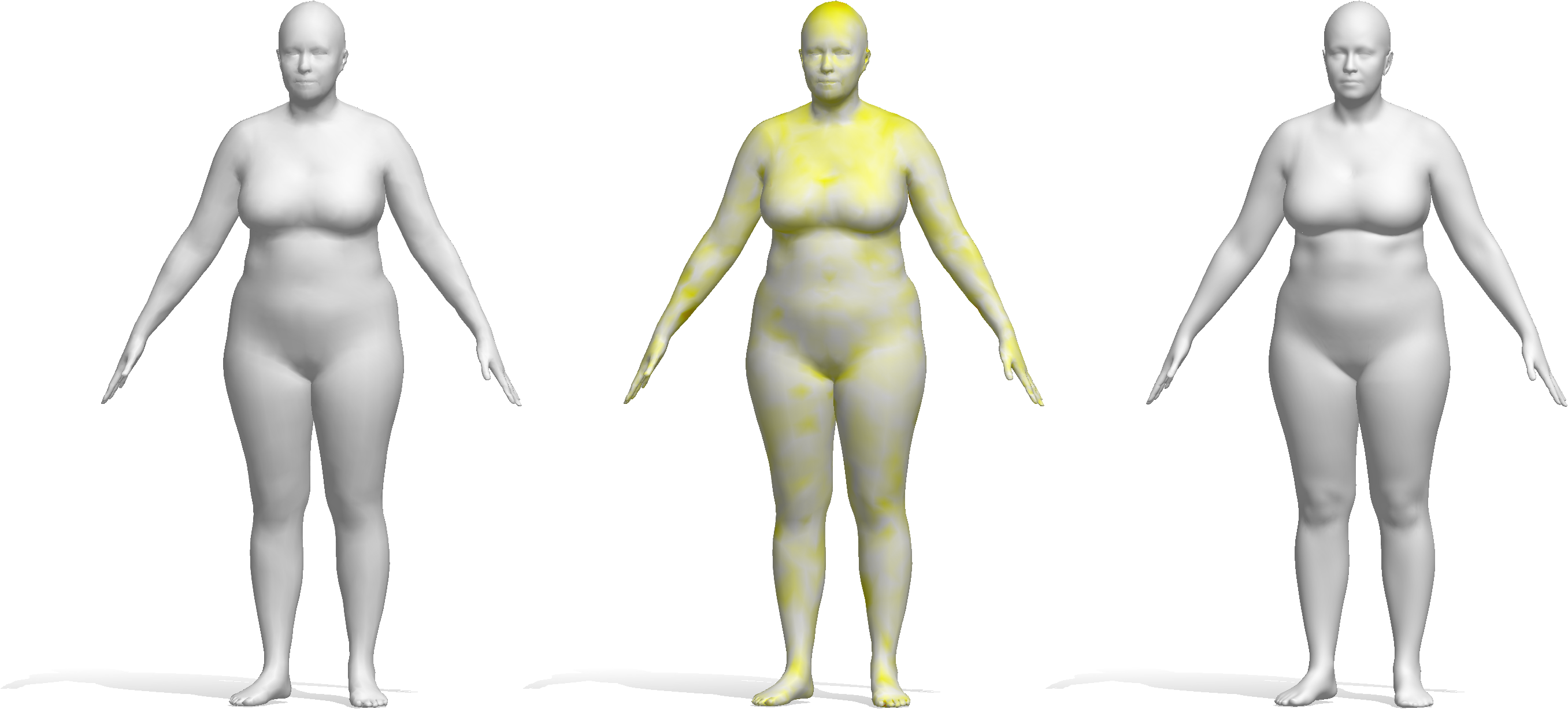}
  
  \put(16,48){\footnotesize{ZOSR}}
  \put(50,50){\footnotesize{}}
  \put(82.5,48){\footnotesize{Target}}
  \end{overpic}
    \label{fig:local}
\end{figure}
\end{center}

 \begin{center}
  \begin{figure}[t!]
  \vspace{0.5cm}
  \begin{overpic}
  [trim=0cm 0cm 0cm 0cm,clip,width=\linewidth]{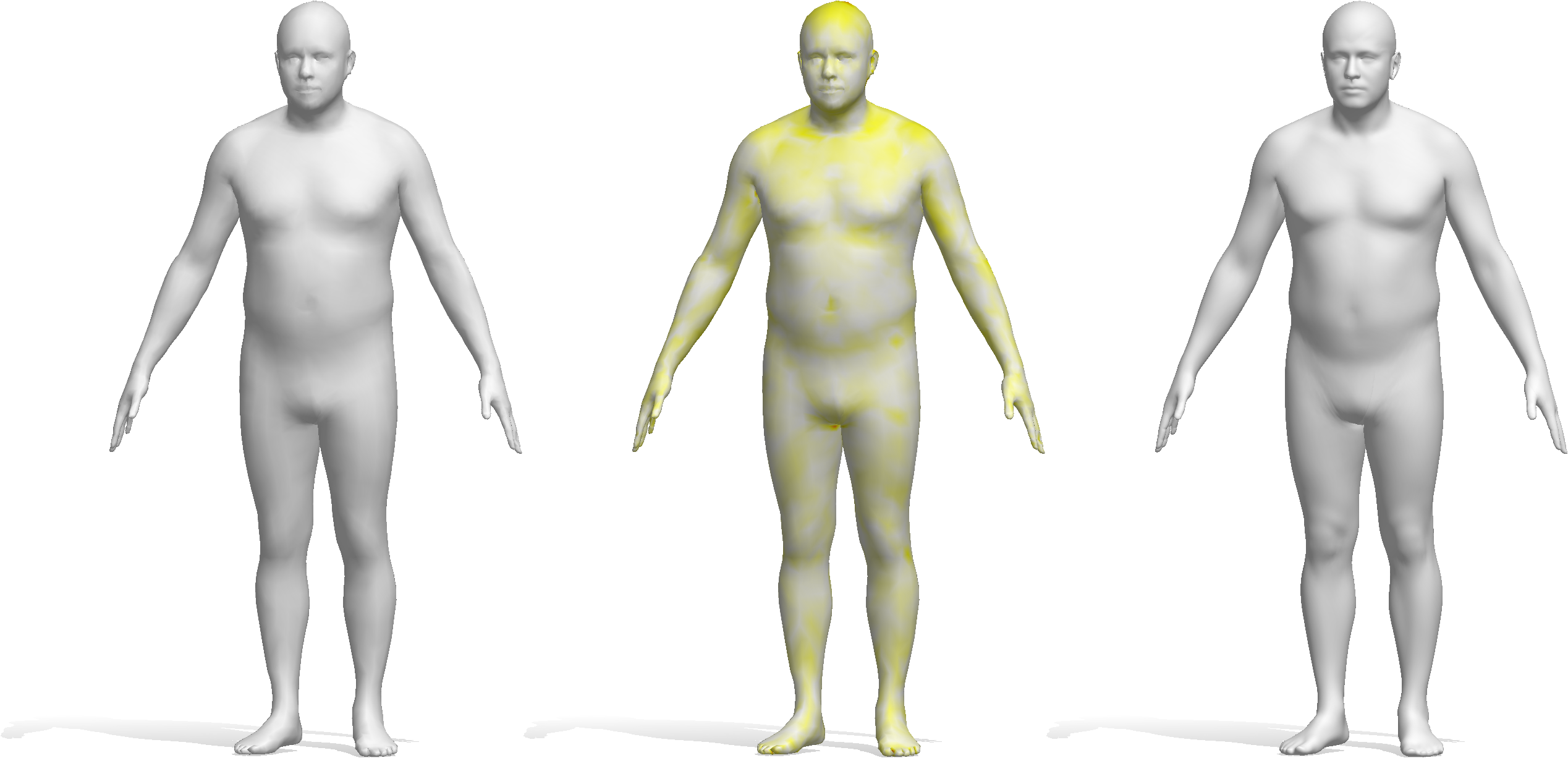}
  
  \put(17,50){\footnotesize{ZOSR}}
  \put(50,50){\footnotesize{}}
  \put(82,50){\footnotesize{Target}}
  \end{overpic}

    \label{fig:local}
\end{figure}
\end{center}

 \begin{center}
  \begin{figure}[t!]
  \vspace{0.5cm}
  \begin{overpic}
  [trim=0cm 0cm 0cm 0cm,clip,width=\linewidth]{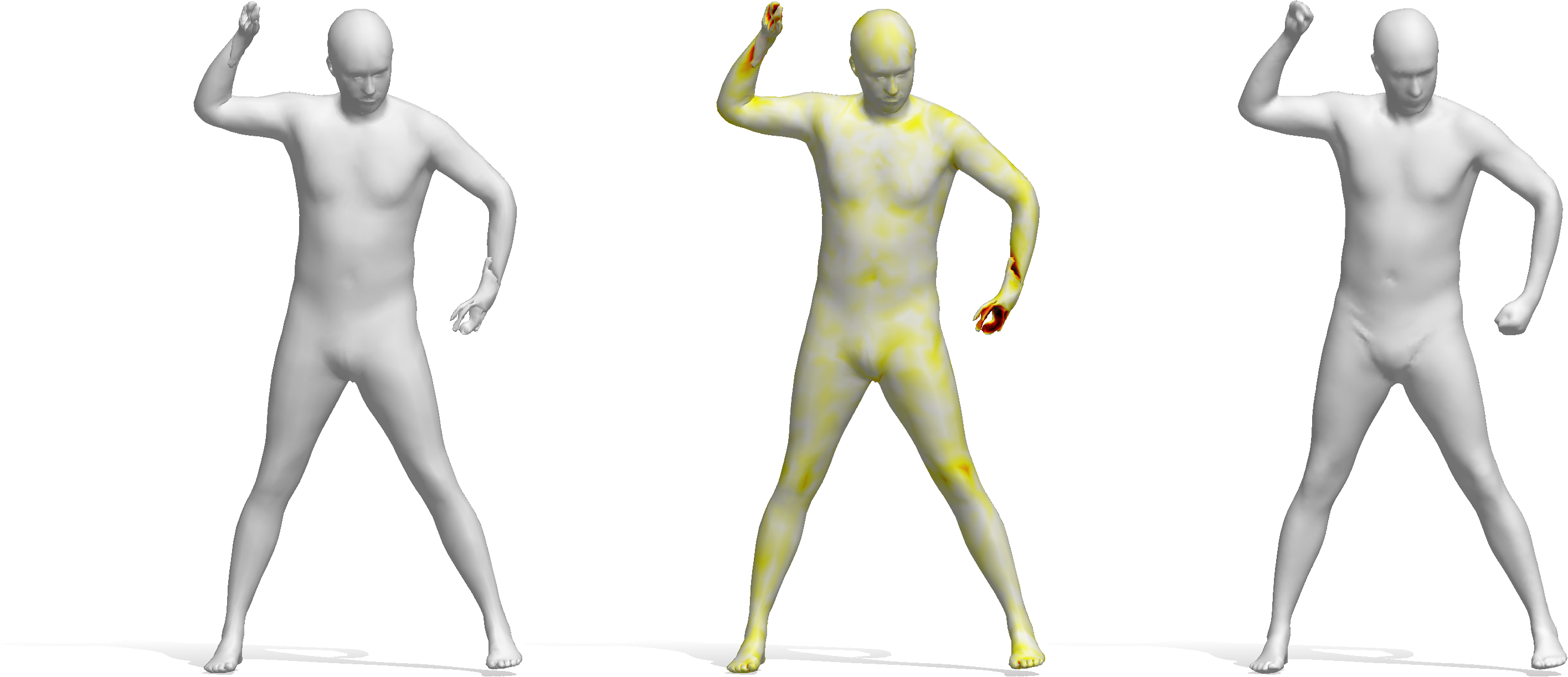}
  
  \put(18,46){\footnotesize{ZOSR}}
  \put(50,47){\footnotesize{}}
  \put(84,46){\footnotesize{Target}}
  \end{overpic}
    \label{fig:local}
\end{figure}
\end{center}

 \begin{center}
  \begin{figure}[t!]
  \vspace{0.5cm}
  \begin{overpic}
  [trim=0cm 0cm 0cm 0cm,clip,width=\linewidth]{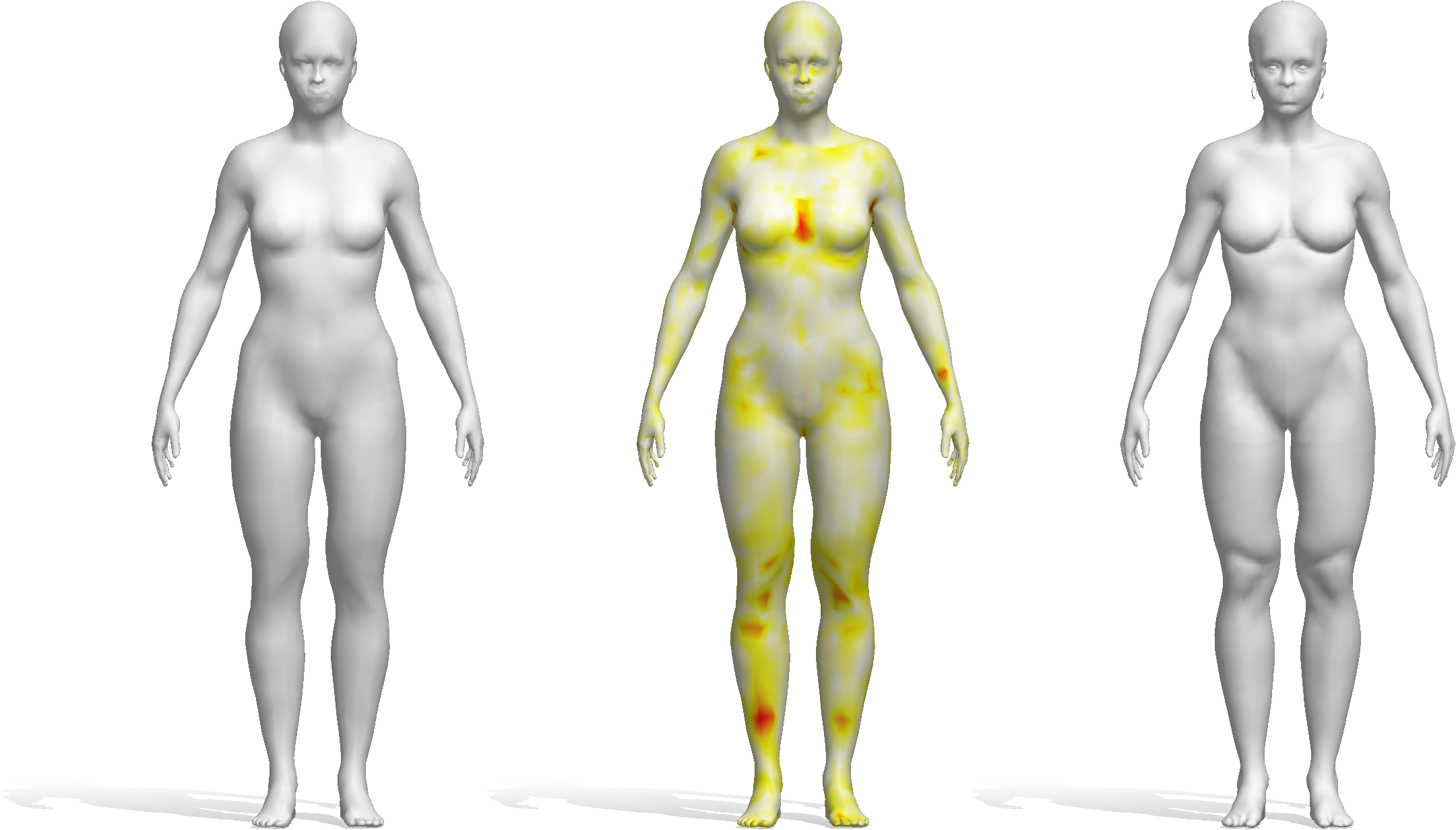}
  
  \put(18,58){\footnotesize{ZOSR}}
  \put(50,50){\footnotesize{}}
  \put(84,58){\footnotesize{Target}}
  \end{overpic}
    \label{fig:local}
\end{figure}
\end{center}


 \begin{center}
  \begin{figure}[t!]
  \vspace{0.5cm}
  \begin{overpic}
  [trim=0cm 0cm 0cm 0cm,clip,width=\linewidth]{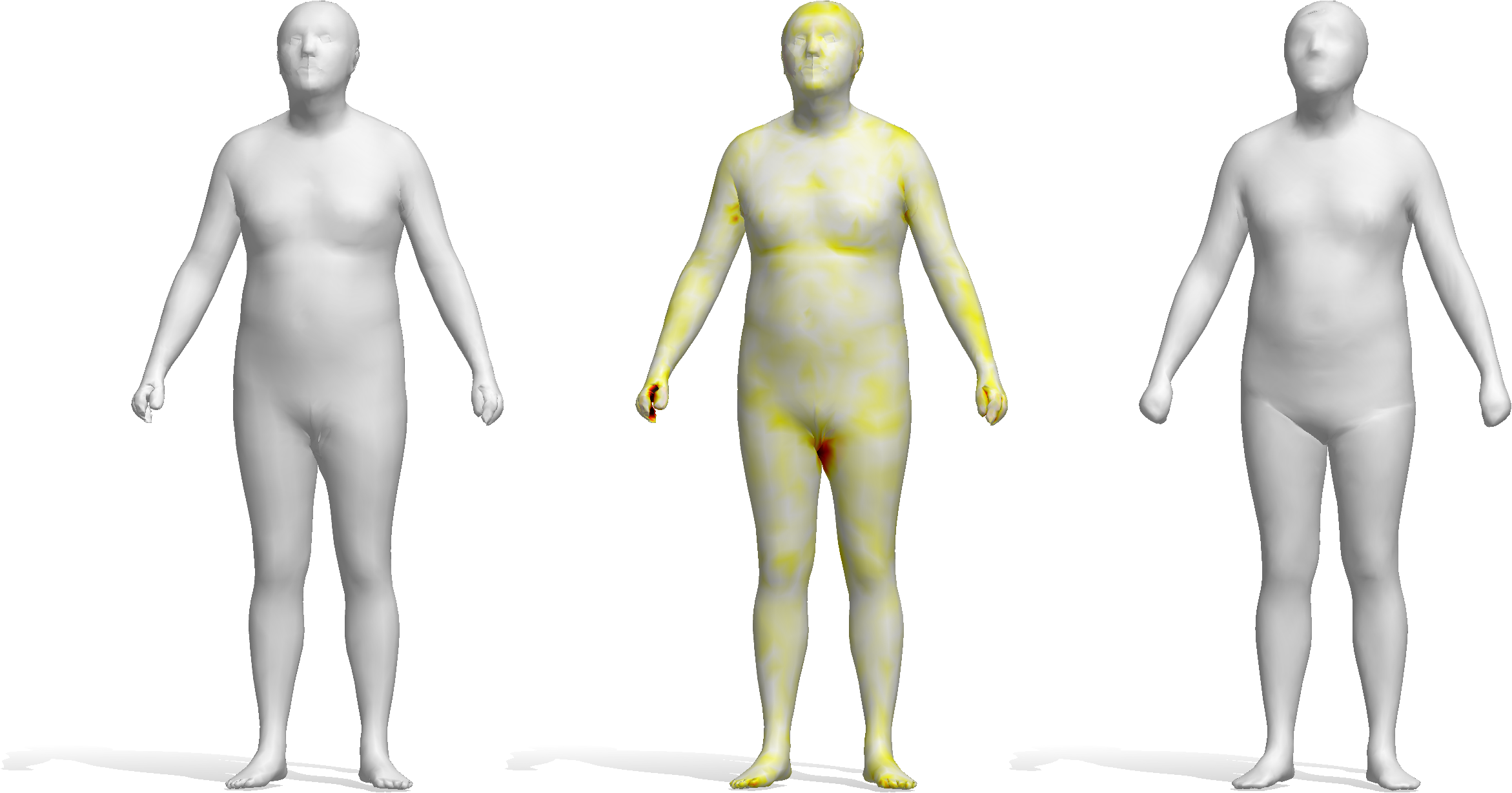}
  
  \put(17,53){\footnotesize{ZOSR}}
  \put(50,50){\footnotesize{}}
  \put(83,53){\footnotesize{Target}}
  \end{overpic}
    \label{fig:local}
\end{figure}
\end{center}

 \begin{center}
  \begin{figure}[t!]
  \vspace{0.5cm}
  \begin{overpic}
  [trim=0cm 0cm 0cm 0cm,clip,width=\linewidth]{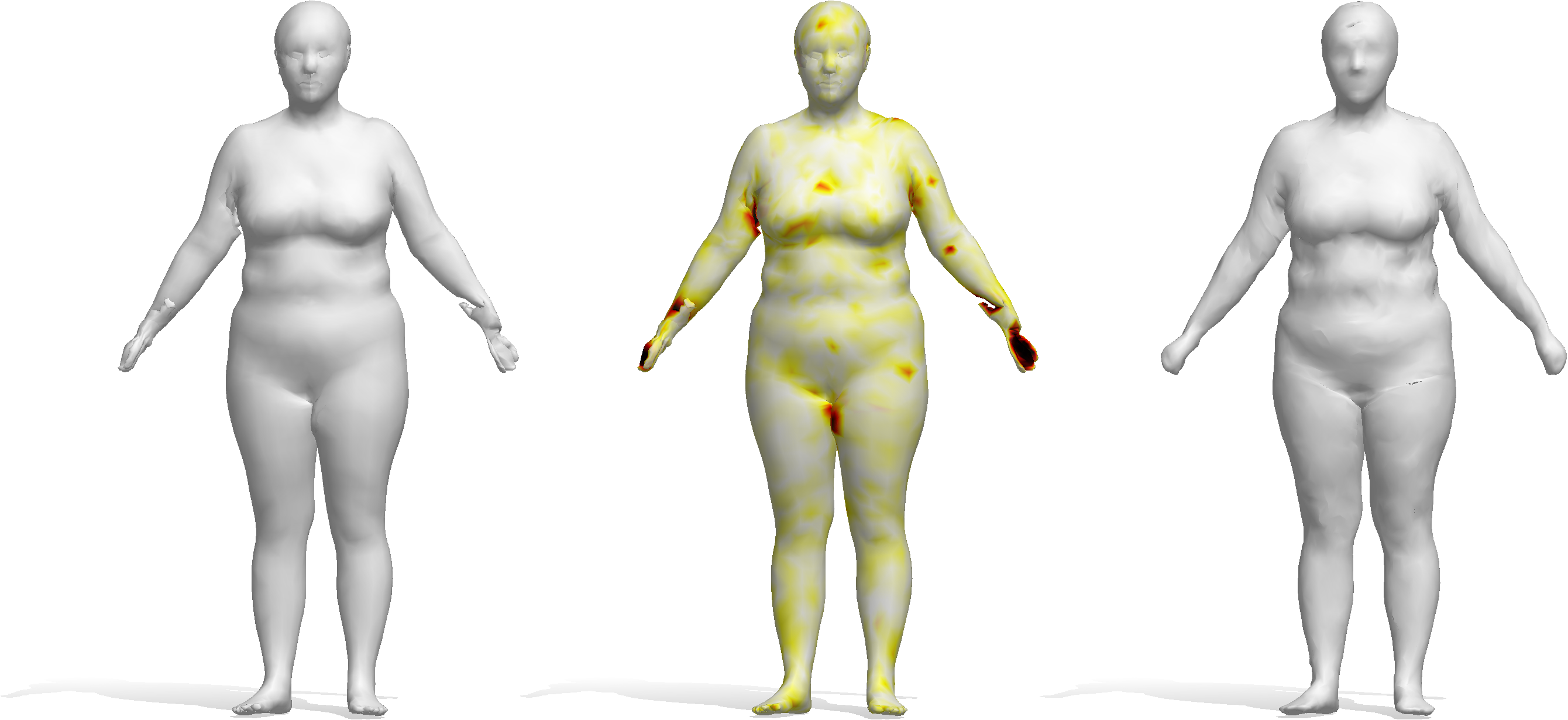}
  
  \put(17,48){\footnotesize{ZOSR}}
  \put(50,50){\footnotesize{}}
  \put(82,48){\footnotesize{Target}}
  \end{overpic}
    \label{fig:local}
\end{figure}
\end{center}

 \begin{center}
  \begin{figure}[t!]
  \vspace{0.5cm}
  \begin{overpic}
  [trim=0cm 0cm 0cm 0cm,clip,width=0.8\linewidth]{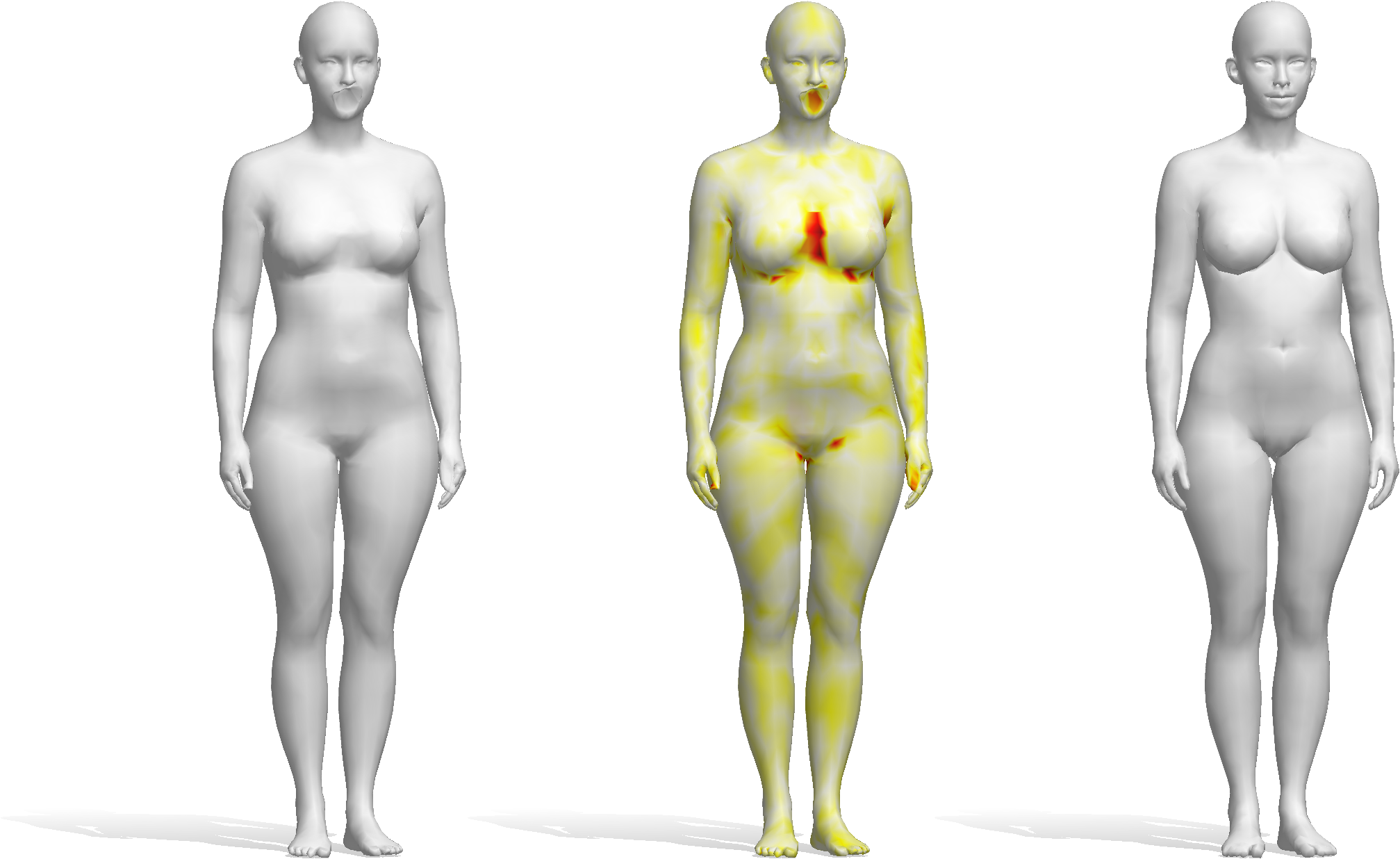}
  
  \put(18.5,64.5){\footnotesize{ZOSR}}
  \put(50,50){\footnotesize{}}
  \put(84,64.5){\footnotesize{Target}}
  \end{overpic}

    \label{fig:local}
\end{figure}
\end{center}

 \begin{center}
  \begin{figure*}[t!]
  \vspace{0.5cm}
  \begin{overpic}
  [trim=0cm 0cm 0cm 0cm,clip,width=\linewidth]{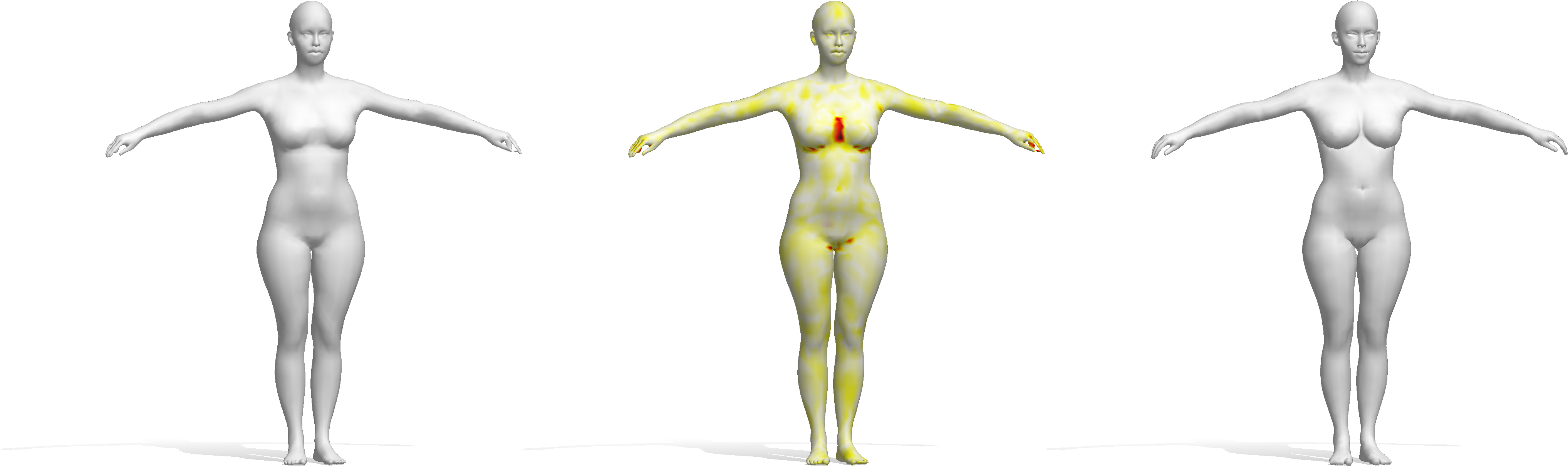}
  
  \put(18,32){\footnotesize{ZOSR}}
  \put(50,50){\footnotesize{}}
  \put(84.5,32){\footnotesize{Target}}
  \end{overpic}
    \label{fig:local}
\end{figure*}
\end{center}

 \begin{center}
  \begin{figure}[t!]
  \vspace{0.5cm}
  \begin{overpic}
  [trim=0cm 0cm 0cm 0cm,clip,width=\linewidth]{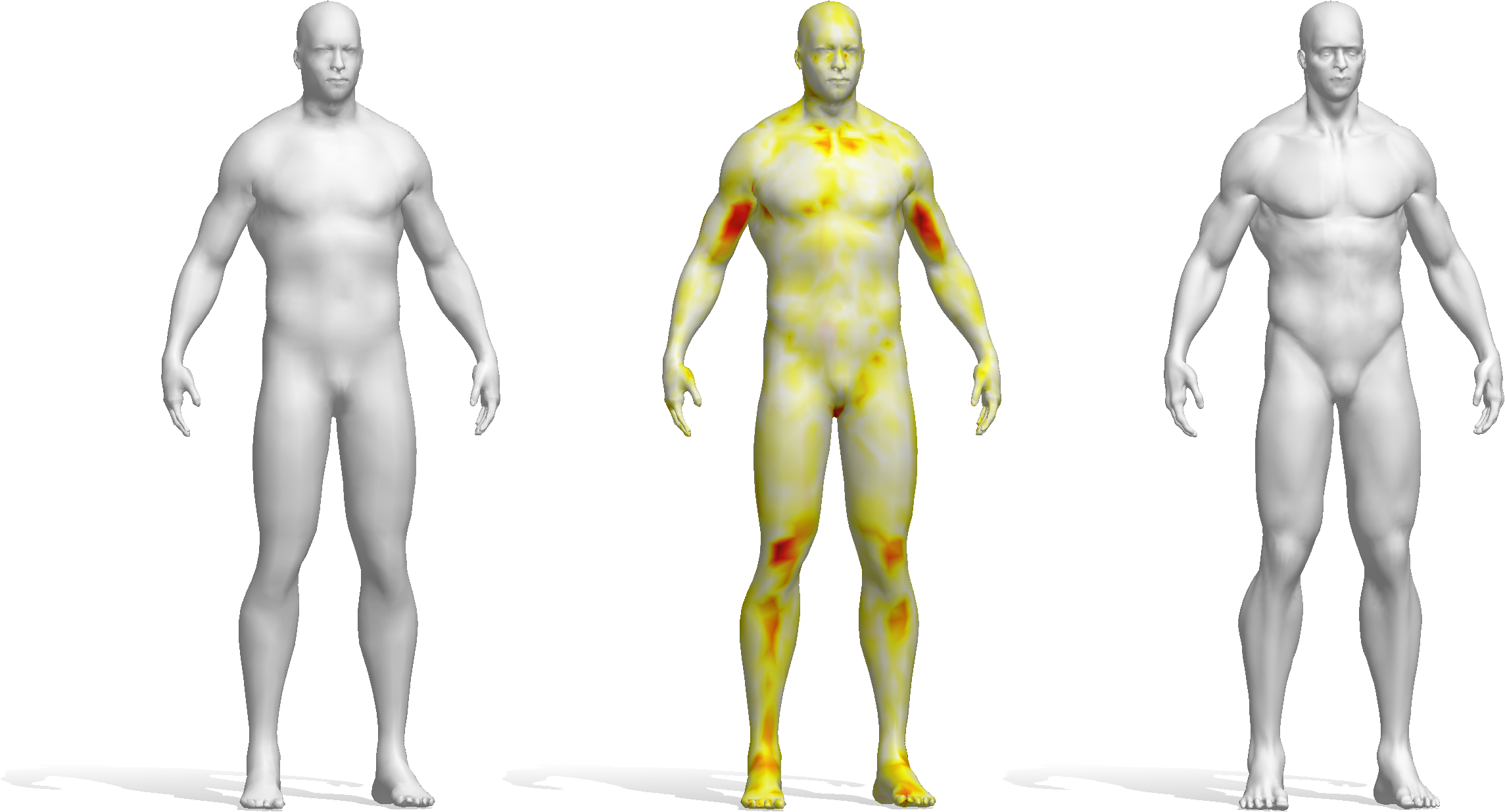}
  \put(18,55){\footnotesize{ZOSR}}
  \put(50,50){\footnotesize{}}
  \put(83.5,55){\footnotesize{Target}}
  \end{overpic}

    \label{fig:local}
\end{figure}
\end{center}

 \begin{center}
  \begin{figure}[t!]
  \vspace{0.5cm}
  \begin{overpic}
  [trim=0cm 0cm 0cm 0cm,clip,width=\linewidth]{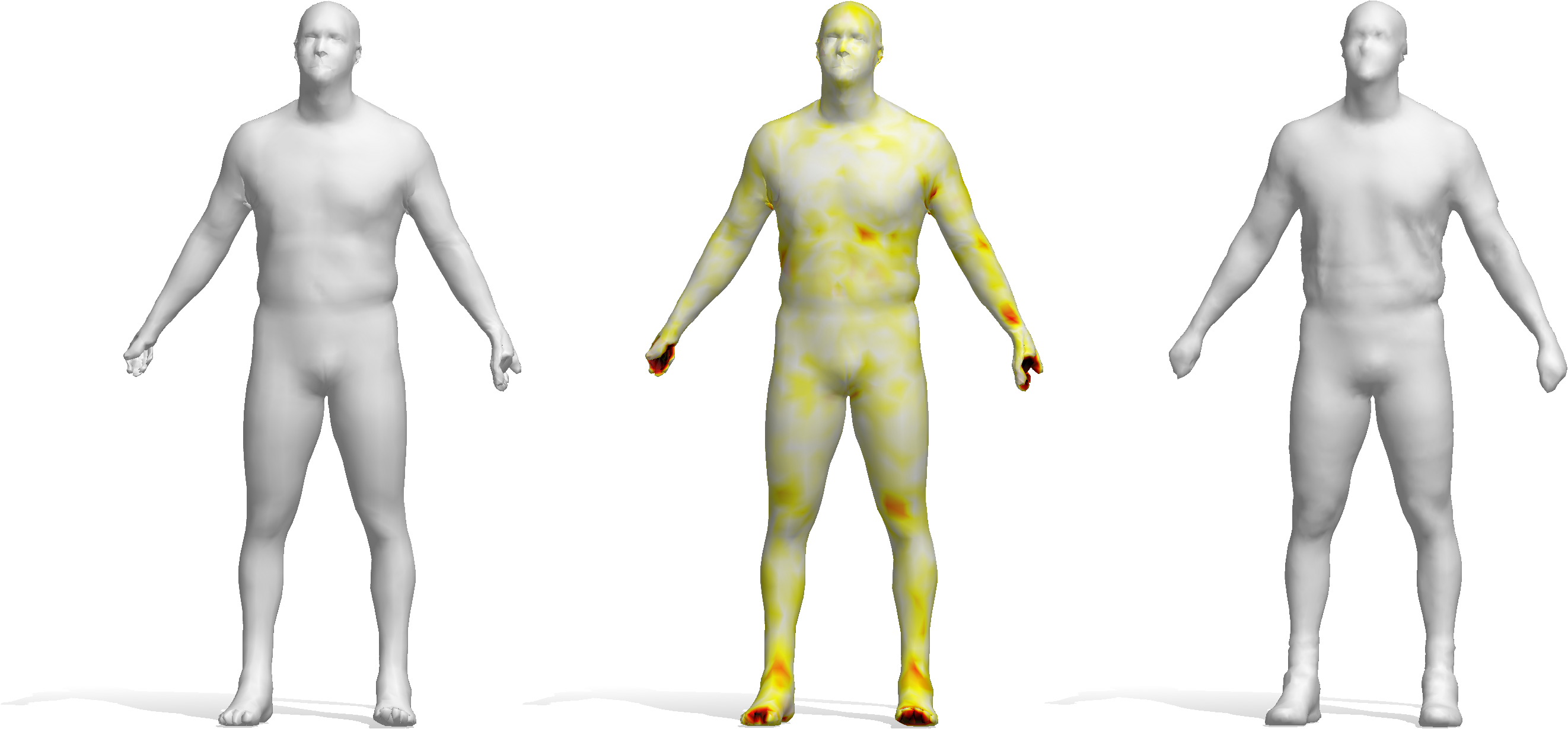}
  
  \put(18,49){\footnotesize{ZOSR}}
  \put(50,50){\footnotesize{}}
  \put(82.5,49){\footnotesize{Target}}
  \end{overpic}

    \label{fig:local}
\end{figure}
\end{center}

 \begin{center}
  \begin{figure}[t!]
  \vspace{0.5cm}
  \begin{overpic}
  [trim=0cm 0cm 0cm 0cm,clip,width=\linewidth]{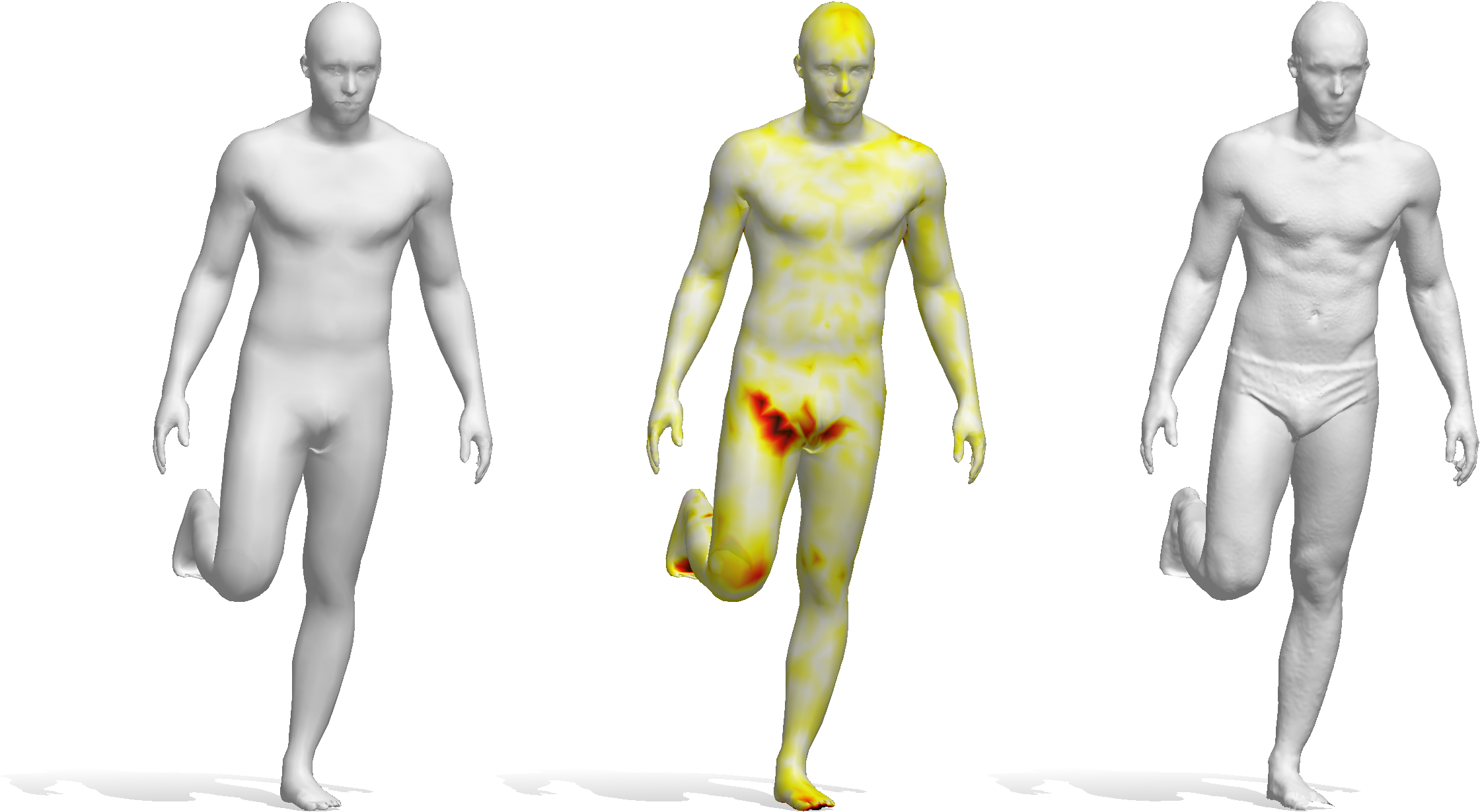}
  
  \put(19,56){\footnotesize{ZOSR}}
  \put(50,50){\footnotesize{}}
  \put(85,56){\footnotesize{Target}}
  \end{overpic}

    \label{fig:local}
\end{figure}
\end{center}

 \begin{center}
  \begin{figure}[t!]
  \vspace{0.5cm}
  \begin{overpic}
  [trim=0cm 0cm 0cm 0cm,clip,width=\linewidth]{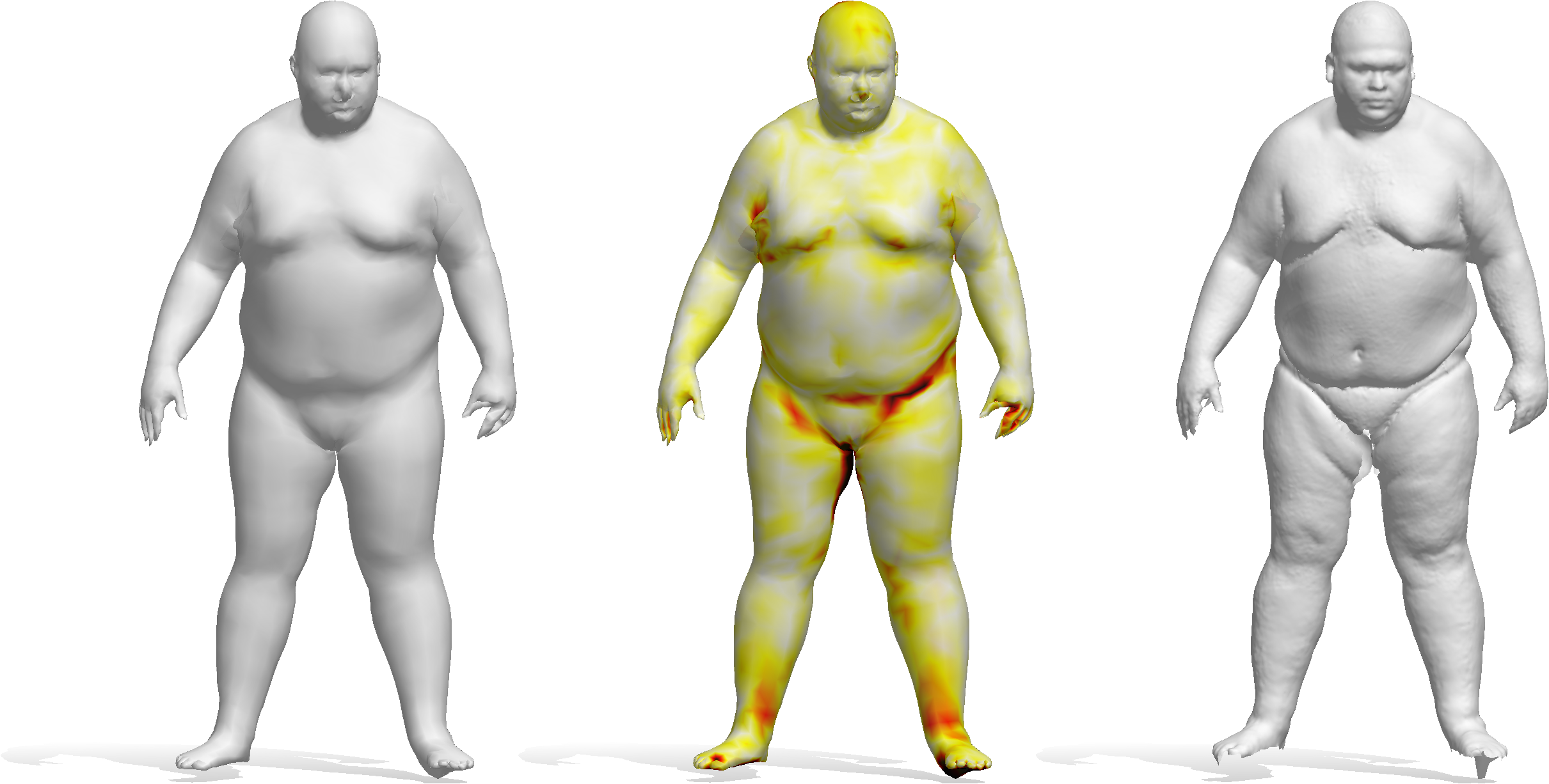}
  
  \put(18,52){\footnotesize{ZOSR}}
  \put(50,50){\footnotesize{}}
  \put(82,52){\footnotesize{Target}}
  \end{overpic}

    \label{fig:local}
\end{figure}
\end{center}

 \begin{center}
  \begin{figure}[t!]
  \vspace{0.5cm}
  \begin{overpic}
  [trim=0cm 0cm 0cm 0cm,clip,width=\linewidth]{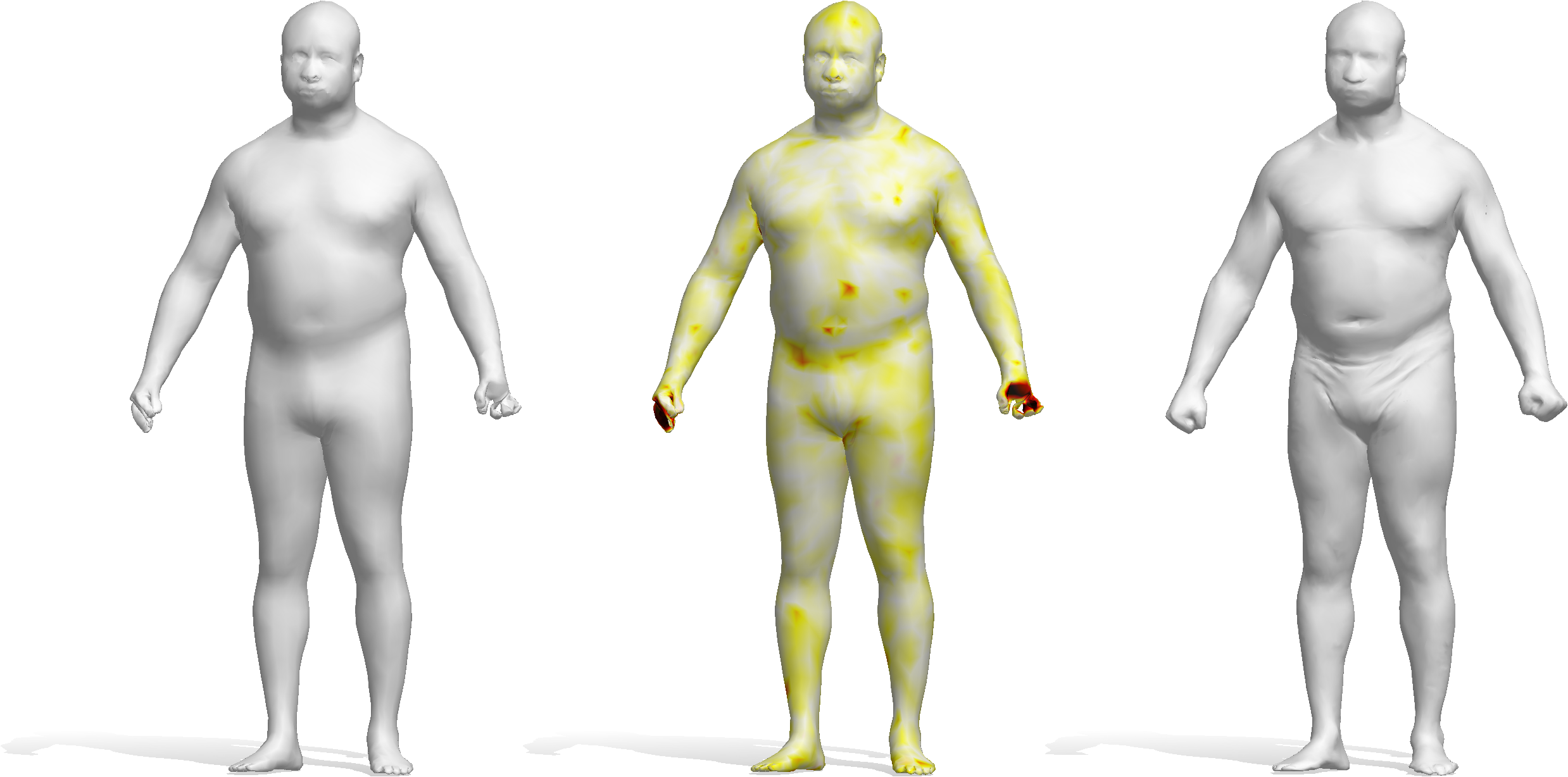}
  
  \put(18,50){\footnotesize{ZOSR}}
  \put(50,50){\footnotesize{}}
  \put(82,50){\footnotesize{Target}}
  \end{overpic}

    \label{fig:local}
\end{figure}
\end{center}

 \begin{center}
  \begin{figure}[t!]
  \vspace{0.5cm}
  \begin{overpic}
  [trim=0cm 0cm 0cm 0cm,clip,width=\linewidth]{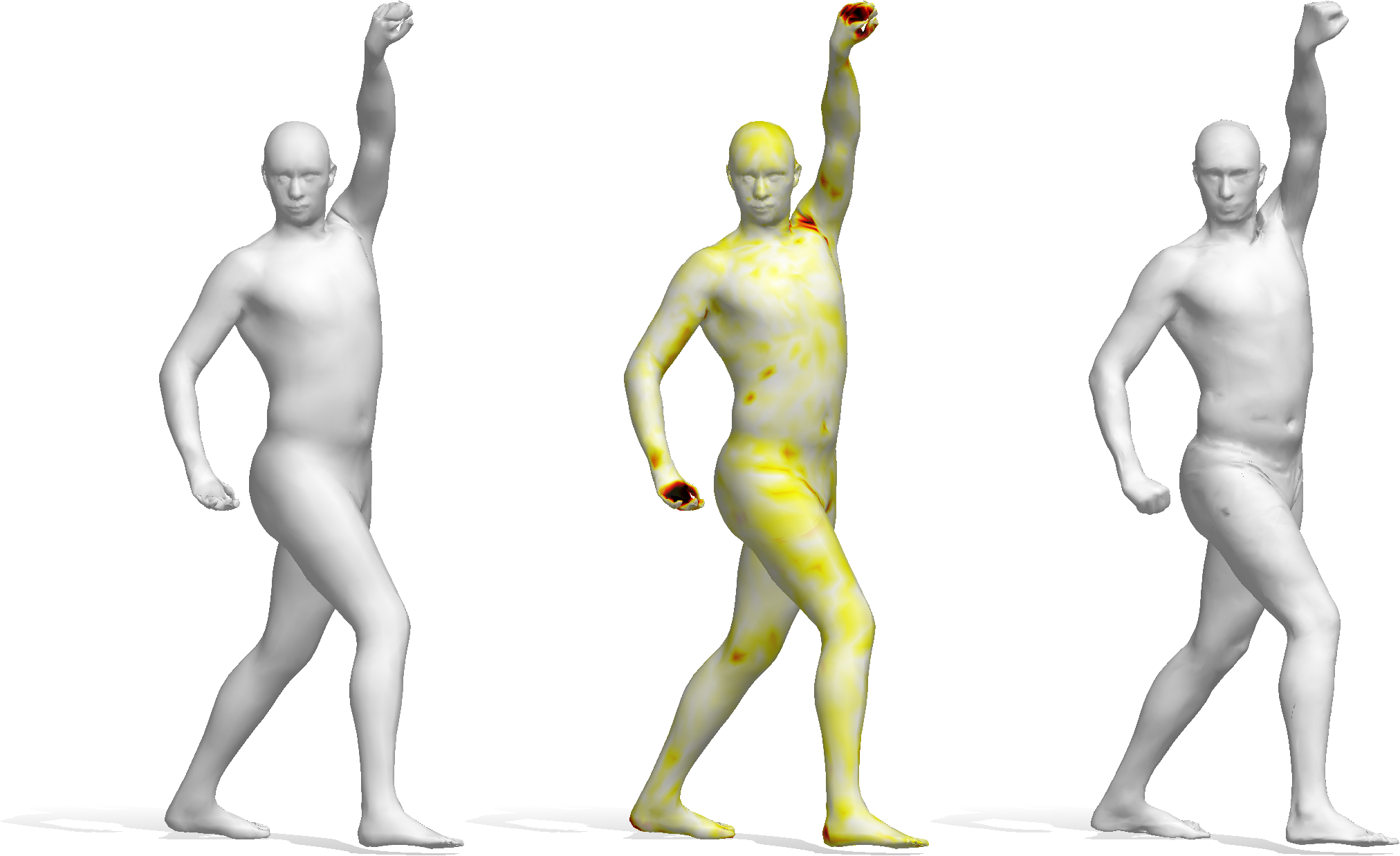}
  
  \put(18,64){\footnotesize{ZOSR}}
  \put(50,50){\footnotesize{}}
  \put(82,64){\footnotesize{Target}}
  \end{overpic}

    \label{fig:local}
\end{figure}
\end{center}

 \begin{center}
  \begin{figure}[t!]
  \vspace{0.5cm}
  \begin{overpic}
  [trim=0cm 0cm 0cm 0cm,clip,width=\linewidth]{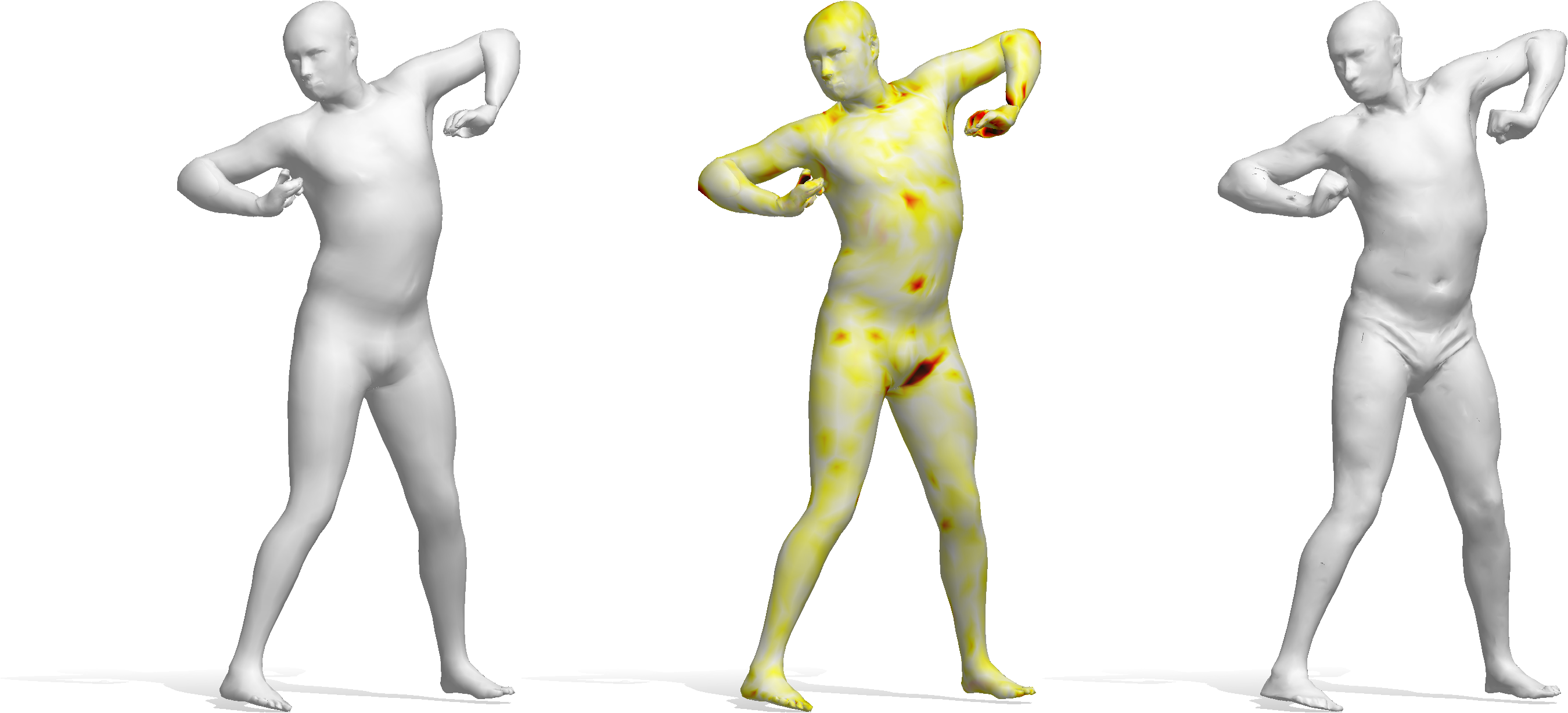}
  
  \put(17,48){\footnotesize{ZOSR}}
  \put(50,50){\footnotesize{}}
  \put(83,48){\footnotesize{Target}}
  \end{overpic}

    \label{fig:local}
\end{figure}
\end{center}

 \begin{center}
  \begin{figure}[t!]
  \vspace{0.5cm}
  \begin{overpic}
  [trim=0cm 0cm 0cm 0cm,clip,width=\linewidth]{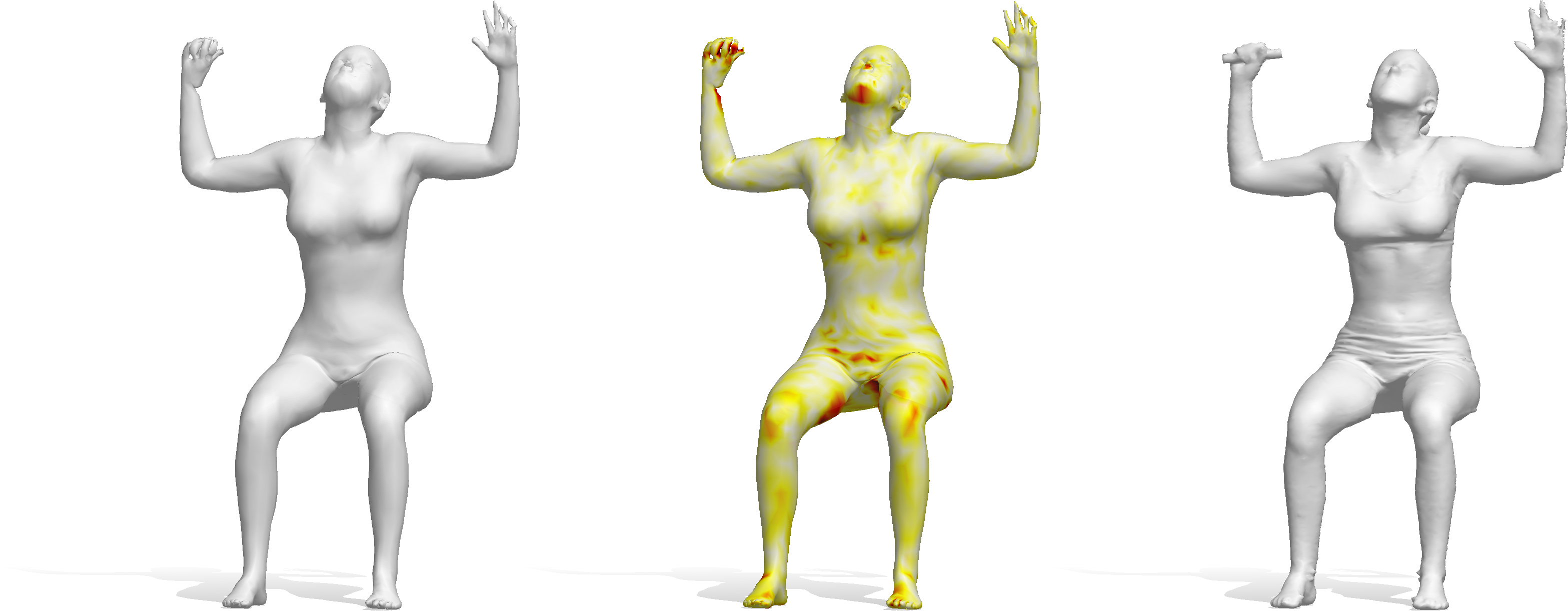}
  
  \put(18,41){\footnotesize{ZOSR}}
  \put(50,50){\footnotesize{}}
  \put(84,41){\footnotesize{Target}}
  \end{overpic}

    \label{fig:local}
\end{figure}
\end{center}

 \begin{center}
  \begin{figure}[t!]
  \vspace{0.5cm}
  \begin{overpic}
  [trim=0cm 0cm 0cm 0cm,clip,width=\linewidth]{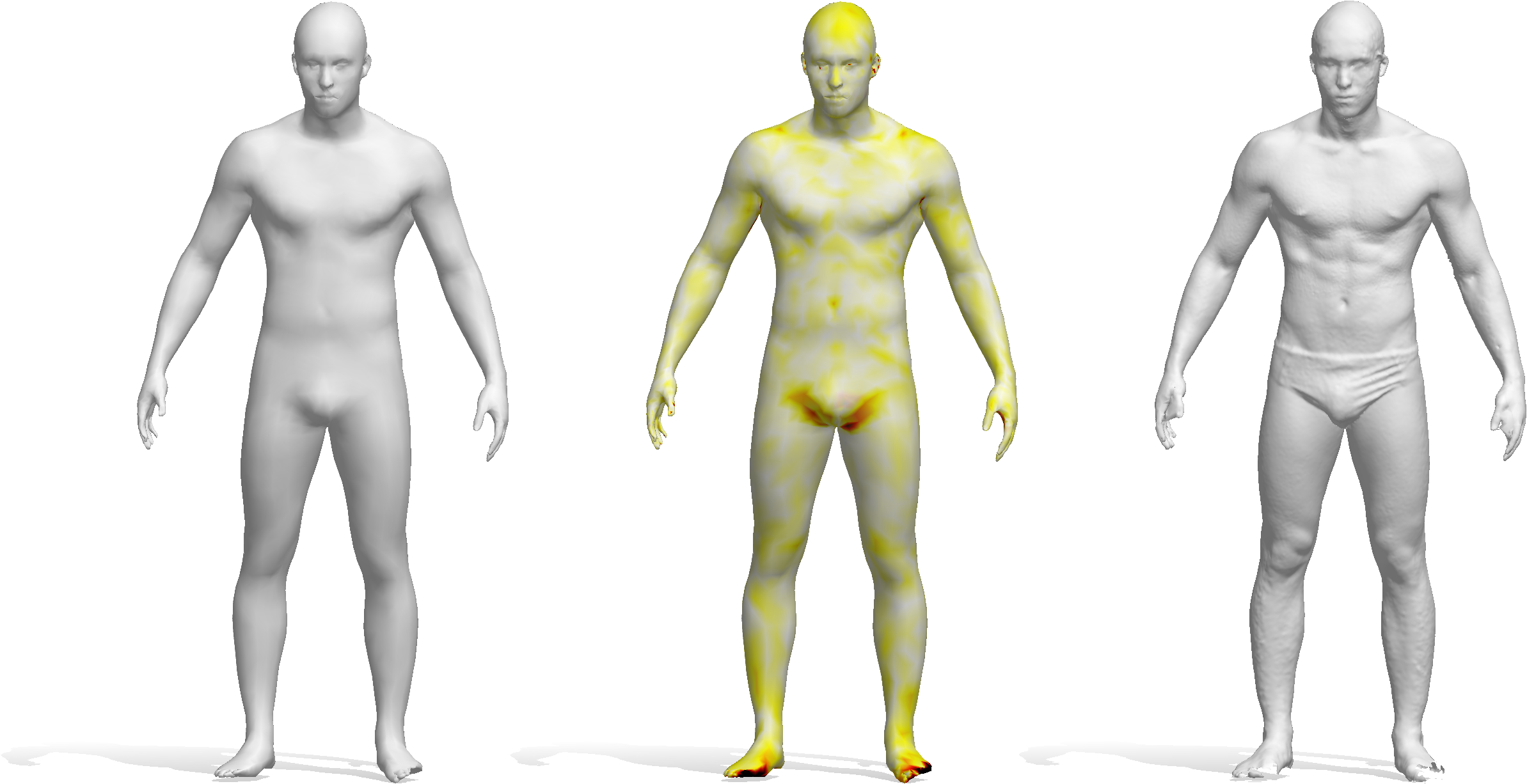}
  
  \put(18,53){\footnotesize{ZOSR}}
  \put(50,50){\footnotesize{}}
  \put(84,53){\footnotesize{Target}}
  \end{overpic}

    \label{fig:local}
\end{figure}
\end{center}

 \begin{center}
  \begin{figure*}[t!]
  \vspace{0.5cm}
  \begin{overpic}
  [trim=0cm 0cm 0cm 0cm,clip,width=\linewidth]{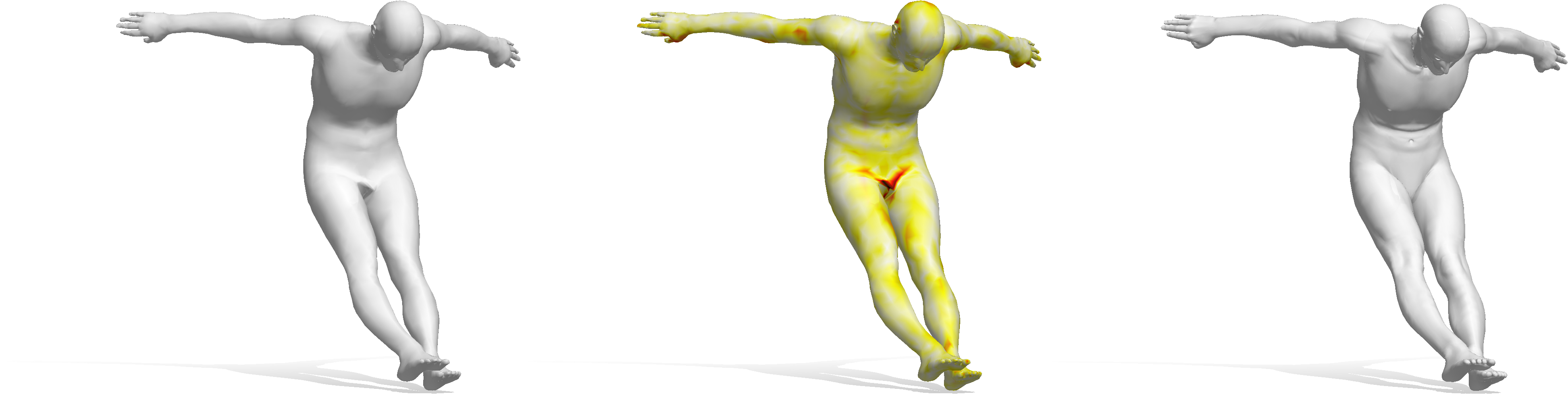}
  \put(23,27){\footnotesize{ZOSR}}
  \put(50,50){\footnotesize{}}
  \put(89,27){\footnotesize{Target}}
  \end{overpic}
    \label{fig:local}
\end{figure*}
\end{center}

\begin{center}
  \begin{figure*}[t!]
  \vspace{0.5cm}
  \begin{overpic}
  [trim=0cm 0cm 0cm 0cm,clip,width=\linewidth]{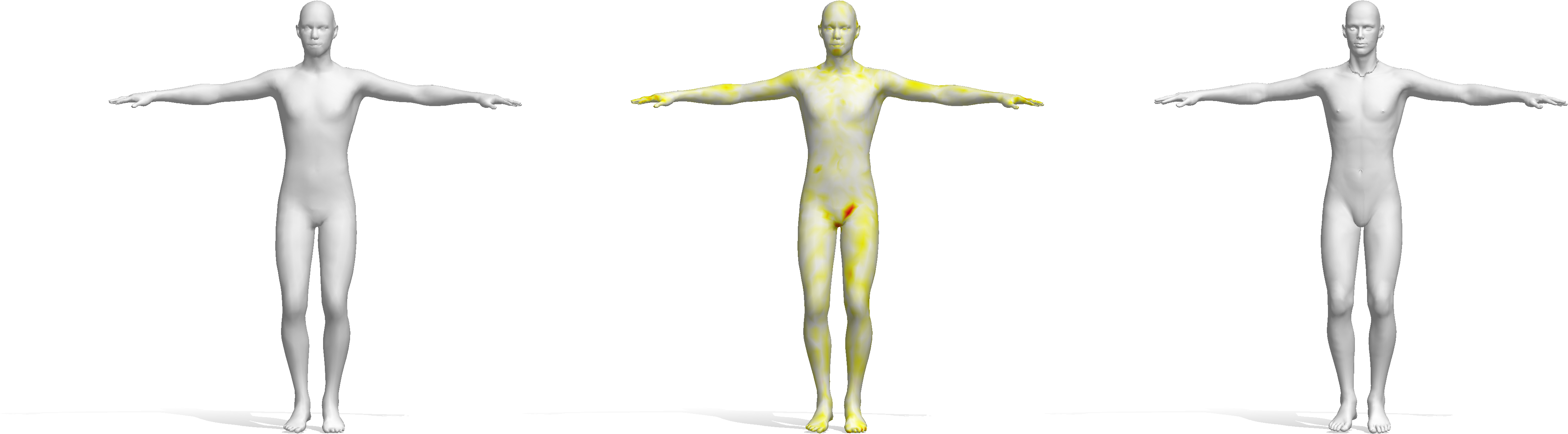}
  
  \put(18,29){\footnotesize{ZOSR}}
  \put(50,50){\footnotesize{}}
  \put(85,29){\footnotesize{Target}}
  \end{overpic}
    \label{fig:local}
\end{figure*}
\end{center}

 \begin{center}
  \begin{figure*}[t!]
  \vspace{0.5cm}
  \begin{overpic}
  [trim=0cm 0cm 0cm 0cm,clip,width=\linewidth]{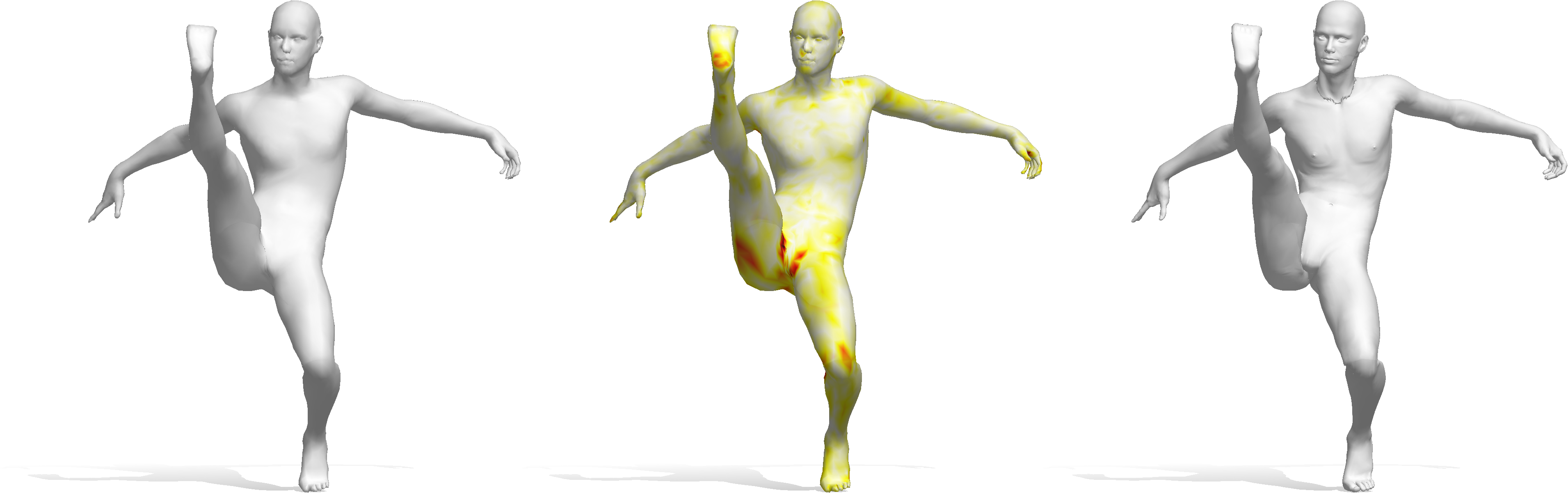}
  
  \put(17,33){\footnotesize{ZOSR}}
  \put(50,50){\footnotesize{}}
  \put(83.5,33){\footnotesize{Target}}
  \end{overpic}

    \label{fig:local}
\end{figure*}
\end{center}

\clearpage \newpage
\section{SMIL with kids}
\label{sec:SMIL}
Some experiments performed using the SMIL template over the KIDS dataset. We perform reasonably well over different pose and identities. Notice that majors artifacts are related to the different body proportions between SMIL and template and to the unnatural hands of SMIL. On the other hand, ZOSR is able to catch the majority of the geometry without any tuning of pipeline parameters. We are robust enough to have a fair representation. Please, see also related \textbf{video} for an example of fitting using SMIL.

 \begin{center}
  \begin{figure}[t!]
  \vspace{0.5cm}
  \begin{overpic}
  [trim=0cm 0cm 0cm 0cm,clip,width=0.5\linewidth]{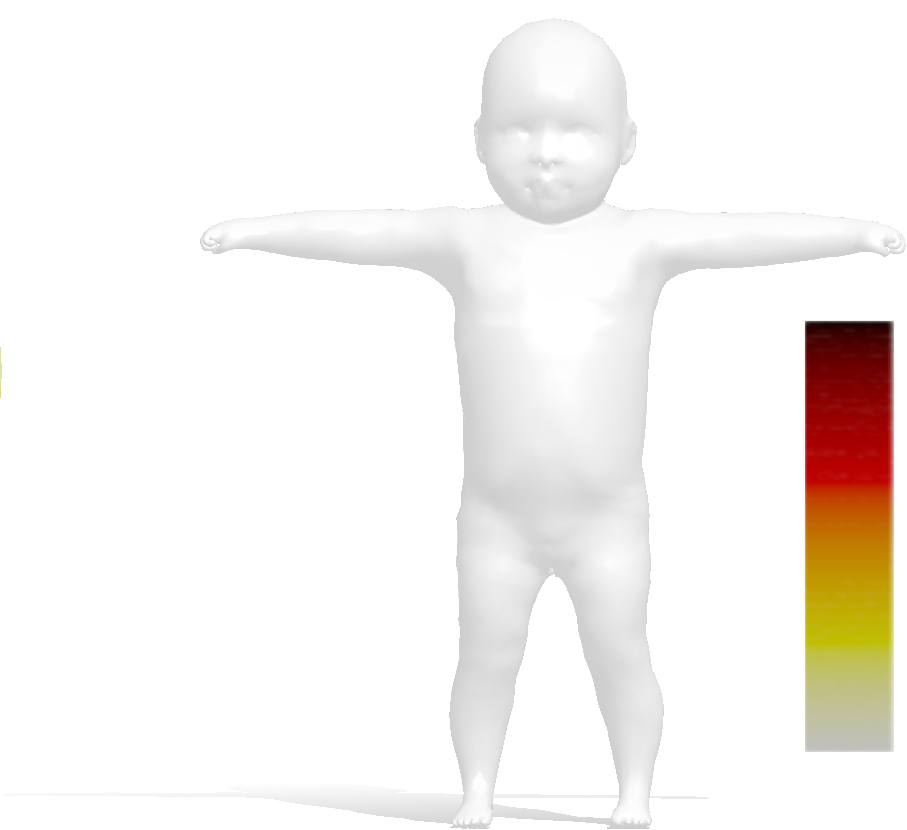}
  
  \put(100,55){\footnotesize{0.5cm}}
  \put(100,2){\footnotesize{0cm}}
  \end{overpic}

    \label{fig:smil}
\end{figure}
\end{center}

 \begin{center}
  \begin{figure}[t!]
  \vspace{0.5cm}
  \begin{overpic}
  [trim=0cm 0cm 0cm 0cm,clip,width=0.8\linewidth]{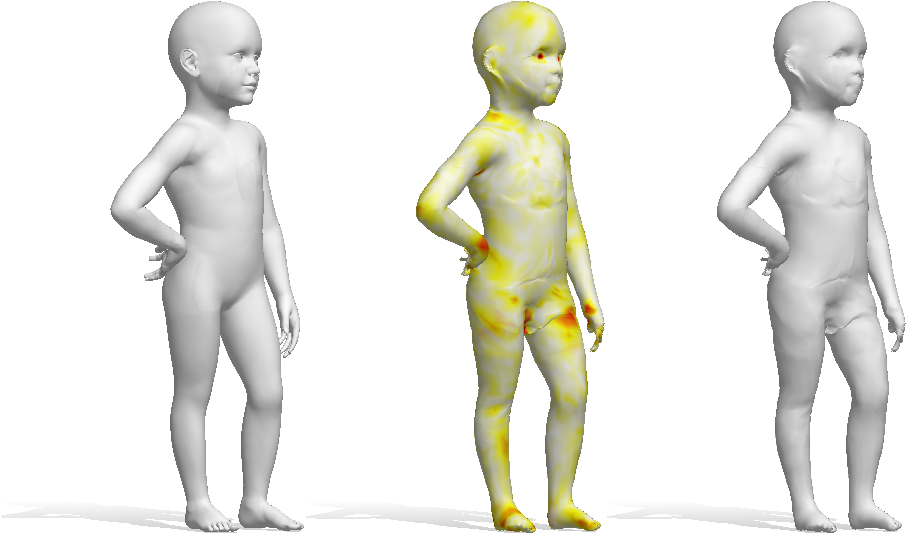}
  
  \put(18,60){\footnotesize{Target}}
  \put(50,50){\footnotesize{}}
  \put(84,60){\footnotesize{ZOSR}}
  \end{overpic}
    \label{fig:smil}
\end{figure}
\end{center}

\begin{center}
  \begin{figure}[t!]
  \vspace{0.5cm}
  \begin{overpic}
  [trim=0cm 0cm 0cm 0cm,clip,width=0.7\linewidth]{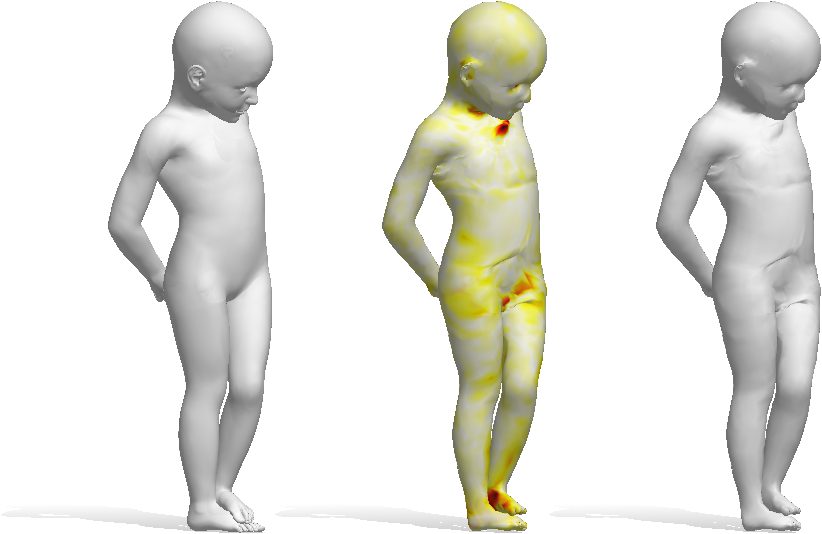}
 
  \put(20,66){\footnotesize{Target}}
  \put(50,50){\footnotesize{}}
  \put(86,66){\footnotesize{ZOSR}}
  \end{overpic}

    \label{fig:smil}
\end{figure}
\end{center}

\begin{center}
  \begin{figure}[t!]
  \vspace{0.5cm}
  \begin{overpic}
  [trim=0cm 0cm 0cm 0cm,clip,width=\linewidth]{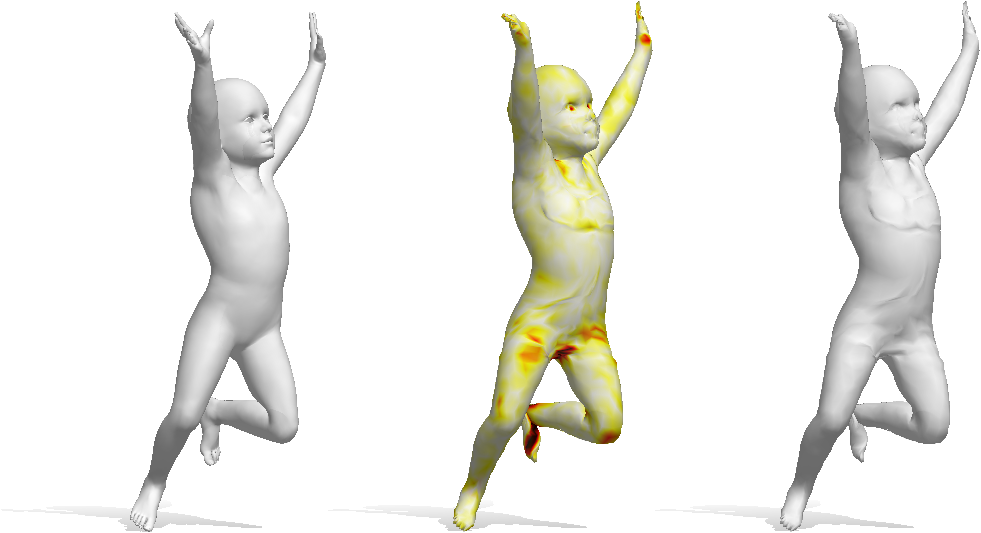}
  
  \put(19,56){\footnotesize{Target}}
  \put(50,50){\footnotesize{}}
  \put(87,56){\footnotesize{ZOSR}}
  \end{overpic}

    \label{fig:smil}
\end{figure}
\end{center}

\begin{center}
  \begin{figure}[t!]
  \vspace{0.5cm}
  \begin{overpic}
  [trim=0cm 0cm 0cm 0cm,clip,width=\linewidth]{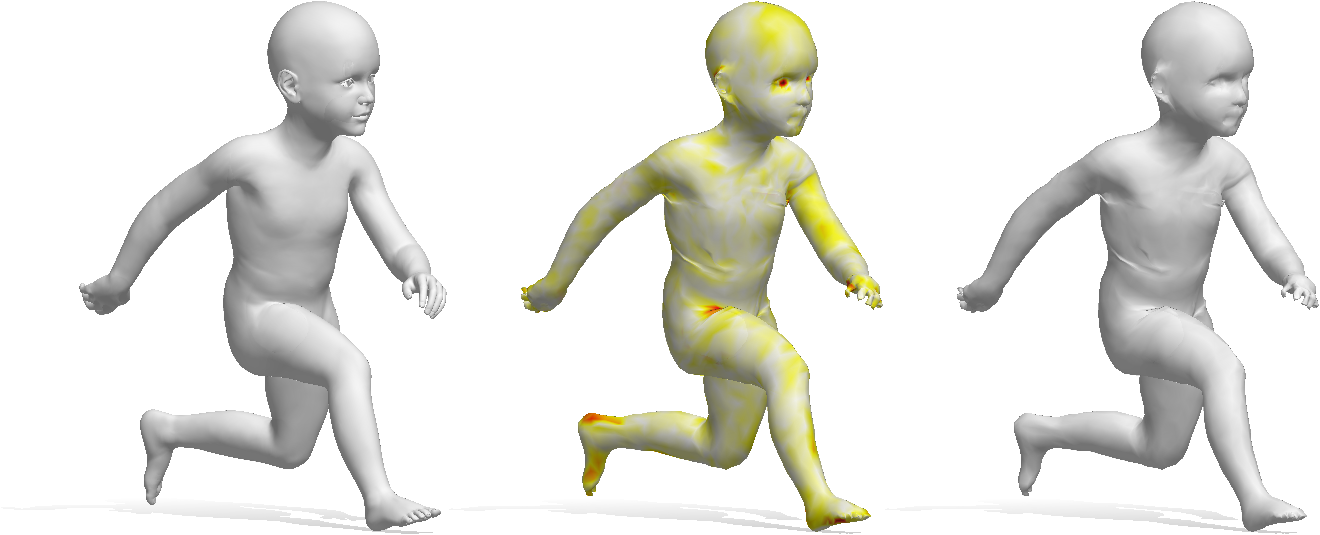}
  \put(19,43){\footnotesize{Target}}
  \put(50,50){\footnotesize{}}
  \put(85,43){\footnotesize{ZOSR}}
  \end{overpic}
    \label{fig:smil}
\end{figure}
\end{center}

\begin{center}
  \begin{figure}[t!]
  \vspace{0.5cm}
  \begin{overpic}
  [trim=0cm 0cm 0cm 0cm,clip,width=\linewidth]{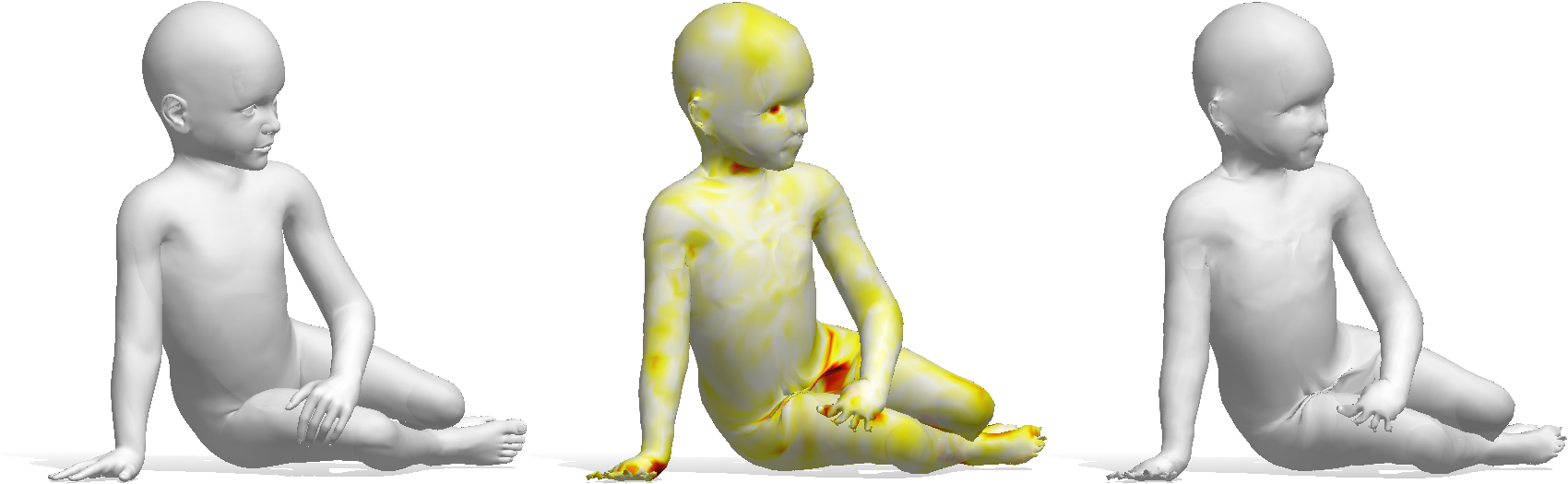}
  
  \put(10,34){\footnotesize{Target}}
  \put(50,50){\footnotesize{}}
  \put(76,34){\footnotesize{ZOSR}}
  \end{overpic}

    \label{fig:smil}
\end{figure}
\end{center}

\begin{center}
  \begin{figure}[t!]
  \vspace{0.5cm}
  \begin{overpic}
  [trim=0cm 0cm 0cm 0cm,clip,width=\linewidth]{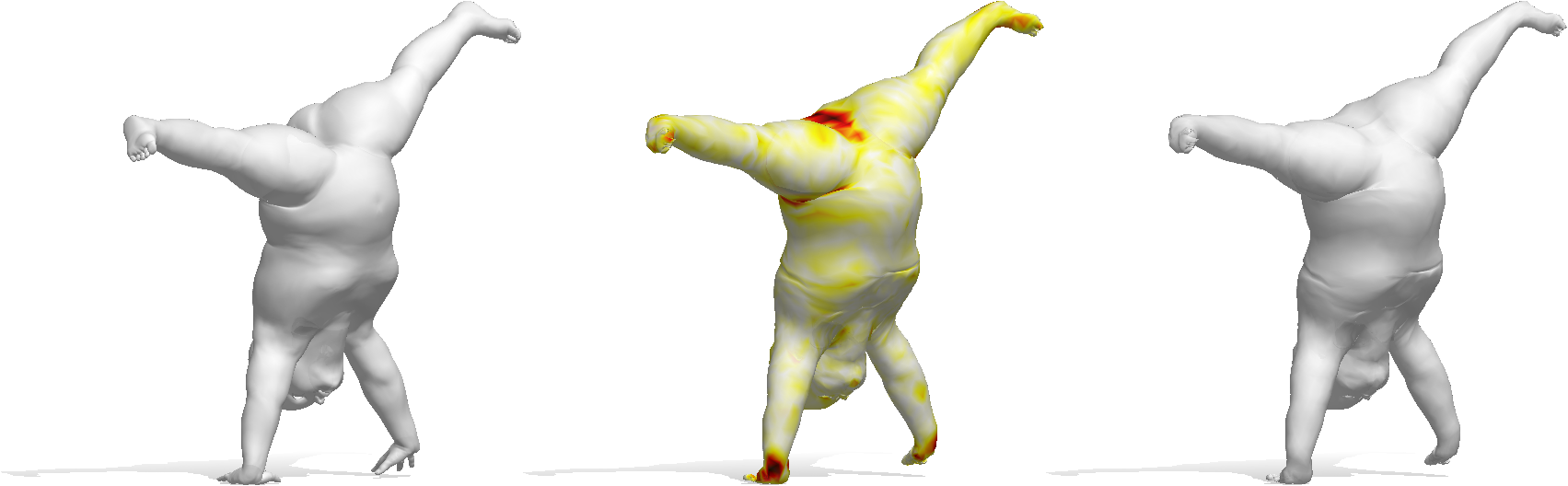}
  
  \put(18,35){\footnotesize{Target}}
  \put(50,50){\footnotesize{}}
  \put(82,35){\footnotesize{ZOSR}}
  \end{overpic}

    \label{fig:smil}
\end{figure}
\end{center}

\begin{center}
  \begin{figure*}[t!]
  \vspace{0.5cm}
  \begin{overpic}
  [trim=0cm 0cm 0cm 0cm,clip,width=\linewidth]{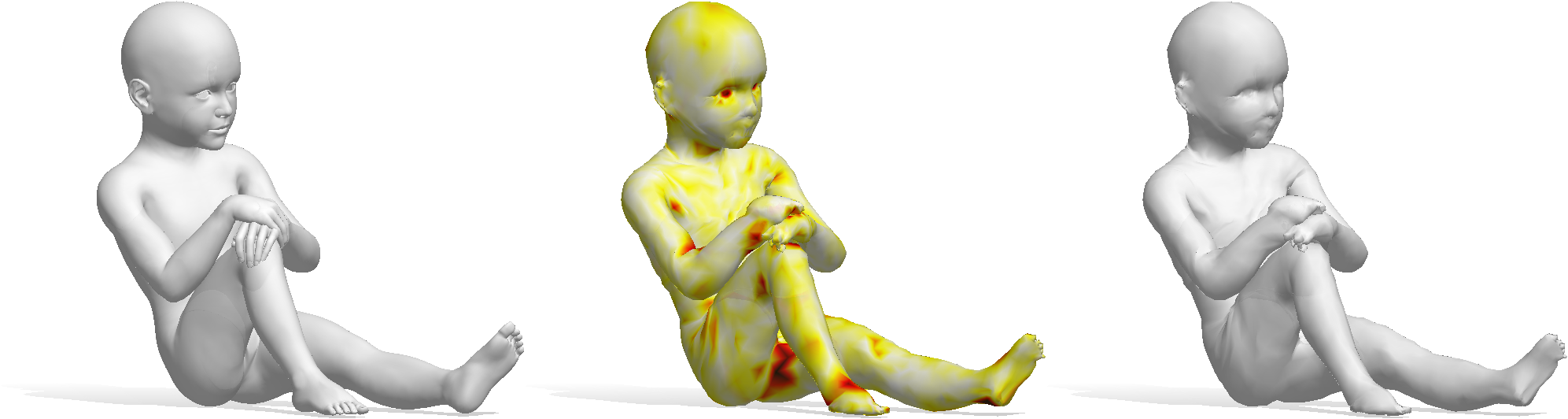}
  
  \put(10,28){\footnotesize{Target}}
  \put(50,50){\footnotesize{}}
  \put(78,28){\footnotesize{ZOSR}}
  \end{overpic}

    \label{fig:smil}
\end{figure*}
\end{center}

\begin{center}
  \begin{figure*}[t!]
  \vspace{0.5cm}
  \begin{overpic}
  [trim=0cm 0cm 0cm 0cm,clip,width=\linewidth]{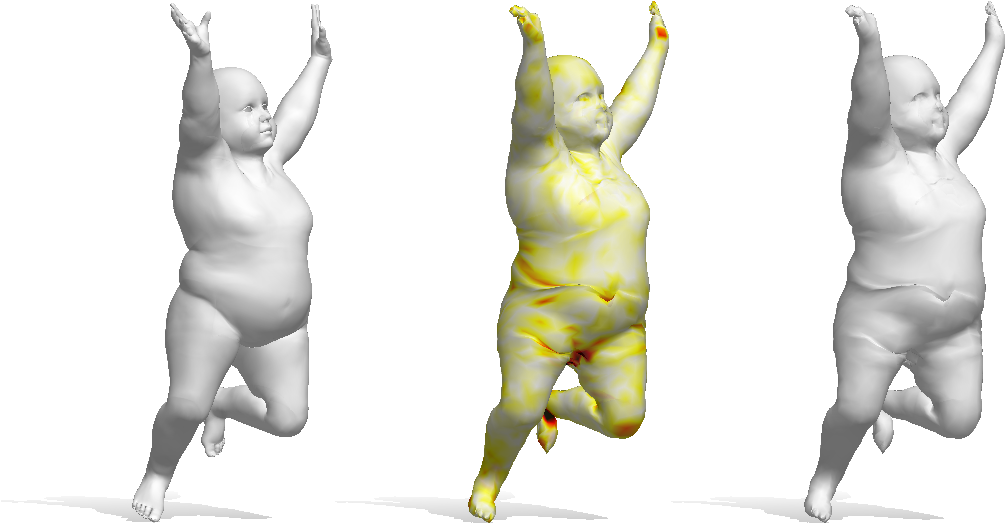}
  
  \put(20,54){\footnotesize{Target}}
  \put(50,50){\footnotesize{}}
  \put(89,54){\footnotesize{ZOSR}}
  \end{overpic}

    \label{fig:smil}
\end{figure*}
\end{center}

\clearpage \newpage
\section{Texture Transfer}
\label{sec:TextureTransfer}
\begin{center}
\begin{figure}[t!]
  \vspace{0.5cm}
  \begin{overpic}
  [trim=0cm 0cm 0cm 0cm,clip,width=\linewidth]{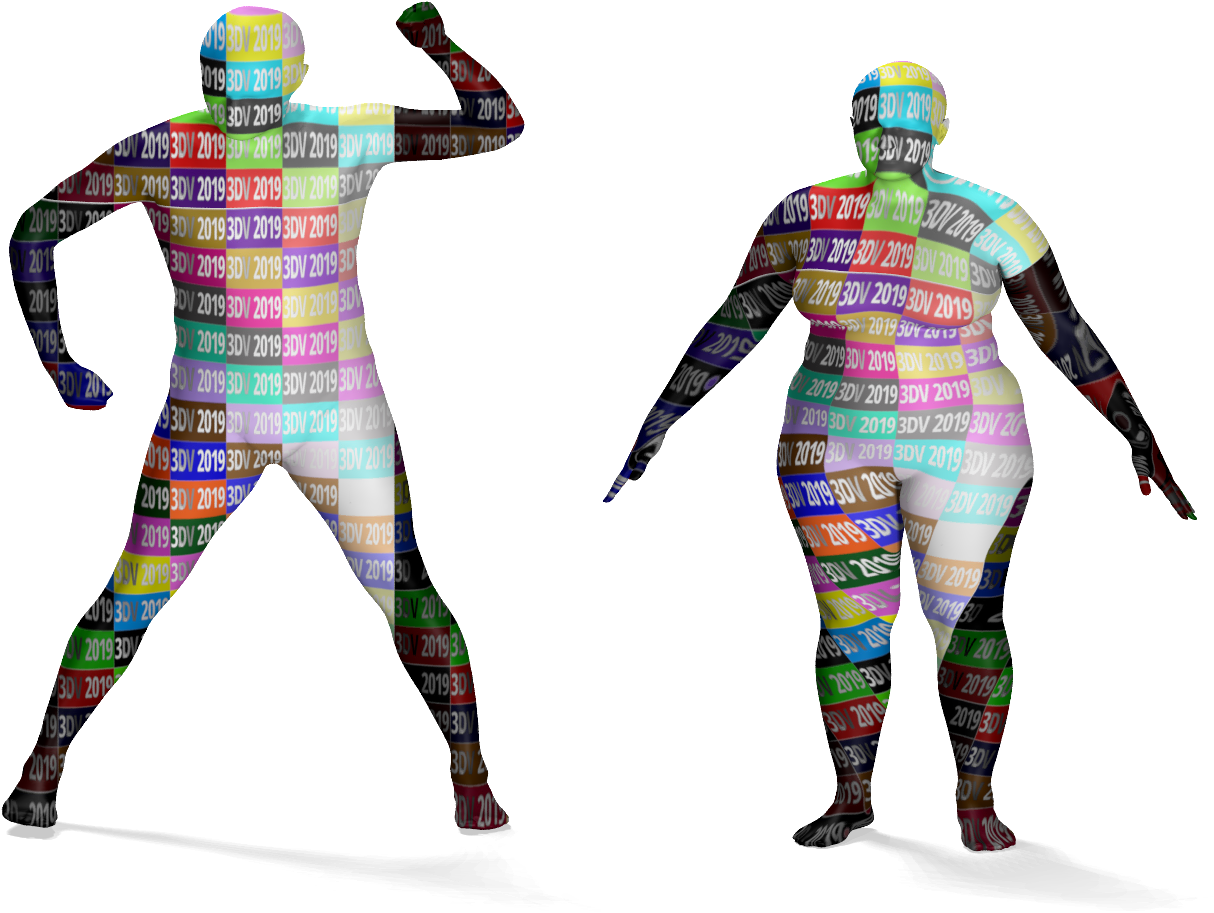}
  
  \put(15,-2.5){\footnotesize{Original}}
  \put(69,-2.5){\footnotesize{Transfer}}
  \end{overpic}
  \vspace{0.15cm}
      \caption{}
    \label{fig:text1}
\end{figure}
\end{center}

\begin{center}
\begin{figure}[t!]
  \vspace{0.5cm}
  \begin{overpic}
  [trim=0cm 0cm 0cm 0cm,clip,width=\linewidth]{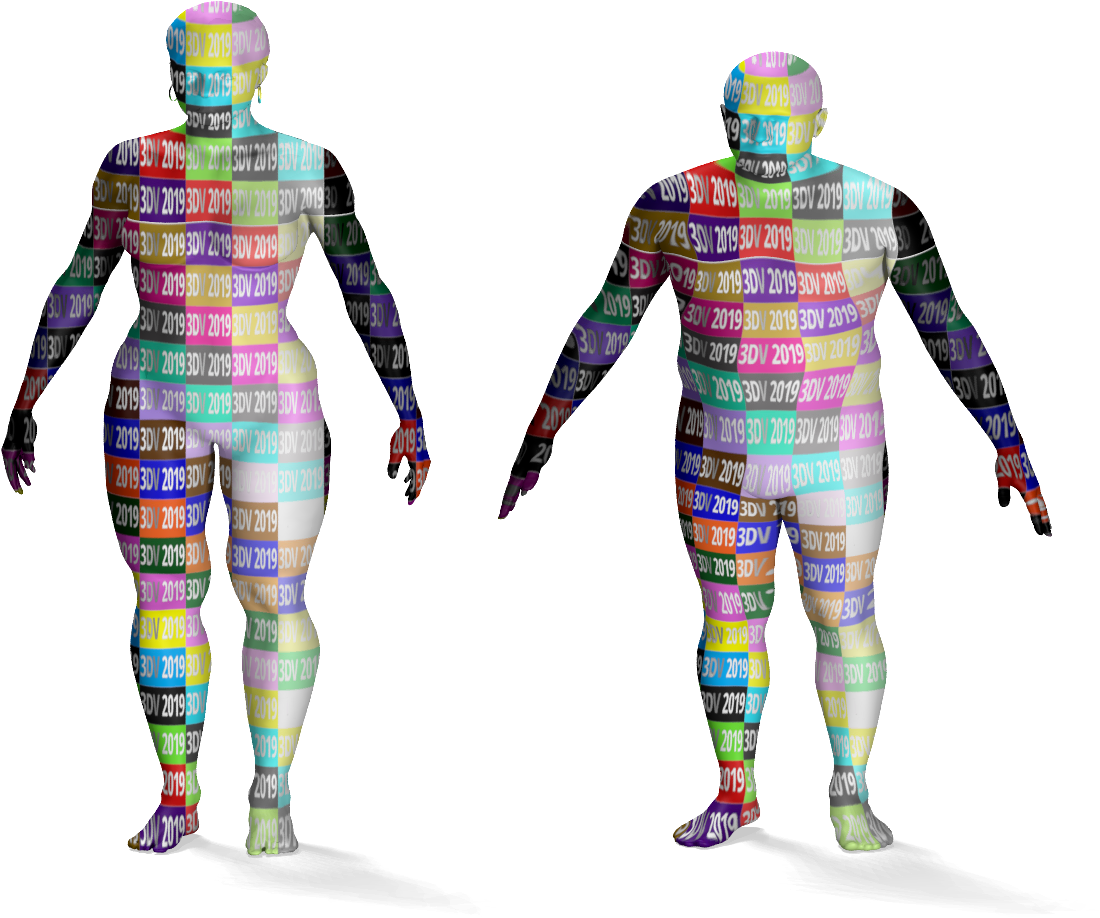}
  
  \put(15,-2.5){\footnotesize{Original}}
  \put(69,-2.5){\footnotesize{Transfer}}
  \end{overpic}
  \vspace{0.15cm}
      \caption{}
    \label{fig:text2}
\end{figure}
\end{center}

\begin{center}
\begin{figure}[t!]
  \vspace{0.5cm}
  \begin{overpic}
  [trim=0cm 0cm 0cm 0cm,clip,width=\linewidth]{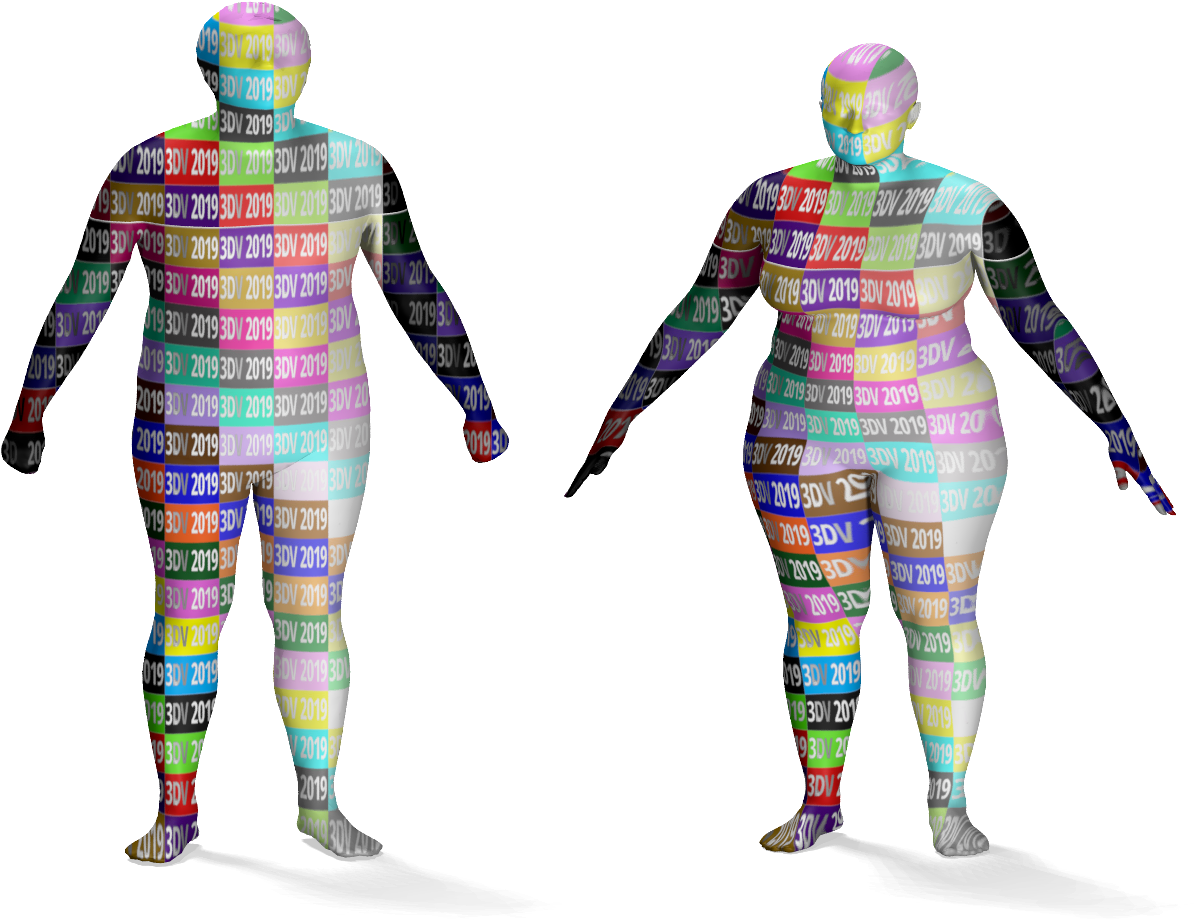}
  
  \put(15,-2.5){\footnotesize{Original}}
  \put(69,-2.5){\footnotesize{Transfer}}
  \end{overpic}
  \vspace{0.15cm}
      \caption{}
    \label{fig:text3}
\end{figure}
\end{center}

\begin{center}
\begin{figure}[t!]
  \vspace{0.5cm}
  \begin{overpic}
  [trim=0cm 0cm 0cm 0cm,clip,width=\linewidth]{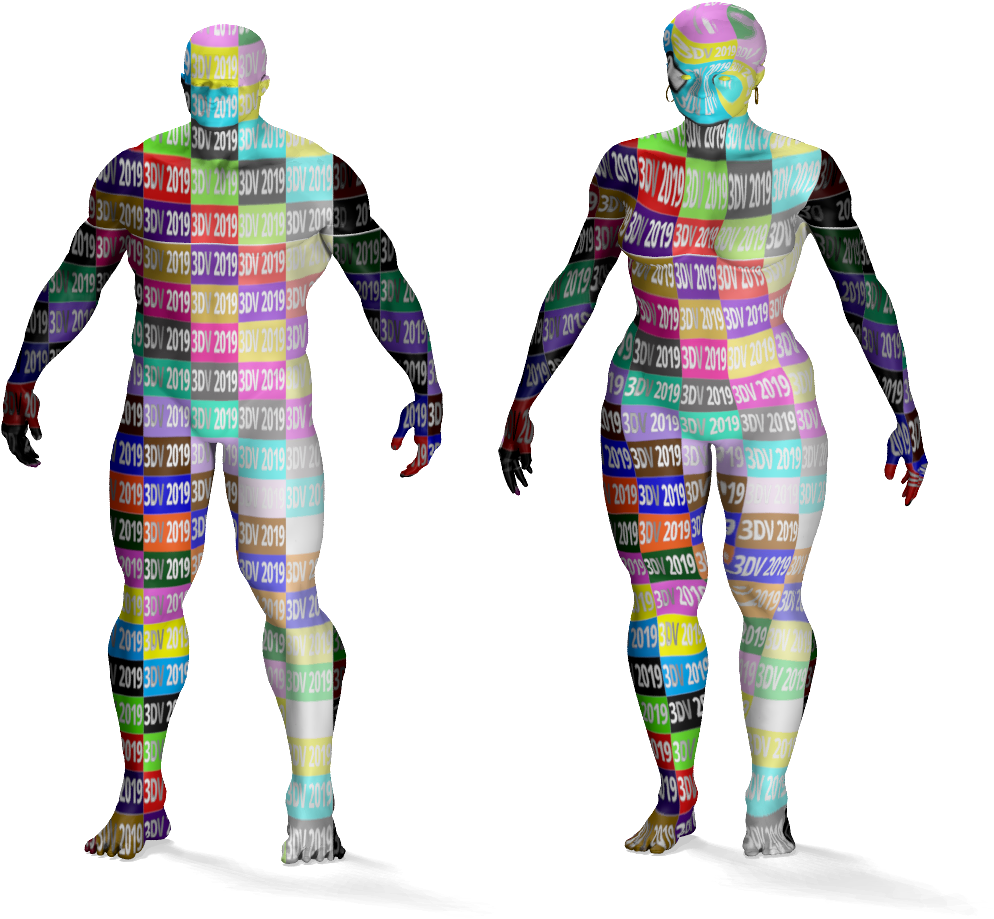}
  
  \put(15,-2.5){\footnotesize{Original}}
  \put(69,-2.5){\footnotesize{Transfer}}
  \end{overpic}
  \vspace{0.15cm}
      \caption{}
    \label{fig:text4}
\end{figure}
\end{center}

\begin{center}
\begin{figure}[t!]
  \vspace{0.5cm}
  \begin{overpic}
  [trim=0cm 0cm 0cm 0cm,clip,width=\linewidth]{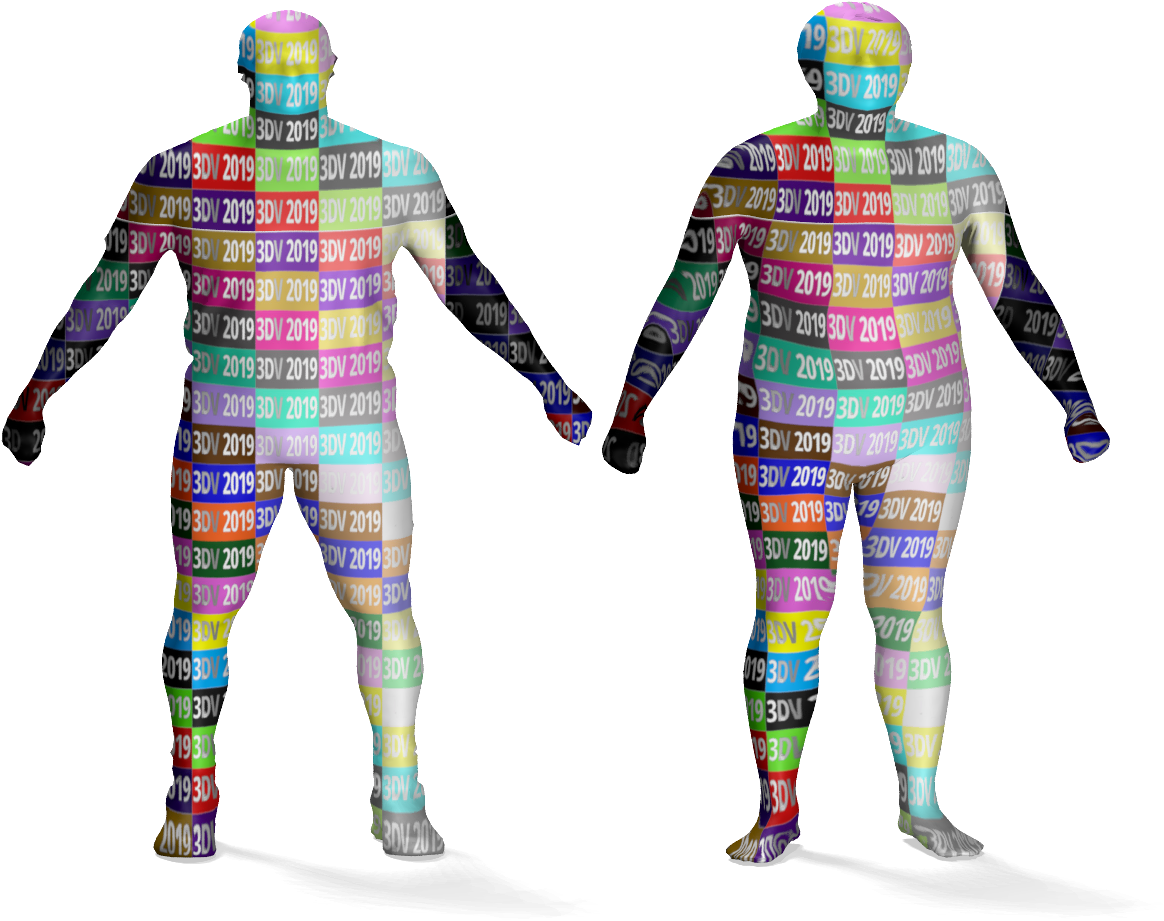}
  
  \put(15,-2.5){\footnotesize{Original}}
  \put(69,-2.5){\footnotesize{Transfer}}
  \end{overpic}
  \vspace{0.15cm}
      \caption{}
    \label{fig:text_}
\end{figure}
\end{center}

\begin{center}
\begin{figure}[t!]
  \vspace{0.5cm}
  \begin{overpic}
  [trim=0cm 0cm 0cm 0cm,clip,width=\linewidth]{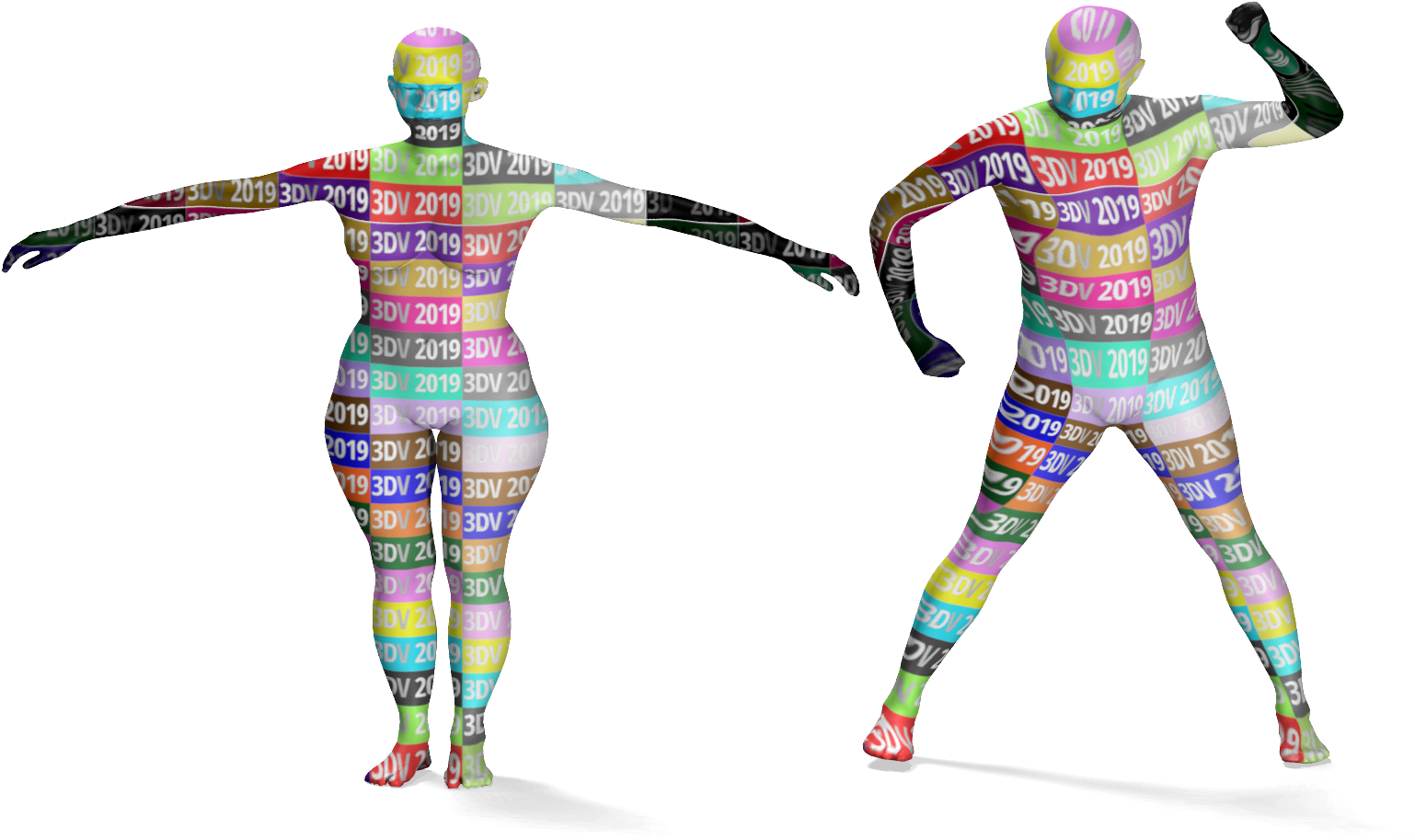}
  
  \put(15,-2.5){\footnotesize{Original}}
  \put(69,-2.5){\footnotesize{Transfer}}
  \end{overpic}
  \vspace{0.15cm}
      \caption{}
    \label{fig:text5}
\end{figure}
\end{center}

In Figures from~\ref{fig:text1} to ~\ref{fig:text5} we show some of the texture transfer results obtained on random pairs from the SHREC 19 connectivity track benchmark~\cite{SHREC19}. We remark that these shape are completely not isometric and represented by meshes with different connectivity and thus they can be considered very challenging pairs.

\clearpage \newpage

\end{document}